\documentclass{article} 
\usepackage{iclr2027_conference,times}

\usepackage{amsmath,amsfonts,bm}

\usepackage{hyperref}
\usepackage{url}

\usepackage{xcolor}
\usepackage{xspace}
\usepackage{graphicx}
\usepackage[caption=false,font=footnotesize]{subfig}
\usepackage{booktabs}
\usepackage{colortbl}
\usepackage{multirow}
\usepackage{wrapfig}
\usepackage{float}
\usepackage{tcolorbox}
\usepackage{longtable}

\newcommand{\dataset}{\textsc{LiveTS-Suite}\xspace}
\newcommand{\method}{\textsc{TimeOmni-Live}\xspace}

\title{\includegraphics[height=1em]{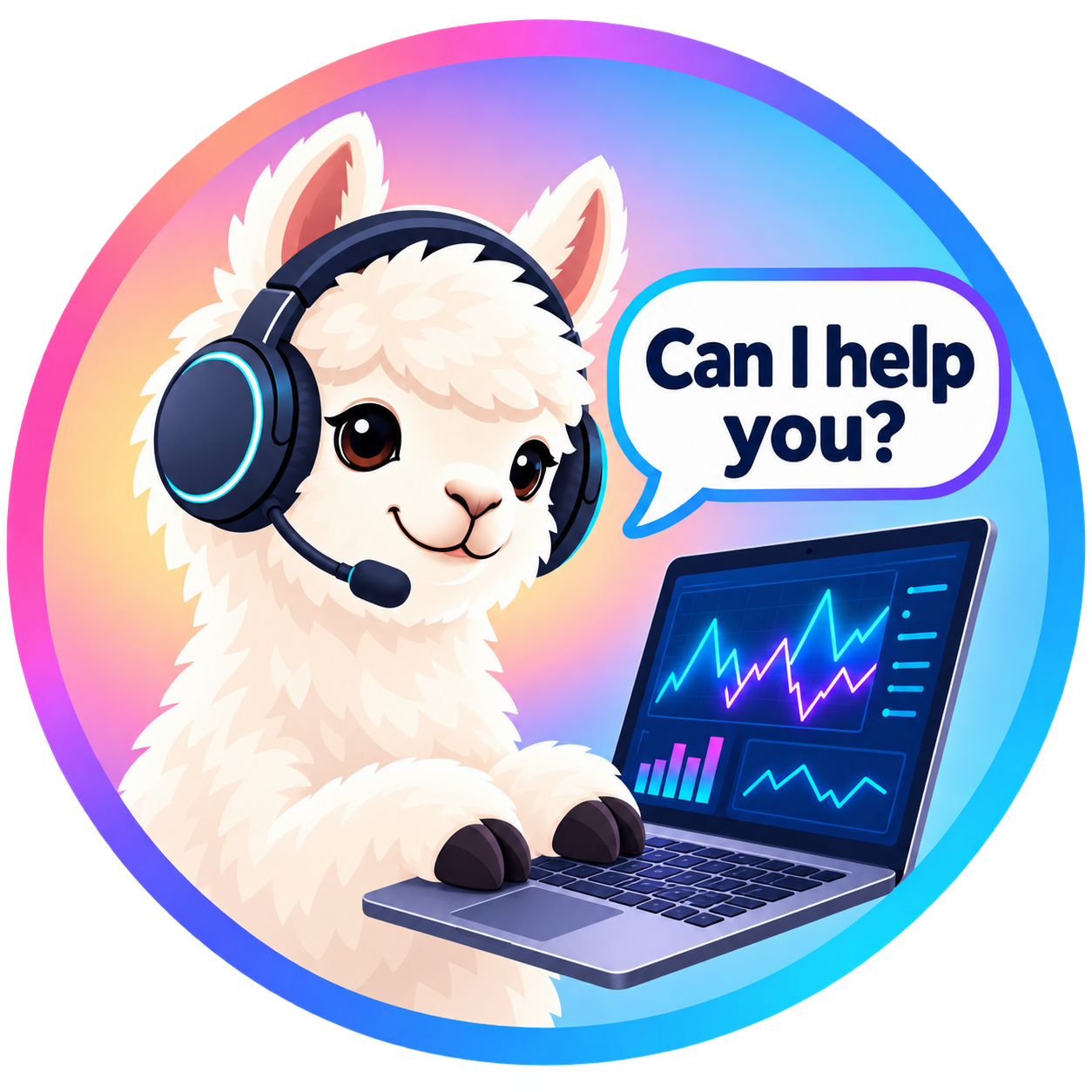}\hspace{0.3em}
\method: Continuous \& Live Time Series Understanding with Long-Term Memory}

\author{Tong Guan$^{1,2}$\quad Alexander Chemeris$^{3}$\quad Yiqing Guo$^{4}$\quad Warren Jin$^{4}$\quad Zhao Li$^{5}$ \\
 \textbf{Shirui Pan$^{1}$\thanks{Correspondence to: S. Pan $<$s.pan@griffith.edu.au$>$ and M. Jin $<$mingjinedu@gmail.com$>$ } \quad Ming Jin$^{1}$\footnotemark[1]} \\
 \\
  $^{1}$Griffith University \quad
  $^{2}$Zhejiang University \quad
  $^{3}$Langotime \quad
  $^{4}$CSIRO \quad 
  $^{5}$Zhejiang Lab
}

\iclrfinalcopy 
\usepackage{graphicx,booktabs,tabularx,array,xcolor}
\newcolumntype{Y}{>{\raggedright\arraybackslash}X}

\begin{document}

\maketitle

\begin{abstract}
Existing time series language models predominantly follow an offline, one-shot paradigm: they process fixed-length inputs independently and do not retain past states. This static approach fundamentally misaligns with the continuous nature of real-world streaming applications and fails to maintain the historical context necessary for cross-window reasoning. To bridge this gap, we introduce Continuous \& Live Time Series Understanding (CL-TSU), a novel paradigm that shifts the focus toward maintaining a persistent, stateful understanding over open-ended data streams. To systematically evaluate this paradigm, we present \textbf{\dataset}, the first comprehensive data suite tailored for CL-TSU, supporting data generation, model training, and systematic evaluation across 18 tasks spanning six core capabilities. It comprises over 5K streams totaling 2.2B timesteps, annotated with 15K specific response events. We further design an evaluation protocol that jointly assesses response timeliness, content accuracy, and computational efficiency. Building on this foundation, we propose \textbf{\method}, the first live time series understanding model. It integrates a hybrid time series encoding for natively temporal input, a stream-persistent hybrid memory mechanism to effectively preserve exact historical evidence and long-term task states, and an elastic stream-inference synchronization strategy to reduce processing lag. Evaluations show that \method establishes a strong baseline for CL-TSU, outperforming frontier LLMs (90.0 vs. 79.3 for Opus 5) with only 4B parameters and achieving $206\times$ lower p95 latency than its backbone. Deployment on a Jetson edge device further shows its ability to support real-time continuous temporal understanding locally at the edge.
\end{abstract}

\section{Introduction}
Time series are innately a continuous modality, consisting of observations collected sequentially over time~\citep{TiRex2}. These observations are typically generated by sensors that continuously measure ongoing processes without a definitive endpoint~\citep{tsisSensorData}. Yet, most existing time series research is primarily designed for offline settings, where a model receives a predefined input window in a single model call, produces a single prediction or decision, and then halts. Conventional models typically do not preserve task states across independent windows~\citep{wu2020connecting}. Typical tasks in this category include forecasting and classification under fixed input lengths~\citep{timebenchmark, cauker}. Recent work has broadened the scope of the field to include one-shot time series understanding and reasoning~\citep{timemqa,timeomni-1}.

However, this offline modeling paradigm has two major limitations. (1) Fixed-window, post-hoc analysis is poorly aligned with applications that require timely decisions as measurements arrive. For instance, modern monitoring systems generate high-frequency observations that necessitate online analysis, ranging from millisecond-level telemetry in IoT scheduling networks~\citep{TSNSurvey} to minute-level traffic flow data critical for real-time urban dispatching~\citep{Traffic-R1}. (2) This paradigm discards valuable contextual information beyond the predefined temporal boundaries. For example, streamflow forecasting and flood-risk estimation draw on historical streamflow records and previously observed flood events, which may lie far outside the immediate input window~\citep{Climate-Extremes}. Accordingly, an appropriate modeling paradigm should align computation with continuous data arrival while retaining and reasoning over accumulated temporal context.

The broader multimodal learning community has recently begun addressing analogous requirements in continuously generated audio, video, and embodied intelligence~\citep{lin2026stepaudio3realtimetechnical, zhangStreamArenaContinuousInteractive2026, blackRealTimeExecutionAction2025, Wang_2026_CVPR_StreamEQA}. Audio-Interaction~\citep{AudioInteractionModel} adopts an always-on perceive--decide--respond loop, while AURA~\citep{Always-On-video-Understanding} continuously processes video streams for real-time question answering and proactive assistance. Vinci2~\citep{Vinci2ProactiveVideo} reasons over accumulated egocentric context to determine whether and when assistance is warranted. These studies represent a paradigm shift from one-shot analysis with isolated inputs to continuous understanding supported by persistent state. Yet, an equivalent setting has not been systematically formulated or evaluated for time series intelligence (TSI). We therefore introduce {Continuous \& Live Time Series Understanding (CL-TSU)}, a new paradigm where a model operates continuously over open-ended streams and responds dynamically as observations and instructions evolve.

However, two key limitations hinder the development of CL-TSU: \textbf{(1) The lack of dedicated data systematically targeting CL-TSU capability.} The vast majority of time series evaluations remain largely confined to the one-shot setting~\citep{gifteval,timeomni-vl}. Although some studies on real-time systems evaluate models by processing streams continuously, they primarily focus on timeliness and latency~\citep{streamingad,Real-Time-ECG}, largely ignoring whether earlier observations influence later decisions. Consequently, such tasks do not measure a model's ability to maintain context, update internal states, and reason across a continuing stream. \textbf{(2) The architectural paradigm required to support CL-TSU remains largely unexplored.} Currently, processing streaming time series primarily relies on executing predefined code over the data stream~\citep{stream-survey,rule-based-monitoring-1}. These approaches struggle to capture long-term temporal dependencies and lack the flexibility to perform continuous multimodal understanding. Although some recent agent frameworks attempt to introduce capabilities for automatically updating predefined code~\citep{coding_monitoring-2,agentic_coding_monitoring-1}, these methods fundamentally lack a native understanding of the time series modality. This leaves CL-TSU without a unified modeling framework capable of natively processing continuous time series streams and maintaining persistent memory over time.

Filling these gaps drives us to formulate CL-TSU first. We argue that CL-TSU goes beyond faster one-shot analysis and requires a systematic assessment of a model's ability to retain and utilize historical context over time. Guided by this principle, we introduce \textbf{\dataset}, the first data suite for CL-TSU, which covers six core capabilities across 18 tasks. The evaluation begins with \textit{Instantaneous Conditions (C1)} in current observations and extends to \textit{Sequential Conditions (C2)} requiring recent historical reasoning. To assess persistent state, it challenges models with \textit{Event Tracking (C3)} for ongoing records and \textit{Event Reference (C4)} to retrieve distant historical data. Finally, \dataset evaluates dynamic adaptability via \textit{Online Instructions (C5)} for mid-stream queries and rule changes, culminating in \textit{Ultra-Long Reliability (C6)} to test sustained execution. 
Building on this data foundation, we present \textbf{\method}, the first streaming foundation model for CL-TSU. We introduce \textit{(a) hybrid time series encoding} to natively support the time series modality. To allow subsequent decisions to be informed by earlier related observations, we design a \textit{(b) stream-persistent hybrid memory} mechanism, which retains memory of both textual and time series modalities. Furthermore, we introduce \textit{(c) elastic stream-inference synchronization} to reduce processing lag. During inference, \method generates responses and updates its memory while continuously processing the incoming data stream.
Our contributions lie in three aspects:
\begin{enumerate}
\setlength{\leftskip}{-2.1em}
\item \textbf{New Paradigm.} We formulate Continuous \& Live Time Series Understanding (CL-TSU), a novel setting that shifts the paradigm from conventional offline, one-shot analysis to maintaining a persistent, stateful understanding over continuous time series streams.
\item \textbf{New Datasets and Testbed.} We introduce \dataset, the first data suite supporting both the training and evaluation of CL-TSU models. It covers 18 tasks across six capabilities, with over 5K streams totaling 2.2B timesteps and 15K annotated response events.
\item \textbf{New Model and Empirical Findings.} We present \method, the first native foundation model for CL-TSU, which integrates hybrid time series encoding, stream-persistent hybrid memory, and elastic stream-inference synchronization. Evaluations show that \method outperforms frontier LLMs on the in-distribution test set (90.0 vs. 79.3 for Opus~5).
Notably, \method is highly robust to instruction paraphrasing (a minimal score drop of 1.0, compared to 17.6 for Opus~5). Additionally, we validate its practical efficiency by evaluating deployment performance on a real edge device (RTF 0.358 on NVIDIA Jetson AGX Orin).
\end{enumerate}

\section{Related Work}
\textbf{Large Time Series Models.} Time series research has increasingly shifted from task-specific architectures toward general-purpose pre-trained models. \textbf{Time series foundation models (TSFMs)} use large and diverse corpora for zero-shot forecasting~\citep{jin2024timellm, shi2025timemoe, zhou2026eidoslatentspacepredictivelearning} and, more recently, general-purpose classification~\citep{MOMENT, feofanov2026mantisv2closingzeroshotgap, cauker}. Recognizing that other modalities can provide complementary information, the community subsequently developed \textbf{time series language models (TSLMs)} and multimodal architectures that align temporal data with LLMs for understanding and complex reasoning~\citep{timemqa, timeomni-1}. Despite these impressive capabilities, the underlying modeling paradigm remains fundamentally one-shot.
Although recent efforts such as ARGOS~\citep{agentic_coding_monitoring-1} attempt to address continuous scenarios by utilizing LLM agents to generate executable code, this generation process is fundamentally offline. At runtime, the system relies on fixed code execution, making online user adaptation difficult~\citep{Real-Time-Streams-ad}. Furthermore, such approaches remain confined to local analysis around the current timestep. They do not accumulate historical evidence across sequential inputs and are incapable of handling cross-window comparative reasoning.

\textbf{Streaming Multimodal Models.} To process continuously generated data modalities, such as video and audio, the broader multimodal community has begun moving from static clips and utterances toward continuously operating models. In the vision domain, streaming video models incrementally process incoming frames to support online understanding, event tracking, and proactive assistance~\citep{VideoLLM-online, Always-On-video-Understanding,zhangStreamArenaContinuousInteractive2026,Wang_2026_CVPR_StreamEQA}. In audio, real-time and always-on models continuously perceive acoustic streams and interact without predefined segment boundaries~\citep{Real-Time-Dialogue,AudioInteractionModel,lin2026stepaudio3realtimetechnical}. To support these continuous streams without exhausting GPU memory, various system-level optimizations have emerged. Most notably, approaches like StreamingLLM~\citep{streamingllm} employ KV cache eviction strategies and attention sinks to maintain stable generation in theoretically infinite contexts. However, these optimizations primarily focus on maintaining decoding efficiency and low latency by systematically discarding historical tokens~\citep{ProactiveLLM}. They largely neglect the necessity of a persistent memory mechanism capable of preserving exact historical evidence and long-term task states. Furthermore, these general-purpose multimodal models fundamentally lack a native understanding of time series waveforms. Consequently, maintaining a persistent memory over continuous time series streams remains an unexplored frontier.

\section{Methodology}
\label{sec:problem-formulation}
\textbf{Problem Definition.} We define continuous \& live time series understanding as stateful interaction over an open-ended stream. In conventional one-shot tasks, a model receives a bounded sequence $X_{1:T}\!=\!(x_1,\ldots,x_T)$ and an instruction $q$, produces a single output
$y\!=\!f(X_{1:T},q)$, and then terminates. In contrast, a CL-TSU system processes an open-ended stream $X\!=\!(x_1,x_2,\ldots)$ under an initial instruction $q$, maintains persistent state, and determines whether and what to respond at each new observation.
Formally: $(s_t, a_t) = H(s_{t-1}, x_t, q, u_t)$, where $a_t \in \mathcal{Y} \cup \{\emptyset\}$.
Here, $s_t$ represents the persistent task-relevant state; $x_t$ is the current chunk of observations; $q$ is the standing instruction; $u_t$ is an optional mid-stream query or rule update, with $u_t=\emptyset$ when no update is received. The action $a_t$ is either a response in the response space $\mathcal{Y}$, or $\emptyset$, indicating that the system remains silent at step $t$. Unlike static models, a CL-TSU system is characterized by three properties: \textbf{(1) Open-ended processing}, continuously handling incoming observations without a predefined termination point; \textbf{(2) Stateful historical reasoning}, maintaining and updating the internal state $s_t$ so that later decisions can use information accumulated across earlier windows; and \textbf{(3) Dynamic adaptability}, incorporating asynchronous mid-stream queries or rule changes $u_t$ on the fly.

\subsection{\dataset}
\label{sec:dataset}
\begin{figure}[t]
    \centering
    \includegraphics[width=\linewidth]{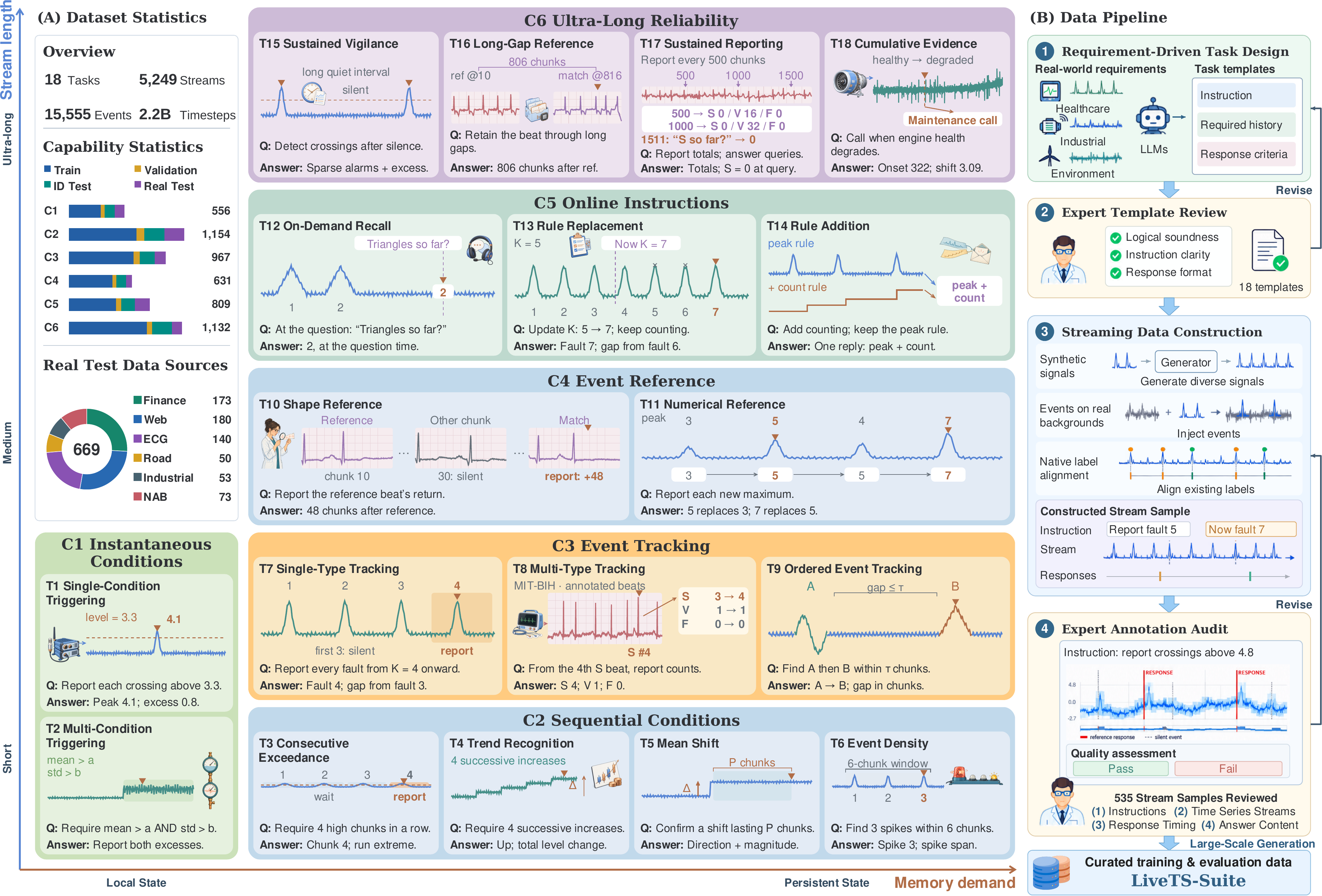}
    \caption{\textbf{Overview of \dataset.} The 18 tasks span six capabilities, organized by stream length (vertical) and memory demand (horizontal). Each task illustrates the instruction, the open-ended stream, and example responses. \textbf{(a)} Dataset statistics summarize stream counts by capability and Real Test data sources (see Appendix~\ref{app:data-statistics} for details). \textbf{(b)} Data pipeline details the four-stage data construction process used to generate large-scale training and evaluation data for CL-TSU.}
    \label{fig:taskmap}
\end{figure}

To address the scarcity of data for CL-TSU, we construct \dataset, the first comprehensive data suite tailored to this setting and designed to support both model training and evaluation. Its construction is guided by the three defining properties introduced in Section~\ref{sec:problem-formulation}.
As shown in Figure~\ref{fig:taskmap}, \dataset contains 18 tasks spanning six core capabilities, organized along two axes. The vertical axis, \textbf{stream length}, dictates the required duration of continuous operation, assessing the capacity for \textit{open-ended processing}. More importantly, distinguishing our paradigm from existing offline tasks, the horizontal axis, \textbf{memory demand}, dictates what historical information must be retained and how it supports current decisions, directly evaluating \textit{stateful historical reasoning}. Starting with conditions in current observations (\textbf{C1: Instantaneous Conditions}), the suite extends to conditions involving recent history (\textbf{C2: Sequential Conditions}), persistent event records (\textbf{C3: Event Tracking}), and historical references for subsequent judgments (\textbf{C4: Event Reference}). It subsequently introduces dynamic queries and rule changes during the stream (\textbf{C5: Online Instructions}) to explicitly test \textit{dynamic adaptability}, culminating in tests of sustained reliability over ultra-long streams (\textbf{C6: Ultra-Long Reliability}). \dataset comprises 5{,}249 streams across the Train, Validation, ID, and Real Test splits, with 15{,}555 annotated events and 2.2B timesteps. Experts reviewed all 18 task templates and audited 535 stream instances via a dedicated interface, including the numerical-shift and instruction-paraphrasing test sets designed to evaluate model generalization (Figure~\ref{fig:generalization}).

\subsubsection{Capabilities and Tasks}
\label{sec:dataset-tasks}
\dataset decomposes CL-TSU into 18 specific tasks to systematically evaluate a model's ability to maintain persistent states over data streams. Dataset statistics are provided in Figure~\ref{fig:taskmap}(a). 
Below, we detail the tasks. \textbf{Instantaneous Conditions (C1).} 
This capability establishes condition recognition within the current chunk, from a single criterion to a conjunction of criteria. \textit{Single-Condition Triggering \textbf{(T1)}} reports a signal crossing a specified threshold, while \textit{Multi-Condition Triggering \textbf{(T2)}} requires multiple properties (e.g., mean and variability) to satisfy joint constraints.

\textbf{Sequential Conditions (C2).} 
Extending local judgments to recent history, C2 covers four conditions involving persistence, direction, mean-level change, and event density. \textit{Consecutive Exceedance \textbf{(T3)}} recognizes a sustained run above a threshold; \textit{Trend Recognition \textbf{(T4)}} identifies successive changes in one direction; \textit{Mean Shift \textbf{(T5)}} recognizes a sustained departure from the preceding mean; and \textit{Event Density \textbf{(T6)}} identifies a concentration of events within a given interval.

\textbf{Event Tracking (C3).} 
Once events are detected, their records must remain available to inform later decisions. \textit{Single-Type Tracking \textbf{(T7)}} counts recurring events of one type and reports their count and inter-event interval when the instructed threshold is reached. \textit{Multi-Type Tracking \textbf{(T8)}} maintains separate counts for multiple event types, such as ECG beat classes, and reports them when the designated class triggers a response. \textit{Ordered Event Tracking \textbf{(T9)}} captures their temporal relationships by identifying a prescribed sequence of event types within an allowed interval.

\textbf{Event Reference (C4).} 
Historical observations also serve as references for interpreting new inputs. C4 distinguishes shape and numerical references: \textit{Shape Reference \textbf{(T10)}} retains an early waveform example to recognize its recurrence, whereas \textit{Numerical Reference \textbf{(T11)}} compares incoming values with the historical extremes and updates the record when a new high or low is reached.

\textbf{Online Instructions (C5).} 
Beyond the initial instruction $q$, C5 introduces mid-stream requests $u_t$. \textit{On-Demand Recall \textbf{(T12)}} responds to asynchronous queries, such as requests for the count of a specified event type; \textit{Rule Replacement \textbf{(T13)}} applies a revised rule to the existing event state; and \textit{Rule Addition \textbf{(T14)}} introduces a new rule while preserving the previously active rule. These tasks cover querying, revising, and extending the requirements governing a continuous stream.

\textbf{Ultra-Long Reliability (C6).} 
C6 extends the capabilities along the stream-length axis to test reliability over prolonged operation. \textit{Sustained Vigilance \textbf{(T15)}} tests responsiveness to sparse events amid long quiet periods; \textit{Long-Gap Reference \textbf{(T16)}} evaluates reference retention across distant recurrences; \textit{Sustained Reporting \textbf{(T17)}} reports cumulative counts at scheduled intervals while answering intervening queries; and \textit{Cumulative Evidence \textbf{(T18)}} accumulates subtle deviations to capture gradual shifts that remain hidden within individual chunks. Unlike the localized mean shift in T5, this trigger relies strictly on long-term accumulated evidence. Collectively, by tightly coupling stream length with memory demand, \dataset operationalizes the \mbox{CL-TSU} formulation, requiring models to maintain and dynamically update states over time.

\subsubsection{Streaming Data Construction and Evaluation}
\label{sec:dataset-pipeline}
\textbf{Streaming Data Pipeline.}
Our data construction comprises four stages (Figure~\ref{fig:taskmap}(b); details in Appendix~\ref{app:data-construction}).
\textbf{(1) Requirement-Driven Task Design.} LLMs survey real-world continuous time series understanding requirements and define task templates specifying instructions, required history, and response criteria. \textbf{(2) Expert Template Review.} Human experts review all 18 task templates for logical soundness, instruction clarity, and response format, revising them as needed. \textbf{(3) Streaming Data Construction.} We construct streams through \textit{(a) synthetic signal generation}, \textit{(b) event injection into real-world backgrounds}, and \textit{(c) native label alignment}.
\textbf{(4) Expert Annotation Audit.} PhD-level annotators use a review interface to audit 535 sampled stream instances, verifying consistency among the instructions, time series stream, response timing, and answer content. Failed cases are revised before the validated pipeline is used to generate training and evaluation data at scale.

\textbf{Streaming Evaluation Protocol.} We assess models along two dimensions: \textbf{response quality} and \textbf{real-time efficiency}. Quality demands temporal precision and exact information, measured by multiplying \textit{Timing F1} and \textit{Content Accuracy} into a unified \textbf{\textit{Joint Score}}. Efficiency tracks operational overhead (including \textit{real-time factor}, \textit{chunk latency}, \textit{input tokens / 1k chunks}) and is evaluated exclusively on locally deployable models. Appendix~\ref{app:streaming-evaluation} details all matching and scoring mechanics.

\subsection{\method}
\label{sec:method}
\begin{figure}[t]
    \centering
    \includegraphics[width=\linewidth]{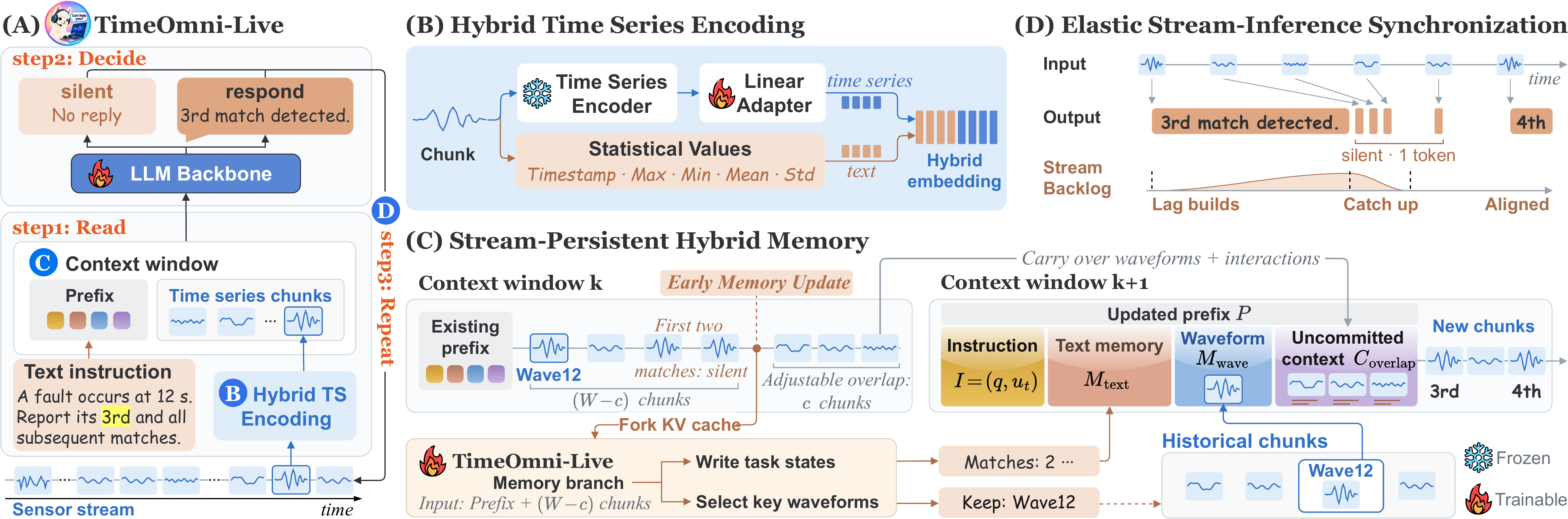}
    \caption{\textbf{\method architecture.} \textbf{(a)} The continuous Read-Decide-Repeat inference loop. \textbf{(b)} Hybrid time series encoding combining continuous waveform embeddings with explicit textual statistics. \textbf{(c)} Stream-persistent hybrid memory retaining both text states and selected waveforms, utilizing early KV-cache forking to prevent stalling. \textbf{(d)} Elastic stream-inference synchronization that buffers chunks during response generation and realigns via fast single-token silence decisions.}
    \label{fig:method}
\end{figure}

\method realizes continuous time series understanding through the three-step \textbf{Read-Decide-Repeat} loop illustrated in Figure~\ref{fig:method}(a). At each iteration, \method reads an incoming time series chunk together with the existing context and then decides whether to remain silent or initiate a response. If a response is triggered, the same backbone generates it while subsequently arriving chunks are buffered; the loop then processes the buffered and newly arriving chunks in order. Although the architecture is backbone-agnostic, we instantiate it with Qwen3.5-4B~\citep{qwen35blog}; Figure~\ref{fig:backbone-analysis} provides a detailed backbone comparison. The following three core mechanisms reliably and efficiently support operation over open-ended streams.

\textbf{Hybrid Time Series Encoding.} As shown in Figure~\ref{fig:method}(b), each incoming time series chunk is processed through two complementary modalities to capture both natively temporal patterns and exact numerical values. A frozen time series encoder~\citep{feofanov2026mantisv2closingzeroshotgap} extracts time series features, which a trainable adapter aligns with the backbone's input dimensions. In parallel, a textual representation provides the chunk's timestamp, maximum, minimum, mean, and standard deviation. The hybrid-modality embedding is then fed into the backbone alongside the existing context.

\textbf{Stream-Persistent Hybrid Memory.} A bounded context cannot retain an entire stream, yet later decisions may depend on earlier events, reference waveforms, or rule updates. As shown in Figure~\ref{fig:method}(c), our hybrid memory preserves two complementary forms of information in the context prefix. The text modality stores compact task states (e.g., event counts and reference thresholds), while abstracting away signal details. The time series modality retains selected historical waveform embeddings, preserving fine-grained shape information for subsequent comparisons. 
For each context window $\textbf{\text{k}}$ containing $W$ new chunks, the memory branch forks the KV cache after the first $W\!-\!c$ chunks, rather than waiting until the window ends. The forking provides a stable context snapshot from which the model asynchronously generates an updated textual state $M_{\text{text}}$ and selected waveform embeddings $M_{\text{wave}}$. Meanwhile, the main branch continues processing the remaining $c$ chunks. The adjustable overlap length $c$ therefore controls how early the memory update begins: a larger $c$ gives slower hardware more time to complete the update, at the cost of carrying more overlap into the next window.
Because the memory branch forks before the window ends, its updated state does not include the remaining $c$ uncommitted chunks and their interactions. We thus carry these chunks forward as an overlap $C_{\text{overlap}}$, containing their raw waveforms and interaction history. The next context window $\textbf{\text{k\!+\!1}}$ begins with the updated prefix $P\!=\![I, M_{\text{text}}, M_{\text{wave}}, C_{\text{overlap}}]$, where $I\!=\!(q,u_t)$ denotes the standing instruction and optional rule update. Newly arriving chunks are then appended to $P$ in order. By preserving the post-fork context explicitly, this asynchronous memory mechanism allows stream processing to continue without waiting for the memory update to finish.

\textbf{Elastic Stream-Inference Synchronization.}
Continuous chunks arrive at fixed intervals, but inference latency varies depending on whether the model is silent or generates a response. As shown in Figure~\ref{fig:method}(d), chunks arriving during response generation are buffered and subsequently processed in order, preventing observation loss. Silence decisions require only one output token, allowing the model to clear the backlog and catch up with the stream when processing is faster than arrival. 

\textbf{Training.} We train only the time series adapter and LoRA parameters of the LLM backbone. \textbf{Training samples and objectives.} To mirror continuous inference, each training window contains the prefix $P$ followed by up to $W$ incoming chunks. We use two forms of supervision: \textit{\textbf{(a)} Response samples} supervise chunk-wise silence or response decisions and reference responses; \textit{\textbf{(b)} Memory samples} supervise the updated text state and indices of historical waveforms to retain. Memory targets are derived deterministically from reference events. Token-level cross-entropy applies only to these targets; input and context tokens are masked from the loss. \textbf{Tailored Training Designs.} We use three training strategies for CL-TSU: \textbf{(1) Silence down-weighting.} To prevent the $>95\%$ of non-response chunks from biasing the model toward permanent silence, we down-weight the loss on silence tokens by a factor of 0.2 (versus 1.0 for other target tokens). \textbf{(2) Memory-prefix perturbation.} To prevent self-written memory errors from compounding across continuous windows, we artificially perturb the memory prefix with detectable defects in $30\%$ of training samples, forcing the model to repair imperfect states rather than blindly trusting the prefix. \textbf{(3) Variable overlap $c$.} To avoid tying the model to a fixed hardware assumption, we dynamically sample the overlap length $c\!\in\!\{10,15,20\}$ per stream during training. 
This enables flexible adjustment of the overlap length $c$ at deployment, giving memory updates more time to finish before the next window and reducing stalls. Table~\ref{tab:efficiency} shows the effect of $c$ on a real edge device. Appendix~\ref{app:training} provides full training details.

\section{Experiments}
\label{sec:experiments}
\textbf{Baselines.}
We compare \method against leading models in two categories: \textbf{(1) General-purpose LLMs}, including proprietary models (Opus~5, Sonnet~5) and five open-source LLMs (DeepSeek-V4.1-Flash, MiniMax-M3, Qwen3.8-27B, Qwen3.5-9B, Qwen3.5-4B); and \textbf{(2) Time series language models}, TimeOmni-1~\citep{timeomni-1} and TS-Reasoner~\citep{yu2025tsreasoner}.

\begin{table}[!b]
\vspace{-5mm}
\caption{Joint Score $S$ on in-distribution (ID) and out-of-distribution Real Testbeds, with results aggregated from 18 tasks into six capabilities. Detailed per-task Timing F1 $F_j$ and Content Accuracy $C_j$ are provided in Tables~\ref{tab:task-metrics-id} and~\ref{tab:task-metrics-real} of Appendix~\ref{app:result-tables}. \textcolor{red}{\textbf{Red}}: the best, \textcolor{blue}{\underline{Blue}}: the 2nd best.}
\label{tab:main}\label{tab:capabilities}
\centering
\begingroup
\newcommand{\mainbest}[1]{\textcolor{red}{\textbf{#1}}}
\newcommand{\mainsecond}[1]{\textcolor{blue}{\underline{#1}}}
\newcommand{\capheader}[2]{\raisebox{\dimexpr.5ex-.5\height\relax}{\shortstack{\footnotesize\textbf{#1}\\[-1pt]\footnotesize\textbf{#2}}}}
\renewcommand{\arraystretch}{1.25}
\setlength{\tabcolsep}{1mm}
\newlength{\mainmetricwidth}
\settowidth{\mainmetricwidth}{\capheader{C1: Instantaneous}{Conditions}}
\setlength{\mainmetricwidth}{\dimexpr\mainmetricwidth/2-\tabcolsep+1pt\relax}
\resizebox{\linewidth}{!}{%
\begin{tabular}{cc*{14}{>{\centering\arraybackslash}p{\mainmetricwidth}}}
\toprule
\multirow{2}{*}[-.6ex]{\textbf{Method}} & \multirow{2}{*}[-.6ex]{\textbf{Base LLMs}} & \multicolumn{2}{c}{\capheader{C1: Instantaneous}{Conditions}} & \multicolumn{2}{c}{\capheader{C2: Sequential}{Conditions}} & \multicolumn{2}{c}{\capheader{C3: Event}{Tracking}} & \multicolumn{2}{c}{\capheader{C4: Event}{Reference}} & \multicolumn{2}{c}{\capheader{C5: Online}{Instructions}} & \multicolumn{2}{c}{\capheader{C6: Ultra-Long}{Reliability}} & \multicolumn{2}{c}{\textbf{Overall}}\\
\cmidrule(lr){3-4}\cmidrule(lr){5-6}\cmidrule(lr){7-8}\cmidrule(lr){9-10}\cmidrule(lr){11-12}\cmidrule(lr){13-14}\cmidrule(lr){15-16}
 & & ID & Real & ID & Real & ID & Real & ID & Real & ID & Real & ID & Real & ID & Real\\
\midrule
\textbf{Proprietary Models} &  &  &  &  &  &  &  &  &  &  &  &  &  &  & \\
Opus 5 &  & 98.0 & \mainsecond{98.0} & \mainbest{96.1} & \mainbest{95.0} & \mainsecond{86.6} & \mainbest{71.4} & 74.4 & \mainbest{62.1} & 49.0 & \mainbest{56.7} & \mainsecond{72.8} & 28.3 & \mainsecond{79.3} & \mainbest{66.5}\\
Sonnet 5 &  & \mainbest{100.0} & \mainbest{100.0} & 88.3 & \mainsecond{90.6} & 39.4 & 33.9 & \mainsecond{75.6} & 41.4 & \mainsecond{50.7} & 50.4 & 71.5 & 35.3 & 70.0 & \mainsecond{57.7}\\
\textbf{Open-Source Models} &  &  &  &  &  &  &  &  &  &  &  &  &  &  & \\
DeepSeek-V4.1-Flash &  & \mainbest{100.0} & \mainbest{100.0} & 87.8 & 89.0 & 64.9 & \mainsecond{62.1} & 0.0 & 4.5 & 0.0 & 0.0 & 62.7 & \mainbest{41.5} & 55.4 & 50.9\\
MiniMax-M3 &  & \mainsecond{99.9} & \mainbest{100.0} & 82.7 & 87.3 & 13.6 & 16.8 & 22.0 & 0.0 & 47.3 & \mainsecond{50.6} & 66.6 & 25.9 & 56.9 & 47.5\\
Qwen3.8-27B &  & 54.2 & 46.0 & 56.4 & 58.4 & 34.9 & 33.2 & 17.8 & 25.7 & 38.1 & 41.4 & 50.3 & 27.3 & 43.9 & 39.4\\
Qwen3.5-9B &  & 53.4 & 49.4 & 8.2 & 8.4 & 0.0 & 0.0 & 0.0 & 0.0 & 11.6 & 17.8 & 0.0 & 2.6 & 9.7 & 10.9\\
Qwen3.5-4B &  & 9.3 & 7.8 & 6.7 & 7.1 & 0.0 & 0.0 & 0.0 & 1.4 & 0.1 & 0.1 & 0.0 & 0.0 & 2.5 & 2.6\\
Qwen3.5-4B (streaming) &  & 2.1 & 0.7 & 0.0 & 0.8 & 0.0 & 0.4 & 0.0 & 0.0 & 0.0 & 0.3 & 0.0 & 0.0 & 0.2 & 0.4\\
\midrule
\textbf{Time Series Language Models} &  &  &  &  &  &  &  &  &  &  &  &  &  &  & \\
TimeOmni-1-4B & Qwen3.5-4B & 0.0 & 0.0 & 0.0 & 0.0 & 0.0 & 0.0 & 0.0 & 0.0 & 4.3 & 5.4 & 0.0 & 0.0 & 0.7 & 0.9\\
TS-Reasoner-7B (streaming) & Qwen2.5-7B & 1.0 & 0.8 & 1.1 & 1.5 & 0.1 & 0.1 & 0.0 & 0.1 & 0.5 & 0.4 & 0.0 & 0.0 & 0.5 & 0.5\\
\midrule
\textbf{Ours} &  &  &  &  &  &  &  &  &  &  &  &  &  &  & \\
\rowcolor{blue!5}
\method & Qwen3.5-4B & 87.6 & 83.9 & \mainsecond{92.6} & 80.4 & \mainbest{95.2} & 41.9 & \mainbest{93.8} & \mainsecond{54.7} & \mainbest{84.2} & 43.7 & \mainbest{86.9} & \mainsecond{38.3} & \mainbest{90.0} & 56.1\\
\bottomrule
\end{tabular}}
\endgroup
\end{table}

\textbf{Metrics.} For \textbf{response quality}, we report the Joint Score $S\!=\!\frac{100}{|\mathcal{J}|}\sum_{j\in\mathcal{J}}F_j C_j$ to capture both response timing and content accuracy over the evaluated task subset  $\mathcal{J}$. For each task $j\!\in\!\mathcal{J}$, $F_j$ is the mean per-stream timing F1, and $C_j$ is the mean per-stream content accuracy over streams with matched responses ($C_j=0$ if no responses match). 
For \textbf{real-time efficiency}, we evaluate locally deployable models using three metrics: (1) Processed input tokens per 1{,}000 chunks; (2) Real-time factor (RTF), the active processing time divided by stream replay duration; and (3) p95 chunk latency, the wall-clock delay from data arrival to completed processing, including queueing. Appendix~\ref{app:streaming-evaluation} details these metrics, and Appendix~\ref{app:case-studies} illustrates the task-specific scoring criteria.

\subsection{Main Results}
\label{sec:main-results}
\textbf{\underline{Setup.}} To ensure a fair comparison in Table~\ref{tab:main}, we evaluate the baselines under two specific settings:
\textbf{(1) Code execution.} Unless explicitly marked as streaming, all baseline models generate executable programs that run in a sandbox. We adopt this protocol for two reasons: \textit{\textbf{(a) serializing numerical chunks as text is costly}} (averaging 2.56M \textit{tokens/stream} with the Qwen3.5 tokenizer), and \textit{\textbf{(b) direct numerical prompting yields far weaker results}} (Qwen3.5-4B scores only 0.2/0.4 on ID/Real Test in streaming mode, versus 2.5/2.6 when writing code).
\textbf{(2) Feedback-Driven Curriculum.} Because baseline performance improves with execution feedback and additional examples, a conventional zero-shot evaluation may underestimate these models. We therefore allow baselines to iteratively revise their programs using execution feedback. Figure~\ref{fig:codegen-curriculum} shows the resulting gains, which vary across models. All code baselines follow the same schedule, and we report their final converged results. 
\textbf{\underline{Results.}} \textbf{Strong CL-TSU performance.} With only 4B parameters and purely streaming inference, \method achieves an ID Test score of 90.0, surpassing the strongest code baseline (Opus 5) by \textbf{10.7} points. On the entirely unseen, out-of-distribution (OOD) Real Test, it achieves 56.1, clearly outperforming DeepSeek-V4.1-Flash (50.9) while remaining substantially stronger than other streaming baselines such as Qwen3.5-4B~(streaming) and TS-Reasoner-7B. \textbf{Robust instruction following.} Paraphrasing reduces its score by only 1.0 point (89.5 to 88.5; Figure~\ref{fig:generalization}(a)), demonstrating the robustness to variations in instruction wording.

\subsection{More Analysis}
\label{sec:more-analysis}

\textbf{Ablation Studies.}
\textbf{\underline{Setup.}}
Table~\ref{tab:ablation} ablates two core components. We uniformly sample five complete streams per task per split (180 streams in total, 11.2\% of the full ID/Real evaluation set), using the same samples for all variants. \textit{\textbf{(a) Hybrid Time Series Encoding:}} Variants are trained under a strictly fixed 2{,}500-step budget to isolate modality contributions without confounding by training length. \textit{\textbf{(b) Stream-Persistent Hybrid Memory:}} Interventions are applied directly to the main checkpoint without retraining.
\textbf{\underline{Results.}}
\textbf{Complementary modalities.} The hybrid encoding of both time series and textual modalities achieves the highest overall scores. The closeness of the statistics-only variant is expected: Some tasks can be effectively solved from numerical summaries alone, whereas waveform embeddings remain essential for shape-sensitive tasks in C4.
\textbf{Memory ablations validate the capability organization.} The full memory mechanism consistently outperforms its ablated variants. For C1--C2, the required evidence remains within the backbone's local context window, limiting the need for cross-window memory, so all variants perform similarly. The benefits become pronounced for C3--C6: removing text memory reduces the C3 score from 95.6 to 30.5. Removing historical waveforms has only a small effect on overall performance because relatively few tasks are shape-sensitive, but the substantial decrease on C4 confirms their importance for cross-window waveform comparisons. These results support the suite's capability organization, where memory demand increases from local condition recognition (C1--C2) to persistent state tracking and historical references (C3--C6). Detailed scores are provided in Appendix~\ref{app:ablation-details}.

\begin{table}[b]
\vspace{-5mm}
\caption{Encoding and memory ablations. Encoding variants are each trained separately for exactly 2,500 steps; memory variants use the main checkpoint without retraining. \textbf{Bold}: the best.}
\label{tab:ablation}\label{tab:perception}\label{tab:memory}
\vspace{1mm}
\centering
\begingroup
\newcommand{\capheader}[2]{\raisebox{\dimexpr.5ex-.5\height\relax}{\shortstack{\footnotesize\textbf{#1}\\[-1pt]\footnotesize\textbf{#2}}}}
\renewcommand{\arraystretch}{1.25}
\setlength{\tabcolsep}{1mm}
\newlength{\ablationmetricwidth}
\settowidth{\ablationmetricwidth}{\capheader{C1: Instantaneous}{Conditions}}
\setlength{\ablationmetricwidth}{\dimexpr\ablationmetricwidth/2-\tabcolsep+1pt\relax}
\resizebox{\linewidth}{!}{%
\begin{tabular}{c*{14}{>{\centering\arraybackslash}p{\ablationmetricwidth}}}
\toprule
\multirow{2}{*}[-.6ex]{\textbf{Variant}} & \multicolumn{2}{c}{\capheader{C1: Instantaneous}{Conditions}} & \multicolumn{2}{c}{\capheader{C2: Sequential}{Conditions}} & \multicolumn{2}{c}{\capheader{C3: Event}{Tracking}} & \multicolumn{2}{c}{\capheader{C4: Event}{Reference}} & \multicolumn{2}{c}{\capheader{C5: Online}{Instructions}} & \multicolumn{2}{c}{\capheader{C6: Ultra-Long}{Reliability}} & \multicolumn{2}{c}{\textbf{Overall}}\\
\cmidrule(lr){2-3}\cmidrule(lr){4-5}\cmidrule(lr){6-7}\cmidrule(lr){8-9}\cmidrule(lr){10-11}\cmidrule(lr){12-13}\cmidrule(lr){14-15}
 & ID & Real & ID & Real & ID & Real & ID & Real & ID & Real & ID & Real & ID & Real\\
\midrule
\textbf{(a) Hybrid Time Series Encoding} &  &  &  &  &  &  &  &  &  &  &  &  &  & \\
\rowcolor{blue!5}
\method (trained for 2,500 steps) & 83.1 & \textbf{84.9} & 89.3 & \textbf{77.2} & 97.9 & \textbf{45.3} & \textbf{93.8} & 53.9 & 76.8 & 54.8 & \textbf{76.3} & 43.9 & \textbf{85.6} & \textbf{59.0}\\
Time Series Embeddings Only & 11.7 & 15.0 & 76.5 & 20.5 & \textbf{98.7} & 19.2 & 72.3 & 25.0 & 74.1 & 21.6 & 48.9 & 13.2 & 66.0 & 18.7\\
Statistical Values Only & \textbf{85.6} & 74.9 & \textbf{95.5} & 73.4 & 84.5 & 38.8 & 76.0 & \textbf{56.8} & \textbf{83.3} & \textbf{59.4} & 65.1 & \textbf{47.1} & 81.6 & 57.8\\
\midrule
\textbf{(b) Stream-Persistent Hybrid Memory} &  &  &  &  &  &  &  &  &  &  &  &  &  & \\
\rowcolor{blue!5}
\method & \textbf{88.1} & \textbf{84.9} & 93.2 & \textbf{77.1} & \textbf{95.6} & \textbf{45.0} & \textbf{93.8} & \textbf{52.2} & \textbf{79.4} & \textbf{52.8} & \textbf{88.7} & 39.5 & \textbf{89.8} & \textbf{57.4}\\
w/o Text Memory & \textbf{88.1} & \textbf{84.9} & 93.2 & \textbf{77.1} & 30.5 & 27.8 & 56.0 & 28.6 & 32.8 & 24.8 & 53.1 & 37.6 & 59.1 & 46.9\\
w/o Waveform Memory & \textbf{88.1} & \textbf{84.9} & \textbf{93.4} & \textbf{77.1} & \textbf{95.6} & \textbf{45.0} & 65.7 & 48.1 & \textbf{79.4} & \textbf{52.8} & 68.9 & 40.2 & 82.3 & 57.1\\
w/o Uncommitted Context & \textbf{88.1} & \textbf{84.9} & 93.2 & \textbf{77.1} & 54.6 & 25.4 & 74.5 & 48.0 & 48.5 & 26.1 & 71.5 & \textbf{40.7} & 71.9 & 49.5\\
Memory Delayed by One Window & \textbf{88.1} & \textbf{84.9} & 93.2 & \textbf{77.1} & 42.6 & 35.9 & 68.6 & 35.2 & 39.7 & 32.6 & 58.1 & 38.4 & 64.8 & 50.4\\
\bottomrule
\end{tabular}}
\endgroup
\end{table}

\begin{wrapfigure}{r}{0.27\textwidth}
\vspace{-4mm}
\raggedleft
\includegraphics[width=\linewidth]{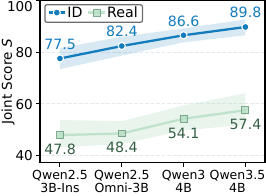}
\vspace{-6mm}
\caption{\small Backbone scores on the 180-stream test subset, with 95\% confidence bands.}
\label{fig:backbone-analysis}
\vspace{-3mm}
\end{wrapfigure}

\textbf{Backbone Analysis.} \textbf{\underline{Setup.}} We compare four language backbones of comparable scale (3B--4B) under identical training settings, evaluated on the same 180-stream subset described above.
\textbf{\underline{Results.}} \textbf{Stronger LLMs yield stronger CL-TSU.} Figure~\ref{fig:backbone-analysis} shows that newer backbones consistently improve overall scores on both ID and Real Test, suggesting that our framework can readily benefit from future advances in language backbones.
\textbf{Lightweight design enables low-cost upgrades.} This trend highlights a key advantage of our 4B-parameter architecture: it allows the framework to be easily and cost-effectively upgraded as new open-source models emerge. Detailed capability scores and training convergence curves are provided in Appendix~\ref{app:backbone}.

\begin{figure}[!htbp]
\centering
\subfloat[Instruction paraphrasing (IP)\label{fig:generalization-ip}]{%
\includegraphics[width=0.34912\linewidth]{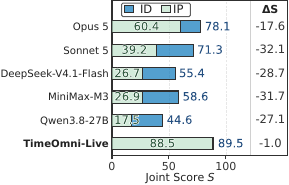}}\hfill
\subfloat[Numerical shift (NS)\label{fig:generalization-ns}]{%
\includegraphics[width=0.34912\linewidth]{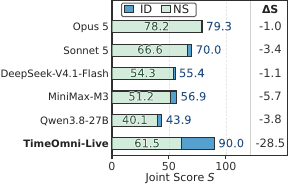}}\hfill
\subfloat[Third-party evaluation\label{fig:generalization-kpi}]{%
\includegraphics[width=0.29296\linewidth]{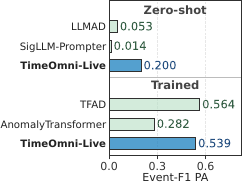}}
\caption{Generalization evaluations. (a) Instruction paraphrasing on 180 paired ID test streams. (b) Numerical shift on the entire ID and NS test sets, where $\Delta\!S$ is the Joint Score change. (c) Third-party evaluation on KPI using mean per-series Event-F1 PA. Details are provided in Appendix~\ref{app:generalization-details}.}
\label{fig:generalization}
\end{figure}

\textbf{Generalization.}
\textbf{\underline{Setup.}}
We evaluate three forms of generalization:
\textit{\textbf{(a) Instruction paraphrasing}}, which rewrites instruction wording while preserving task requirements;
\textit{\textbf{(b) Numerical shifts}}, which introduce new numerical ranges and waveform families; and
\textit{\textbf{(c) Third-party evaluation}} on KPI~\citep{KPI-benchmark}, a real-world anomaly detection dataset with manually labeled incidents.
\textbf{\underline{Results.}}
Figure~\ref{fig:generalization} summarizes the outcomes.
\textbf{Strong instruction generalization.} \method exhibits exceptional robustness to paraphrased instructions compared to brittle code execution baselines, suffering a minimal performance drop ($\Delta S = -1.0$).
\textbf{Competitive numerical generalization.} Under numerical shifts, \method achieves a Joint Score of 61.5 and outperforms DeepSeek-V4.1-Flash (54.3). This larger $\Delta S$ is expected: native language models like \method rely on learned parametric priors for numerical reasoning, making them less robust to out-of-distribution numerical ranges than deterministic sandbox code execution.
\textbf{Effective third-party transfer.}
On KPI, \method surpasses zero-shot LLM baselines (Event-F1 PA 0.200 vs. 0.053 for LLMAD~\citep{liu2025large}) and achieves competitive performance against specialized deep learning methods when adapted (0.539 vs. 0.282 for AnomalyTransformer~\citep{xu2022anomaly}).

\begin{wrapfigure}{r}{0.315\textwidth}
\centering
\vspace{-4mm}
\includegraphics[width=\linewidth,height=0.256676\textwidth]{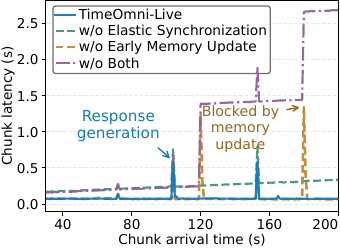}
\setlength{\abovecaptionskip}{3pt}
\vspace{-4mm}
\caption{Scheduling ablations.}
\vspace{-4mm}
\label{fig:runtime-scheduling}
\end{wrapfigure}
\textbf{Elastic Scheduling.}
\textbf{\underline{Setup.}} We test \textbf{\textit{(a) elastic stream-inference synchronization}} and \textbf{\textit{(b) early memory update}}. Figure~\ref{fig:runtime-scheduling} ablates \textit{w/o Elastic Synchronization} (fixed-rate processing), \textit{w/o Early Memory Update} (window-end updates), and \textit{w/o Both}. All replay the same complete stream at 1 chunk/s.
\textbf{\underline{Results.}} \textbf{Elasticity ensures rapid recovery.} \method maintains low latency and quickly recovers after response peaks; without it, delay gradually accumulates. \textbf{Early memory update prevents boundary stalls.} Without it, window boundaries cause temporary stalls, and removing both mechanisms produces severe increases in delay. Full-stream statistics and backlog curves appear in Appendix~\ref{app:efficiency-details}.

\begin{wraptable}{r}{0.47\textwidth}
\centering
\vspace{-7mm}
\setlength{\belowcaptionskip}{2pt}\caption[Efficiency and edge deployment.]{Efficiency and edge deployment.\\Details in Appendix~\ref{app:efficiency-details}.}
\label{tab:runtime}\label{tab:efficiency}\label{tab:jetson}
\fontsize{5.4}{6.6}\selectfont
\setlength{\tabcolsep}{1.5pt}
\renewcommand{\arraystretch}{1.15}
\newlength{\effcontentwidth}
\setlength{\effcontentwidth}{\dimexpr\linewidth-8\tabcolsep-4pt\relax}
\newcommand{\effratio}[2]{\textcolor{#1}{(#2\texttimes)}}
\newsavebox{\efftablebox}
\begin{lrbox}{\efftablebox}
\begin{tabular}{@{\hspace{2pt}}
  >{\raggedright\arraybackslash}m{0.38\effcontentwidth}
  >{\centering\arraybackslash}m{0.085\effcontentwidth}
  >{\centering\arraybackslash}m{0.18\effcontentwidth}
  >{\centering\arraybackslash}m{0.135\effcontentwidth}
  >{\centering\arraybackslash}m{0.22\effcontentwidth}
  @{\hspace{2pt}}}
\toprule
\textbf{Model} & $\boldsymbol{S}\uparrow$ &
\shortstack{\textbf{Input tokens}\\\textbf{/ 1k chunks}$\downarrow$} &
\textbf{RTF}$\downarrow$ &
\shortstack{\textbf{p95 latency}\\\textbf{(s)}$\downarrow$}\\
\midrule
\multicolumn{5}{@{\hspace{2pt}}l@{\hspace{2pt}}}{\textbf{(a) Streaming efficiency on H100}}\\
Qwen3.5-4B (streaming) & 2.0 & 8,558,122 & 0.749 & 16.615\,\effratio{red}{206}\\
TS-Reasoner-7B (streaming) & 0.3 & 594,623 & 0.276 & 0.507\,\effratio{red}{6.3}\\
\method ($c=15$) & \textbf{53.8} & \textbf{56,541} & \textbf{0.117} & \textbf{0.081}\,\effratio{black!55}{1}\\
\midrule
\multicolumn{5}{@{\hspace{2pt}}l@{\hspace{2pt}}}{\textbf{(b) Jetson AGX Orin}}\\
\method ($c=10$) & 53.7 & 53,164 & 0.375 & 2.246\,\effratio{red}{3.8}\\
\method ($c=15$) & 54.3 & 56,511 & 0.358 & 0.649\,\effratio{red}{1.1}\\
\method ($c=20$) & 55.3 & 59,890 & 0.360 & 0.592\,\effratio{black!55}{1}\\
\bottomrule
\end{tabular}
\end{lrbox}
\ifdim\wd\efftablebox>\dimexpr\linewidth+0.1pt\relax
  \PackageError{efficiency-table}{Table 3 exceeds its allocated width}{Check column widths and padding.}
\fi
\usebox{\efftablebox}
\vspace{-3mm}
\end{wraptable}

\textbf{Efficiency.}
\textbf{\underline{Setup.}} We deploy three streaming models on dedicated H100 GPUs, feeding time series at 1 chunk/s. We evaluate them on 54 uniformly sampled Real Test streams using the three real-time efficiency metrics, where RTF below 1 indicates sufficient average processing speed.
\textbf{\underline{Results.}} \textbf{Efficient streaming inference.} Compared with the Qwen3.5-4B streaming baseline, \method reduces processed input tokens by 99.3\% and p95 latency by a factor of 206, while achieving a higher Joint Score (Table~\ref{tab:runtime}).

\textbf{Edge Deployment.}
\textbf{\underline{Setup.}} We deploy the \method checkpoint on a Jetson AGX Orin edge device without quantization, replaying the identical 54 streams used above.
\textbf{\underline{Results.}} \textbf{Real-time edge capability.} At $c=15$, the Jetson achieves a comparable $S$ to the H100 while maintaining an RTF of 0.358 (well below 1), achieving sub-second real-time processing (Table~\ref{tab:runtime}). \textbf{Effectiveness of variable overlap.} Increasing the overlap $c$ from 10 to 20 provides more buffer time for memory updates, which reduces p95 latency by $3.8\times$. Ultimately, $c=20$ yields both the highest score and the lowest latency on the Jetson. Appendix~\ref{app:deployment-details} reports deployment details and overlap sensitivity.

\section{Conclusion}
The work introduces continuous \& live time series understanding (CL-TSU), a paradigm that moves time series understanding beyond offline, one-shot analysis toward persistent reasoning over open-ended streams. We present \textbf{\dataset}, the first data suite for CL-TSU, with 18 streaming tasks spanning six core capabilities and progressively increasing demands on memory and continuous operation. We then develop \textbf{\method}, the first foundation model for CL-TSU, combining hybrid time series encoding, stream-persistent hybrid memory, and elastic stream-inference synchronization. Experiments demonstrate that \method, with only 4B parameters, outperforms frontier models and generalizes to paraphrased instructions, shifted numerical distributions, and third-party data. Ablations further validate the complementary roles of its encoding, memory, and scheduling components. Finally, deployment on a real-world edge device demonstrates its practical efficiency for live, stateful interaction.

\bibliography{reference}
\bibliographystyle{reference}
\clearpage
\appendix

\section{Data Suite Construction and Evaluation}

\subsection{Four-Stage Data Construction}\label{app:data-construction}
Figure~\ref{fig:taskmap}(b) illustrates the four stages of data construction for \dataset.

\textbf{(1) Requirement-Driven Task Design.}
LLMs survey real-world requirements for continuous time series understanding and organize them into task templates. Each template specifies the standing instruction, the historical information needed for subsequent judgments, and when and what to report. These requirements guide the selection of data sources and the temporal structure of each instance.

\textbf{(2) Expert Template Review.}
Human experts inspect the templates for all 18 tasks before stream construction. They check whether the instructions are clear, the required historical information supports the intended task, and the response conditions and answer fields are consistent. Templates that fail these checks are revised before being used to construct instances.

\textbf{(3) Streaming Data Construction.}
Approved templates are instantiated through three complementary routes.
\textit{(a) Synthetic generation} combines white-noise, AR(1), seasonal, or drift backgrounds with task-specific pulses, level or variance changes, trends, gradual drifts, and event sequences, including distractors that do not satisfy the instruction.
\textit{(b) Background-based construction} uses MIT-BIH Arrhythmia signals~\citep{source_mitdb}, NAB series~\citep{source_nab}, and Yahoo Finance daily closing prices.
Segments are cropped or linearly interpolated to the required length, median-centered, normalized by their standard deviation, and randomly scaled before task events are introduced.
\textit{(c) Native-label alignment} uses original beat annotations from MIT-BIH Arrhythmia~\citep{source_mitdb} and MIT-BIH Long-Term ECG~\citep{source_physionet}, and health-state labels from N-CMAPSS~\citep{source_ncmapss}, an engine simulation dataset driven by real flight conditions.
These annotations are aligned to chunks and used to derive task-specific reference responses.
Reference event times and answer fields are derived from these construction records, the signals or source annotations, together with the task rules, and checked programmatically. The resulting instance contains a standing instruction, a sequence of 512-sample chunks, optional timed queries or rule updates, and timestamped reference responses. Event occurrence alone does not necessarily require a response: event order, historical gaps and the active instruction determine the response requirements. For example, T13 retains the cumulative event count across rule updates and applies the new rule to subsequent events. T12 answers later queries using past observations.

\begin{figure}[!ht]
\centering
\fcolorbox{black!55}{white}{\includegraphics[width=0.95\linewidth]{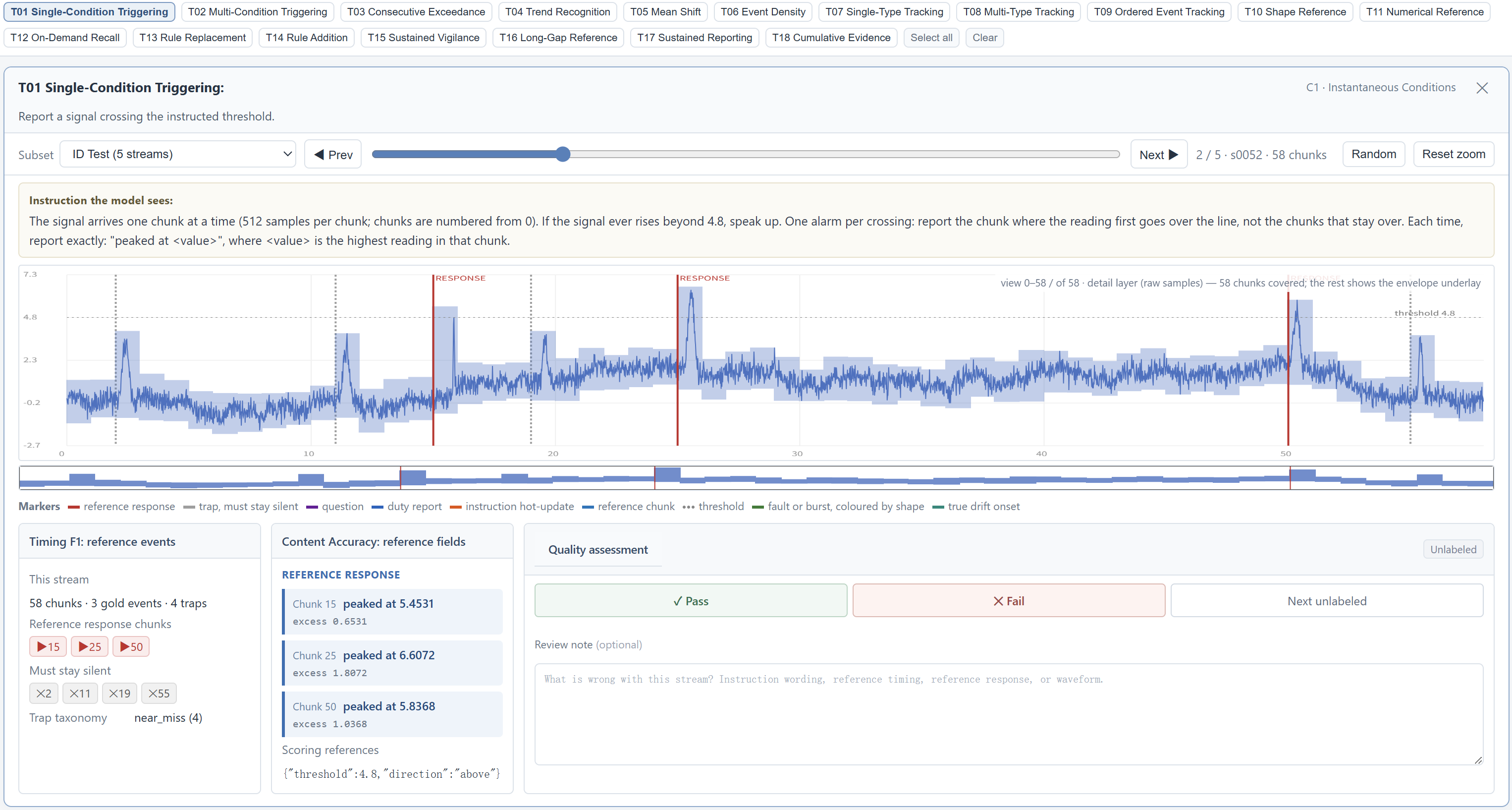}}
\caption{Screenshot of the annotation review interface used in Stage~4 (Expert Annotation Audit).}
\label{fig:annotation-review-interface}
\end{figure}

\textbf{(4) Expert Annotation Audit.}
PhD-level human annotators reviewed 535 stream instances across all 18 tasks: 90 from Train and Validation combined, 90 from ID Test, 85 from Real Test, 90 from Numerical Shift, and 180 from Instruction Paraphrasing. Thus, 265 reviewed instances belong to the main suite splits and 270 to the two generalization evaluations in Figure~\ref{fig:generalization}. The review interface (Figure~\ref{fig:annotation-review-interface}) displays the exact instruction, an overview and zoomable waveform, reference-event and non-response markers, timed queries or updates, and answer fields. Experts inspect whether these elements agree; identified problems lead to revisions of the affected instructions, signal construction or label derivation. The interface supports annotation quality assurance, not model scoring.

\subsection{Data Statistics}
\label{app:data-statistics}
Table~\ref{tab:dataset-statistics} breaks down the 5{,}249 complete streams and 15{,}555 annotated events in \dataset by task and split, totaling 2.2B timesteps. The same table reports maximum and average stream lengths in timesteps, rounded to integers.
\begin{table}[!ht]
\centering
\caption{Stream counts and lengths in \dataset. MAX and AVG pool the four splits; AVG is rounded to the nearest timestep. Events counts annotated response occasions (alarms, scheduled reports, and query answers); each response counts once regardless of its fields. $^{\dagger}$For T2, controlled generation provides precise joint-condition labels for 50 held-out synthetic Real Test streams with independently sampled condition combinations and thresholds.}
\begin{tabular}{lcccccccc}
\toprule
\multirow{2}{*}{\textbf{Task}} & \multirow{2}{*}{\textbf{Train}} & \multirow{2}{*}{\textbf{Validation}} & \multirow{2}{*}{\textbf{ID Test}} & \multirow{2}{*}{\textbf{Real Test}} & \multirow{2}{*}{\textbf{Total}} & \multirow{2}{*}{\textbf{Events}} & \multicolumn{2}{c}{\textbf{Length (timesteps)}}\\
\cmidrule(lr){8-9}
 & & & & & & & \textbf{MAX} & \textbf{AVG}\\
\midrule
T1 & 164 & 20 & 50 & 50 & 284 & 780 & 36,864 & 30,325\\
T2 & 153 & 19 & 50 & 50$^{\dagger}$ & 272 & 674 & 36,864 & 30,850\\
\addlinespace[2pt]
T3 & 171 & 20 & 50 & 50 & 291 & 589 & 36,864 & 30,764\\
T4 & 167 & 20 & 50 & 50 & 287 & 608 & 36,864 & 30,647\\
T5 & 174 & 19 & 50 & 50 & 293 & 377 & 36,864 & 30,362\\
T6 & 164 & 19 & 50 & 50 & 283 & 385 & 36,864 & 30,508\\
\addlinespace[2pt]
T7 & 160 & 20 & 50 & 50 & 280 & 670 & 409,600 & 330,373\\
T8 & 321 & 20 & 50 & 10 & 401 & 1,697 & 409,600 & 236,614\\
T9 & 166 & 20 & 50 & 50 & 286 & 580 & 409,088 & 337,557\\
\addlinespace[2pt]
T10 & 280 & 20 & 50 & 9 & 359 & 1,464 & 409,600 & 257,038\\
T11 & 153 & 19 & 50 & 50 & 272 & 675 & 409,088 & 330,671\\
\addlinespace[2pt]
T12 & 176 & 20 & 50 & 50 & 296 & 895 & 409,600 & 332,098\\
T13 & 194 & 20 & 50 & 50 & 314 & 1,798 & 409,088 & 329,368\\
T14 & 100 & 11 & 39 & 49 & 199 & 1,175 & 409,600 & 335,538\\
\addlinespace[2pt]
T15 & 100 & 10 & 50 & 50 & 210 & 423 & 1,531,392 & 1,288,680\\
T16 & 208 & 10 & 50 & 11 & 279 & 1,060 & 2,149,376 & 1,525,665\\
T17 & 159 & 12 & 50 & 14 & 235 & 1,310 & 1,529,344 & 1,148,771\\
T18 & 312 & 20 & 50 & 26 & 408 & 395 & 1,536,000 & 1,091,363\\
\midrule
\textbf{Total} & \textbf{3,322} & \textbf{319} & \textbf{889} & \textbf{719} & \textbf{5,249} & \textbf{15,555} & \textbf{2,149,376} & \textbf{418,831}\\
\bottomrule
\end{tabular}

\label{tab:dataset-statistics}\label{tab:stream-lengths}
\end{table}

\clearpage
\subsection{Streaming Evaluation Protocol}
\label{app:streaming-evaluation}
We assess models along two dimensions: \textbf{response quality} and \textbf{real-time efficiency}. Streams and instruction updates are presented in temporal order, and evaluation continues after each response.

\subsubsection{Response Quality}
\textbf{Timing F1.}
For each reference event at chunk $a$, we match the earliest unused response within $[a,a+2]$ for T1--T17. T18 requires accumulating small deviations over a long history, similar to numerical integration. To tolerate errors in this accumulated estimate, we use the symmetric window $[a-60,a+60]$. Matching is one-to-one: unmatched responses, including duplicates, are false positives (FP), and unmatched reference events are false negatives (FN). With TP denoting matched responses, each stream $i$ receives
\begin{equation}
F_i=\frac{2\mathrm{TP}_i}{2\mathrm{TP}_i+\mathrm{FP}_i+\mathrm{FN}_i},
\end{equation}
with zero assigned when the denominator is zero.

\textbf{Content Accuracy.}
We compare the requested fields of each matched response with their references. For numerical fields, relative error is $e=|\hat y-y|/\max(|y|,d_{\min})$, where $d_{\min}$ is a field-specific floor. The score of an individual numerical field is
\begin{equation}
c(e)=
\begin{cases}
1, & e\leq 0.05,\\
\dfrac{0.15-e}{0.10}, & 0.05<e<0.15,\\
0, & e\geq 0.15.
\end{cases}
\end{equation}
T18 onset positions instead use absolute errors of 30 and 90 chunks as the full-credit and zero-credit boundaries. Discrete fields require exact agreement after normalization; missing or unparseable fields score zero. Let $\bar c_{ik}$ denote the mean field score of matched response $k$ in stream $i$. With $m_i$ matched responses, its stream-level Content Accuracy is
\begin{equation}
C_i=\frac{1}{m_i}\sum_{k=1}^{m_i}\bar c_{ik},\qquad m_i>0.
\end{equation}
Task-specific fields are listed in Appendix~\ref{app:case-studies}.

\textbf{Joint Score and aggregation.}
For task $j$, $F_j$ averages Timing F1 over all evaluated streams. Let $\mathcal I_j^+$ contain its streams with matched responses. Task-level Content Accuracy $C_j$ and Joint Score $S$ are
\begin{equation}
C_j=\frac{1}{|\mathcal I_j^+|}\sum_{i\in\mathcal I_j^+}C_i,\qquad
S=\frac{100}{|\mathcal J|}\sum_{j\in\mathcal J}F_jC_j.
\end{equation}
We set $C_j=0$ when $\mathcal I_j^+$ is empty.
Capability scores average their constituent tasks; Overall averages all 18 tasks equally. Multiplication precedes averaging across tasks, and rounding is applied only to the displayed results.

\subsubsection{Real-Time Efficiency}
For locally deployable models, we report the three metrics under the same stream replay rate.

\textbf{Input tokens / 1k chunks.}
The total input positions processed by the language model, including text tokens and time series embeddings, divided by the number of chunks and multiplied by 1,000. Reprocessed prefixes count each time; generated output tokens are excluded.

\textbf{Real-time factor (RTF).}
Total active processing time divided by stream replay duration. Processing includes encoding, response generation, and memory updates, but excludes waiting for arrivals. RTF below 1 indicates sufficient average processing speed.

\textbf{p95 chunk latency.}
The 95th percentile of the delay from chunk arrival to completed processing, including queueing. We pool chunk latencies across the evaluated streams before computing the percentile; lower values indicate faster responses to incoming data.

\clearpage

\section{Implementation Details}
\label{app:training}
\subsection{Training Windows and Supervision}
For short tasks (T1--T6), each training window contains the whole stream. For longer streams, windows follow the inference grid with $W=60$ new chunks each, except for a shorter final window. Candidates cover responses, instruction updates, and annotated signal events; each neighboring window is added with probability $0.35$. A uniformly sampled window is added with probability $0.25$ (always for T15 and T18). We retain up to six candidates, then additionally include the first and second windows with probabilities $0.30$ and $0.5$, respectively; the first window is always included for T18. This sampling covers different response positions and both empty memories and memories containing earlier events. We sample the 18 tasks in proportion to the square root of their sample counts, with a minimum sampling weight of $0.6\times$ the mean task weight before normalization.

For each supervised token, cross-entropy receives weight 0.2 for a silence token and 1 for other target tokens; the loss is normalized by the sum of these weights. Context tokens do not contribute to the loss. Memory-prefix perturbation is attempted with probability 0.3 for nonempty memory. Perturbations include a tally inconsistent with the retained event list, shuffled lines, stale reporting positions, or altered whitespace. Historical events are neither inserted nor deleted, and protected memory fields remain unchanged. These perturbations preserve the information needed to recover the reference state. The overlap length is sampled from $\{10,15,20\}$ for each stream during training; evaluation uses $c=15$ with $W=60$ new chunks per window.

\subsection{Training Recipes and Checkpoint Selection}
\label{app:training-recipes}
We train the time series adapter and rank-16 LoRA ($\alpha=32$) on Qwen3.5-4B, keeping the time series encoder and remaining backbone parameters frozen. The adapter and LoRA contain 13.13M and 9.83M trainable parameters, respectively. Their peak learning rates are $10^{-3}$ and $2\times10^{-4}$, with an effective batch size of 32, gradient clipping at 1.0, and a sequence limit of 4,608 tokens. Each optimization run uses a fresh optimizer and a cosine schedule with 100 warm-up steps. Table~\ref{tab:training-recipes} summarizes the main-model and encoding-ablation recipes.
\begin{table}[b]
\vspace{-7mm}
\centering\small
\caption{\textbf{Qwen3.5-4B training recipes.} Encoding ablations correspond to Table~\ref{tab:ablation}(a); memory ablations in Table~\ref{tab:ablation}(b) reuse the main checkpoint.}
\label{tab:training-recipes}
\begin{tabular}{p{0.25\linewidth}p{0.33\linewidth}p{0.33\linewidth}}
\toprule
\textbf{Setting} & \textbf{Main checkpoint} & \textbf{Encoding ablations}\\
\midrule
Seed & 0 & 1337\\
Initial training & Up to 5,000 steps; selected step 2,500 & Fixed 1,000 steps\\
Continuation & Fixed 1,500 steps & Fixed 1,500 steps\\
Continuation sampling & T18 share 0.5 & Original task sampling\\
Channel dropout & 0.25 in training and continuation & 0 in initial training; 0.25 for hybrid in continuation; 0 for single-channel inputs\\
Warm-up & 100 steps per optimization run & 100 steps per optimization run\\
Validation interval & 100 steps & 250 steps\\
Continuation initialization & Initial training validation best & Initial training final\\
Reported checkpoint & Continuation final & Continuation final\\
\bottomrule
\end{tabular}
\end{table}

\textbf{Main checkpoint.}
Initial training selects step 2,500 using mean validation timing F1 plus $0.1\times$ content token accuracy, and stops at step 3,300 after eight validations without improvement. From the selected checkpoint, we continue training for 1,500 steps with half of samples allocated to T18, and report the final checkpoint. Memory ablations in Table~\ref{tab:ablation}(b) and both paraphrasing conditions reuse these weights.

\textbf{Encoding ablations (Table~\ref{tab:ablation}(a)).}
The three variants are \textcircled{\scriptsize 1}~\method{} (trained for 2,500 steps), \textcircled{\scriptsize 2}~Time Series Embeddings Only, and \textcircled{\scriptsize 3}~Statistical Values Only. Each uses a fixed budget of 1,000 initial-training steps plus 1,500 continuation steps, retaining the original task-sampling distribution and its specified input channels. Only \textcircled{\scriptsize 1} applies channel dropout with probability 0.25 during continuation. All three report their final checkpoints. We train \textcircled{\scriptsize 1} separately from the main model used in Table~\ref{tab:ablation}(b).

\subsection{Streaming Inference}
\label{app:inference}
Inference uses greedy decoding, $W=60$ new chunks per window, and overlap $c=15$. An example of the retained memory appears in Appendix~\ref{app:memory}. The Qwen3.5 runtime permits up to 8,192 tokens to accommodate its numerical-readout tokenization. The same runtime settings are used for the main model and encoding variants, with input channels changed only for the corresponding ablations. Hardware timing measurements are described in Appendix~\ref{app:efficiency-details}.

\subsubsection{Memory}
\label{app:memory}
Memory stores task-dependent state and indices of waveform evidence to retain.
\begin{tcolorbox}[
  colback=white,
  colframe=black!50,
  boxrule=0.3mm,
  rounded corners,
  title={Memory Example and Optional Fields},
  fonttitle=\bfseries,
  fontupper=\small,
  before skip=4pt, after skip=4pt,
  left=2mm, right=2mm, top=2mm, bottom=2mm
]
\begin{minipage}[t]{0.43\linewidth}
\textbf{T11: Numerical Reference}\par\smallskip
{\ttfamily
memory @c345\\
record: -24.7 @c133\\
keep: c133\\
last alarm: @c133\par}
\smallskip
\texttt{@c345} covers information before chunk 345. \texttt{keep: c133} retains chunk 133's waveform representation for the next window.
\end{minipage}\hfill
\begin{minipage}[t]{0.54\linewidth}
\textbf{Optional fields in other tasks}\par\smallskip
\texttt{rule}: active rule and update position\\
\texttt{events}: event types and chunk indices\\
\texttt{tally}: cumulative counts\\
\texttt{due}: next reporting position\\
\texttt{baseline}: reference statistics\\
\texttt{step}: accumulation update parameters\\
\texttt{cusum}: accumulated evidence and onset\\
\texttt{streak}/\texttt{trend}: sequential-condition state
\end{minipage}
\end{tcolorbox}

\section{Detailed Experimental Results}
\label{app:result-tables}
The main comparison evaluates code-executing models and \method on all 18 tasks, using 889 ID and 719 Real Test streams.
\subsection{Main Results}
\label{app:main-results}
\textbf{Code-generation convergence.} Each code baseline iteratively refines its program using execution feedback on up to 20 training examples per task ($k\in\{1,3,5,10,15,20\}$). Figure~\ref{fig:codegen-curriculum} evaluates each stage on the same 180 uniformly sampled ID/Real streams (11.2\% of full evaluation). Performance generally stabilizes by $k=20$, whose programs are used for the main comparison.

\begingroup
\setlength{\intextsep}{4pt}
\begin{figure}[H]
\centering
\includegraphics[width=\linewidth]{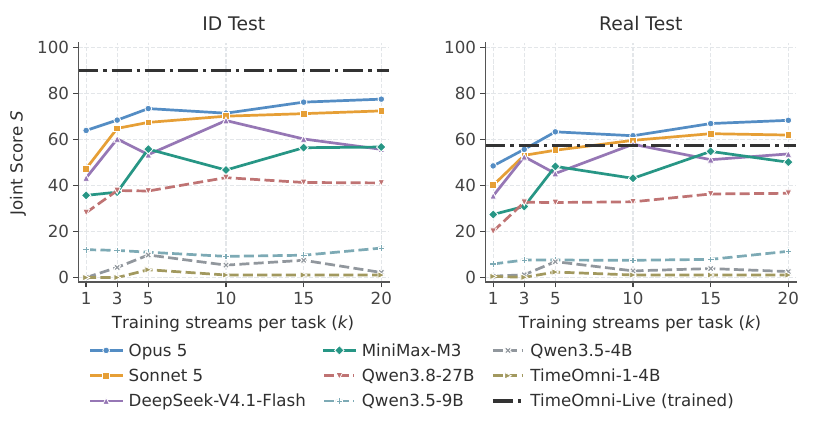}
\vspace{-5mm}
\caption{\textbf{Code-generation convergence on ID and Real Test.} Each stage uses $k$ training streams per task; later stages show diminishing gains with some fluctuations. Horizontal lines mark the trained \method reference.}
\label{fig:codegen-curriculum}
\end{figure}
\endgroup

\clearpage
\textbf{Per-task results.}
Tables~\ref{tab:task-metrics-id} and~\ref{tab:task-metrics-real} give per-task Timing F1, Content Accuracy, and Joint Score for the ID and Real results in Table~\ref{tab:main}. Capability and Overall scores average the corresponding task-level Joint Scores.
\begingroup
\newcommand{\detailours}[3]{\makebox[17.5pt][r]{#1}\hspace{4pt}\makebox[17.5pt][r]{#2}\hspace{4pt}\makebox[17.5pt][r]{#3}}
\setlength{\LTleft}{0pt}
\setlength{\LTright}{0pt}
\setlength{\LTcapwidth}{\linewidth}
\begin{longtable}{@{}p{\linewidth}@{}}
\caption{\textbf{ID Test: per-task main results.} Each model reports Timing F1 ($F_j$), Content Accuracy ($C_j$), and Joint Score ($S_j=100F_jC_j$). For readability, $F_j$ and $C_j$ are displayed as percentages; $S_j$ is on a 0--100 scale. Scores use unrounded values and the same evaluation results as Table~\ref{tab:main}. Overall averages each metric across all 18 tasks; its Joint Score is the mean of task scores, not the product of the aggregate metrics. \method is repeated for comparison.}\label{tab:tasks-0}\label{tab:task-metrics-id}\label{tab:task-metrics}\\
\endfirsthead
\multicolumn{1}{c}{\small Table~\thetable: ID Test per-task main results (continued).}\\[4pt]
\endhead
\endfoot
\endlastfoot
\begin{minipage}{\linewidth}
\centering\scriptsize
\textbf{(a) Proprietary Models}\par\vspace{3pt}
\renewcommand{\arraystretch}{1.18}
\setlength{\tabcolsep}{0.8pt}
\begin{tabular*}{\linewidth}{@{\extracolsep{\fill}}cl*{9}{r}@{}}
\toprule
\multirow{2}{*}{\textbf{Capability}} & \multicolumn{1}{c}{\multirow{2}{*}{\textbf{Task}}} & \multicolumn{3}{c}{\textbf{Opus 5}} & \multicolumn{3}{c}{\textbf{Sonnet 5}} & \multicolumn{3}{>{\columncolor{black!7}}c}{\textbf{\method}}\\
\cmidrule(lr){3-5}\cmidrule(lr){6-8}\cmidrule(lr){9-11}
 & & $F_j$ & $C_j$ & $S_j$ & $F_j$ & $C_j$ & $S_j$ & \multicolumn{3}{>{\columncolor{black!7}}c}{\detailours{$F_j$}{$C_j$}{$S_j$}}\\
\midrule
\multirow{2}{*}{\shortstack{C1: Instantaneous\\Conditions}} & T1 Single-Condition Triggering & 96.0 & 100.0 & 96.0 & 100.0 & 100.0 & 100.0 & \multicolumn{3}{>{\columncolor{black!7}}c}{\detailours{100.0}{100.0}{100.0}}\\
 & T2 Multi-Condition Triggering & 100.0 & 100.0 & 100.0 & 100.0 & 100.0 & 100.0 & \multicolumn{3}{>{\columncolor{black!7}}c}{\detailours{93.4}{80.5}{75.2}}\\
\midrule
\multirow{4}{*}{\shortstack{C2: Sequential\\Conditions}} & T3 Consecutive Exceedance & 100.0 & 100.0 & 100.0 & 68.9 & 100.0 & 68.9 & \multicolumn{3}{>{\columncolor{black!7}}c}{\detailours{100.0}{100.0}{100.0}}\\
 & T4 Trend Recognition & 100.0 & 100.0 & 100.0 & 100.0 & 100.0 & 100.0 & \multicolumn{3}{>{\columncolor{black!7}}c}{\detailours{93.5}{96.8}{90.4}}\\
 & T5 Mean Shift & 100.0 & 84.2 & 84.2 & 100.0 & 84.2 & 84.2 & \multicolumn{3}{>{\columncolor{black!7}}c}{\detailours{100.0}{80.7}{80.7}}\\
 & T6 Event Density & 100.0 & 100.0 & 100.0 & 100.0 & 100.0 & 100.0 & \multicolumn{3}{>{\columncolor{black!7}}c}{\detailours{99.3}{100.0}{99.3}}\\
\midrule
\multirow{3}{*}{\shortstack{C3: Event\\Tracking}} & T7 Single-Type Tracking & 94.8 & 100.0 & 94.8 & 100.0 & 100.0 & 100.0 & \multicolumn{3}{>{\columncolor{black!7}}c}{\detailours{98.0}{99.0}{97.0}}\\
 & T8 Multi-Type Tracking & 80.7 & 88.5 & 71.4 & 65.6 & 28.0 & 18.3 & \multicolumn{3}{>{\columncolor{black!7}}c}{\detailours{96.9}{98.6}{95.6}}\\
 & T9 Ordered Event Tracking & 93.5 & 100.0 & 93.5 & 0.0 & 0.0 & 0.0 & \multicolumn{3}{>{\columncolor{black!7}}c}{\detailours{93.0}{100.0}{93.0}}\\
\midrule
\multirow{2}{*}{\shortstack{C4: Event\\Reference}} & T10 Shape Reference & 74.8 & 95.2 & 71.3 & 81.2 & 100.0 & 81.2 & \multicolumn{3}{>{\columncolor{black!7}}c}{\detailours{100.0}{100.0}{100.0}}\\
 & T11 Numerical Reference & 100.0 & 77.6 & 77.6 & 93.6 & 74.9 & 70.1 & \multicolumn{3}{>{\columncolor{black!7}}c}{\detailours{98.5}{89.0}{87.7}}\\
\midrule
\multirow{3}{*}{\shortstack{C5: Online\\Instructions}} & T12 On-Demand Recall & 100.0 & 16.3 & 16.3 & 100.0 & 14.4 & 14.4 & \multicolumn{3}{>{\columncolor{black!7}}c}{\detailours{100.0}{85.7}{85.7}}\\
 & T13 Rule Replacement & 87.9 & 91.7 & 80.6 & 100.0 & 100.0 & 100.0 & \multicolumn{3}{>{\columncolor{black!7}}c}{\detailours{95.0}{96.0}{91.3}}\\
 & T14 Rule Addition & 88.5 & 56.6 & 50.1 & 78.4 & 48.1 & 37.7 & \multicolumn{3}{>{\columncolor{black!7}}c}{\detailours{99.6}{75.9}{75.6}}\\
\midrule
\multirow{4}{*}{\shortstack{C6: Ultra-Long\\Reliability}} & T15 Sustained Vigilance & 100.0 & 100.0 & 100.0 & 100.0 & 100.0 & 100.0 & \multicolumn{3}{>{\columncolor{black!7}}c}{\detailours{100.0}{100.0}{100.0}}\\
 & T16 Long-Gap Reference & 73.8 & 100.0 & 73.8 & 73.5 & 100.0 & 73.5 & \multicolumn{3}{>{\columncolor{black!7}}c}{\detailours{100.0}{100.0}{100.0}}\\
 & T17 Sustained Reporting & 99.8 & 17.3 & 17.2 & 99.8 & 12.5 & 12.5 & \multicolumn{3}{>{\columncolor{black!7}}c}{\detailours{99.8}{97.9}{97.7}}\\
 & T18 Cumulative Evidence & 100.0 & 100.0 & 100.0 & 100.0 & 100.0 & 100.0 & \multicolumn{3}{>{\columncolor{black!7}}c}{\detailours{89.3}{55.9}{50.0}}\\
\midrule
\multicolumn{2}{l}{\textbf{Overall}} & 93.9 & 84.9 & 79.3 & 86.7 & 75.7 & 70.0 & \multicolumn{3}{>{\columncolor{black!7}}c}{\detailours{97.6}{92.0}{90.0}}\\
\bottomrule
\end{tabular*}
\end{minipage}\\[12pt]
\begin{minipage}{\linewidth}
\centering\scriptsize
\vspace{8pt}
\textbf{(b) Open-Source Models (1/2)}\par\vspace{3pt}
\renewcommand{\arraystretch}{1.18}
\setlength{\tabcolsep}{0.8pt}
\begin{tabular*}{\linewidth}{@{\extracolsep{\fill}}cl*{12}{r}@{}}
\toprule
\multirow{2}{*}{\textbf{Capability}} & \multicolumn{1}{c}{\multirow{2}{*}{\textbf{Task}}} & \multicolumn{3}{c}{\textbf{\begin{tabular}[c]{@{}c@{}}DeepSeek-\\V4.1-Flash\end{tabular}}} & \multicolumn{3}{c}{\textbf{MiniMax-M3}} & \multicolumn{3}{c}{\textbf{Qwen3.8-27B}} & \multicolumn{3}{>{\columncolor{black!7}}c}{\textbf{\method}}\\
\cmidrule(lr){3-5}\cmidrule(lr){6-8}\cmidrule(lr){9-11}\cmidrule(lr){12-14}
 & & $F_j$ & $C_j$ & $S_j$ & $F_j$ & $C_j$ & $S_j$ & $F_j$ & $C_j$ & $S_j$ & \multicolumn{3}{>{\columncolor{black!7}}c}{\detailours{$F_j$}{$C_j$}{$S_j$}}\\
\midrule
\multirow{2}{*}{\shortstack{C1: Instantaneous\\Conditions}} & T1 Single-Condition Triggering & 100.0 & 100.0 & 100.0 & 100.0 & 100.0 & 100.0 & 90.0 & 100.0 & 90.0 & \multicolumn{3}{>{\columncolor{black!7}}c}{\detailours{100.0}{100.0}{100.0}}\\
 & T2 Multi-Condition Triggering & 100.0 & 100.0 & 100.0 & 100.0 & 99.7 & 99.7 & 20.0 & 92.0 & 18.4 & \multicolumn{3}{>{\columncolor{black!7}}c}{\detailours{93.4}{80.5}{75.2}}\\
\midrule
\multirow{4}{*}{\shortstack{C2: Sequential\\Conditions}} & T3 Consecutive Exceedance & 100.0 & 100.0 & 100.0 & 68.9 & 100.0 & 68.9 & 12.7 & 96.1 & 12.2 & \multicolumn{3}{>{\columncolor{black!7}}c}{\detailours{100.0}{100.0}{100.0}}\\
 & T4 Trend Recognition & 90.0 & 100.0 & 90.0 & 94.5 & 95.9 & 90.7 & 88.0 & 100.0 & 88.0 & \multicolumn{3}{>{\columncolor{black!7}}c}{\detailours{93.5}{96.8}{90.4}}\\
 & T5 Mean Shift & 92.0 & 81.7 & 75.2 & 100.0 & 84.2 & 84.2 & 34.7 & 90.8 & 31.5 & \multicolumn{3}{>{\columncolor{black!7}}c}{\detailours{100.0}{80.7}{80.7}}\\
 & T6 Event Density & 86.0 & 100.0 & 86.0 & 88.8 & 97.8 & 86.8 & 94.0 & 100.0 & 94.0 & \multicolumn{3}{>{\columncolor{black!7}}c}{\detailours{99.3}{100.0}{99.3}}\\
\midrule
\multirow{3}{*}{\shortstack{C3: Event\\Tracking}} & T7 Single-Type Tracking & 100.0 & 100.0 & 100.0 & 49.8 & 69.6 & 34.7 & 89.6 & 100.0 & 89.6 & \multicolumn{3}{>{\columncolor{black!7}}c}{\detailours{98.0}{99.0}{97.0}}\\
 & T8 Multi-Type Tracking & 6.1 & 37.5 & 2.3 & 21.6 & 28.1 & 6.1 & 27.1 & 52.3 & 14.2 & \multicolumn{3}{>{\columncolor{black!7}}c}{\detailours{96.9}{98.6}{95.6}}\\
 & T9 Ordered Event Tracking & 92.3 & 100.0 & 92.3 & 8.3 & 0.0 & 0.0 & 1.0 & 100.0 & 1.0 & \multicolumn{3}{>{\columncolor{black!7}}c}{\detailours{93.0}{100.0}{93.0}}\\
\midrule
\multirow{2}{*}{\shortstack{C4: Event\\Reference}} & T10 Shape Reference & 0.0 & 0.0 & 0.0 & 43.9 & 100.0 & 43.9 & 0.0 & 0.0 & 0.0 & \multicolumn{3}{>{\columncolor{black!7}}c}{\detailours{100.0}{100.0}{100.0}}\\
 & T11 Numerical Reference & 61.7 & 0.0 & 0.0 & 38.5 & 0.0 & 0.0 & 38.0 & 93.5 & 35.5 & \multicolumn{3}{>{\columncolor{black!7}}c}{\detailours{98.5}{89.0}{87.7}}\\
\midrule
\multirow{3}{*}{\shortstack{C5: Online\\Instructions}} & T12 On-Demand Recall & 0.0 & 0.0 & 0.0 & 95.0 & 20.7 & 19.7 & 98.6 & 13.0 & 12.8 & \multicolumn{3}{>{\columncolor{black!7}}c}{\detailours{100.0}{85.7}{85.7}}\\
 & T13 Rule Replacement & 0.0 & 0.0 & 0.0 & 100.0 & 100.0 & 100.0 & 79.6 & 100.0 & 79.6 & \multicolumn{3}{>{\columncolor{black!7}}c}{\detailours{95.0}{96.0}{91.3}}\\
 & T14 Rule Addition & 0.0 & 0.0 & 0.0 & 58.3 & 38.2 & 22.3 & 49.7 & 44.1 & 21.9 & \multicolumn{3}{>{\columncolor{black!7}}c}{\detailours{99.6}{75.9}{75.6}}\\
\midrule
\multirow{4}{*}{\shortstack{C6: Ultra-Long\\Reliability}} & T15 Sustained Vigilance & 100.0 & 100.0 & 100.0 & 100.0 & 100.0 & 100.0 & 85.4 & 100.0 & 85.4 & \multicolumn{3}{>{\columncolor{black!7}}c}{\detailours{100.0}{100.0}{100.0}}\\
 & T16 Long-Gap Reference & 36.0 & 100.0 & 36.0 & 94.2 & 100.0 & 94.2 & 0.6 & 12.3 & 0.1 & \multicolumn{3}{>{\columncolor{black!7}}c}{\detailours{100.0}{100.0}{100.0}}\\
 & T17 Sustained Reporting & 99.8 & 15.0 & 15.0 & 99.8 & 3.0 & 3.0 & 99.8 & 15.6 & 15.6 & \multicolumn{3}{>{\columncolor{black!7}}c}{\detailours{99.8}{97.9}{97.7}}\\
 & T18 Cumulative Evidence & 100.0 & 100.0 & 100.0 & 70.0 & 98.8 & 69.2 & 100.0 & 100.0 & 100.0 & \multicolumn{3}{>{\columncolor{black!7}}c}{\detailours{89.3}{55.9}{50.0}}\\
\midrule
\multicolumn{2}{l}{\textbf{Overall}} & 64.7 & 63.0 & 55.4 & 74.0 & 68.7 & 56.9 & 56.1 & 72.8 & 43.9 & \multicolumn{3}{>{\columncolor{black!7}}c}{\detailours{97.6}{92.0}{90.0}}\\
\bottomrule
\end{tabular*}
\end{minipage}\\[12pt]
\begin{minipage}{\linewidth}
\centering\scriptsize
\textbf{(c) Open-Source Models (2/2)}\par\vspace{3pt}
\renewcommand{\arraystretch}{1.18}
\setlength{\tabcolsep}{0.8pt}
\begin{tabular*}{\linewidth}{@{\extracolsep{\fill}}cl*{12}{r}@{}}
\toprule
\multirow{2}{*}{\textbf{Capability}} & \multicolumn{1}{c}{\multirow{2}{*}{\textbf{Task}}} & \multicolumn{3}{c}{\textbf{Qwen3.5-9B}} & \multicolumn{3}{c}{\textbf{Qwen3.5-4B}} & \multicolumn{3}{c}{\textbf{\begin{tabular}[c]{@{}c@{}}Qwen3.5-4B\\(streaming)\end{tabular}}} & \multicolumn{3}{>{\columncolor{black!7}}c}{\textbf{\method}}\\
\cmidrule(lr){3-5}\cmidrule(lr){6-8}\cmidrule(lr){9-11}\cmidrule(lr){12-14}
 & & $F_j$ & $C_j$ & $S_j$ & $F_j$ & $C_j$ & $S_j$ & $F_j$ & $C_j$ & $S_j$ & \multicolumn{3}{>{\columncolor{black!7}}c}{\detailours{$F_j$}{$C_j$}{$S_j$}}\\
\midrule
\multirow{2}{*}{\shortstack{C1: Instantaneous\\Conditions}} & T1 Single-Condition Triggering & 86.7 & 78.6 & 68.1 & 0.0 & 0.0 & 0.0 & 4.3 & 100.0 & 4.3 & \multicolumn{3}{>{\columncolor{black!7}}c}{\detailours{100.0}{100.0}{100.0}}\\
 & T2 Multi-Condition Triggering & 40.9 & 94.9 & 38.8 & 19.6 & 94.9 & 18.6 & 0.0 & 0.0 & 0.0 & \multicolumn{3}{>{\columncolor{black!7}}c}{\detailours{93.4}{80.5}{75.2}}\\
\midrule
\multirow{4}{*}{\shortstack{C2: Sequential\\Conditions}} & T3 Consecutive Exceedance & 0.0 & 0.0 & 0.0 & 0.0 & 0.0 & 0.0 & 0.0 & 0.0 & 0.0 & \multicolumn{3}{>{\columncolor{black!7}}c}{\detailours{100.0}{100.0}{100.0}}\\
 & T4 Trend Recognition & 24.0 & 100.0 & 24.0 & 28.5 & 91.5 & 26.1 & 0.0 & 0.0 & 0.0 & \multicolumn{3}{>{\columncolor{black!7}}c}{\detailours{93.5}{96.8}{90.4}}\\
 & T5 Mean Shift & 8.9 & 79.7 & 7.1 & 0.0 & 0.0 & 0.0 & 0.0 & 0.0 & 0.0 & \multicolumn{3}{>{\columncolor{black!7}}c}{\detailours{100.0}{80.7}{80.7}}\\
 & T6 Event Density & 18.0 & 10.0 & 1.8 & 4.8 & 11.0 & 0.5 & 0.0 & 0.0 & 0.0 & \multicolumn{3}{>{\columncolor{black!7}}c}{\detailours{99.3}{100.0}{99.3}}\\
\midrule
\multirow{3}{*}{\shortstack{C3: Event\\Tracking}} & T7 Single-Type Tracking & 0.0 & 0.0 & 0.0 & 0.0 & 0.0 & 0.0 & 0.4 & 0.0 & 0.0 & \multicolumn{3}{>{\columncolor{black!7}}c}{\detailours{98.0}{99.0}{97.0}}\\
 & T8 Multi-Type Tracking & 0.0 & 0.0 & 0.0 & 0.0 & 0.0 & 0.0 & 0.0 & 0.0 & 0.0 & \multicolumn{3}{>{\columncolor{black!7}}c}{\detailours{96.9}{98.6}{95.6}}\\
 & T9 Ordered Event Tracking & 0.0 & 0.0 & 0.0 & 0.0 & 0.0 & 0.0 & 0.0 & 0.0 & 0.0 & \multicolumn{3}{>{\columncolor{black!7}}c}{\detailours{93.0}{100.0}{93.0}}\\
\midrule
\multirow{2}{*}{\shortstack{C4: Event\\Reference}} & T10 Shape Reference & 0.0 & 0.0 & 0.0 & 0.0 & 0.0 & 0.0 & 0.1 & 0.0 & 0.0 & \multicolumn{3}{>{\columncolor{black!7}}c}{\detailours{100.0}{100.0}{100.0}}\\
 & T11 Numerical Reference & 50.2 & 0.0 & 0.0 & 0.0 & 0.0 & 0.0 & 0.0 & 0.0 & 0.0 & \multicolumn{3}{>{\columncolor{black!7}}c}{\detailours{98.5}{89.0}{87.7}}\\
\midrule
\multirow{3}{*}{\shortstack{C5: Online\\Instructions}} & T12 On-Demand Recall & 31.7 & 40.2 & 12.8 & 0.0 & 0.0 & 0.0 & 0.0 & 0.0 & 0.0 & \multicolumn{3}{>{\columncolor{black!7}}c}{\detailours{100.0}{85.7}{85.7}}\\
 & T13 Rule Replacement & 0.0 & 0.0 & 0.0 & 21.7 & 1.6 & 0.3 & 0.8 & 0.0 & 0.0 & \multicolumn{3}{>{\columncolor{black!7}}c}{\detailours{95.0}{96.0}{91.3}}\\
 & T14 Rule Addition & 42.6 & 51.6 & 22.0 & 0.0 & 0.0 & 0.0 & 0.0 & 0.0 & 0.0 & \multicolumn{3}{>{\columncolor{black!7}}c}{\detailours{99.6}{75.9}{75.6}}\\
\midrule
\multirow{4}{*}{\shortstack{C6: Ultra-Long\\Reliability}} & T15 Sustained Vigilance & 68.0 & 0.0 & 0.0 & 0.0 & 0.0 & 0.0 & 0.0 & 100.0 & 0.0 & \multicolumn{3}{>{\columncolor{black!7}}c}{\detailours{100.0}{100.0}{100.0}}\\
 & T16 Long-Gap Reference & 0.0 & 0.0 & 0.0 & 0.0 & 0.0 & 0.0 & 0.0 & 0.0 & 0.0 & \multicolumn{3}{>{\columncolor{black!7}}c}{\detailours{100.0}{100.0}{100.0}}\\
 & T17 Sustained Reporting & 0.0 & 0.0 & 0.0 & 0.0 & 0.0 & 0.0 & 0.0 & 0.0 & 0.0 & \multicolumn{3}{>{\columncolor{black!7}}c}{\detailours{99.8}{97.9}{97.7}}\\
 & T18 Cumulative Evidence & 0.2 & 80.6 & 0.2 & 0.0 & 4.2 & 0.0 & 0.0 & 0.0 & 0.0 & \multicolumn{3}{>{\columncolor{black!7}}c}{\detailours{89.3}{55.9}{50.0}}\\
\midrule
\multicolumn{2}{l}{\textbf{Overall}} & 20.6 & 29.8 & 9.7 & 4.1 & 11.3 & 2.5 & 0.3 & 11.1 & 0.2 & \multicolumn{3}{>{\columncolor{black!7}}c}{\detailours{97.6}{92.0}{90.0}}\\
\bottomrule
\end{tabular*}
\end{minipage}\\[12pt]
\begin{minipage}{\linewidth}
\centering\scriptsize
\vspace{8pt}
\textbf{(d) Time Series Language Models}\par\vspace{3pt}
\renewcommand{\arraystretch}{1.18}
\setlength{\tabcolsep}{0.8pt}
\begin{tabular*}{\linewidth}{@{\extracolsep{\fill}}cl*{9}{r}@{}}
\toprule
\multirow{2}{*}{\textbf{Capability}} & \multicolumn{1}{c}{\multirow{2}{*}{\textbf{Task}}} & \multicolumn{3}{c}{\textbf{TimeOmni-1-4B}} & \multicolumn{3}{c}{\textbf{\begin{tabular}[c]{@{}c@{}}TS-Reasoner-7B\\(streaming)\end{tabular}}} & \multicolumn{3}{>{\columncolor{black!7}}c}{\textbf{\method}}\\
\cmidrule(lr){3-5}\cmidrule(lr){6-8}\cmidrule(lr){9-11}
 & & $F_j$ & $C_j$ & $S_j$ & $F_j$ & $C_j$ & $S_j$ & \multicolumn{3}{>{\columncolor{black!7}}c}{\detailours{$F_j$}{$C_j$}{$S_j$}}\\
\midrule
\multirow{2}{*}{\shortstack{C1: Instantaneous\\Conditions}} & T1 Single-Condition Triggering & 0.0 & 0.0 & 0.0 & 11.0 & 0.0 & 0.0 & \multicolumn{3}{>{\columncolor{black!7}}c}{\detailours{100.0}{100.0}{100.0}}\\
 & T2 Multi-Condition Triggering & 0.0 & 0.0 & 0.0 & 9.8 & 19.6 & 1.9 & \multicolumn{3}{>{\columncolor{black!7}}c}{\detailours{93.4}{80.5}{75.2}}\\
\midrule
\multirow{4}{*}{\shortstack{C2: Sequential\\Conditions}} & T3 Consecutive Exceedance & 0.0 & 0.0 & 0.0 & 7.1 & 0.0 & 0.0 & \multicolumn{3}{>{\columncolor{black!7}}c}{\detailours{100.0}{100.0}{100.0}}\\
 & T4 Trend Recognition & 0.0 & 0.0 & 0.0 & 6.7 & 50.0 & 3.4 & \multicolumn{3}{>{\columncolor{black!7}}c}{\detailours{93.5}{96.8}{90.4}}\\
 & T5 Mean Shift & 0.0 & 0.0 & 0.0 & 3.8 & 30.0 & 1.1 & \multicolumn{3}{>{\columncolor{black!7}}c}{\detailours{100.0}{80.7}{80.7}}\\
 & T6 Event Density & 0.0 & 0.0 & 0.0 & 4.7 & 0.0 & 0.0 & \multicolumn{3}{>{\columncolor{black!7}}c}{\detailours{99.3}{100.0}{99.3}}\\
\midrule
\multirow{3}{*}{\shortstack{C3: Event\\Tracking}} & T7 Single-Type Tracking & 0.0 & 0.0 & 0.0 & 0.9 & 7.7 & 0.1 & \multicolumn{3}{>{\columncolor{black!7}}c}{\detailours{98.0}{99.0}{97.0}}\\
 & T8 Multi-Type Tracking & 0.0 & 0.0 & 0.0 & 0.8 & 35.0 & 0.3 & \multicolumn{3}{>{\columncolor{black!7}}c}{\detailours{96.9}{98.6}{95.6}}\\
 & T9 Ordered Event Tracking & 0.0 & 0.0 & 0.0 & 0.6 & 0.0 & 0.0 & \multicolumn{3}{>{\columncolor{black!7}}c}{\detailours{93.0}{100.0}{93.0}}\\
\midrule
\multirow{2}{*}{\shortstack{C4: Event\\Reference}} & T10 Shape Reference & 0.0 & 0.0 & 0.0 & 1.3 & 0.0 & 0.0 & \multicolumn{3}{>{\columncolor{black!7}}c}{\detailours{100.0}{100.0}{100.0}}\\
 & T11 Numerical Reference & 0.0 & 0.0 & 0.0 & 0.9 & 0.0 & 0.0 & \multicolumn{3}{>{\columncolor{black!7}}c}{\detailours{98.5}{89.0}{87.7}}\\
\midrule
\multirow{3}{*}{\shortstack{C5: Online\\Instructions}} & T12 On-Demand Recall & 3.5 & 16.7 & 0.6 & 6.3 & 6.7 & 0.4 & \multicolumn{3}{>{\columncolor{black!7}}c}{\detailours{100.0}{85.7}{85.7}}\\
 & T13 Rule Replacement & 0.0 & 0.0 & 0.0 & 1.4 & 3.4 & 0.0 & \multicolumn{3}{>{\columncolor{black!7}}c}{\detailours{95.0}{96.0}{91.3}}\\
 & T14 Rule Addition & 34.7 & 35.4 & 12.3 & 3.7 & 24.9 & 0.9 & \multicolumn{3}{>{\columncolor{black!7}}c}{\detailours{99.6}{75.9}{75.6}}\\
\midrule
\multirow{4}{*}{\shortstack{C6: Ultra-Long\\Reliability}} & T15 Sustained Vigilance & 0.0 & 0.0 & 0.0 & 0.2 & 13.0 & 0.0 & \multicolumn{3}{>{\columncolor{black!7}}c}{\detailours{100.0}{100.0}{100.0}}\\
 & T16 Long-Gap Reference & 0.1 & 0.0 & 0.0 & 0.2 & 3.3 & 0.0 & \multicolumn{3}{>{\columncolor{black!7}}c}{\detailours{100.0}{100.0}{100.0}}\\
 & T17 Sustained Reporting & 0.0 & 0.0 & 0.0 & 0.4 & 16.1 & 0.1 & \multicolumn{3}{>{\columncolor{black!7}}c}{\detailours{99.8}{97.9}{97.7}}\\
 & T18 Cumulative Evidence & 0.0 & 0.0 & 0.0 & 0.1 & 3.8 & 0.0 & \multicolumn{3}{>{\columncolor{black!7}}c}{\detailours{89.3}{55.9}{50.0}}\\
\midrule
\multicolumn{2}{l}{\textbf{Overall}} & 2.1 & 2.9 & 0.7 & 3.3 & 11.9 & 0.5 & \multicolumn{3}{>{\columncolor{black!7}}c}{\detailours{97.6}{92.0}{90.0}}\\
\bottomrule
\end{tabular*}
\end{minipage}\\[12pt]
\end{longtable}
\endgroup

\clearpage
\begingroup
\newcommand{\detailours}[3]{\makebox[17.5pt][r]{#1}\hspace{4pt}\makebox[17.5pt][r]{#2}\hspace{4pt}\makebox[17.5pt][r]{#3}}
\setlength{\LTleft}{0pt}
\setlength{\LTright}{0pt}
\setlength{\LTcapwidth}{\linewidth}
\begin{longtable}{@{}p{\linewidth}@{}}
\caption{\textbf{Real Test: per-task main results.} Each model reports Timing F1 ($F_j$), Content Accuracy ($C_j$), and Joint Score ($S_j=100F_jC_j$). For readability, $F_j$ and $C_j$ are displayed as percentages; $S_j$ is on a 0--100 scale. Scores use unrounded values and the same evaluation results as Table~\ref{tab:main}. Overall averages each metric across all 18 tasks; its Joint Score is the mean of task scores, not the product of the aggregate metrics. \method is repeated for comparison.}\label{tab:tasks-2}\label{tab:task-metrics-real}\\
\endfirsthead
\multicolumn{1}{c}{\small Table~\thetable: Real Test per-task main results (continued).}\\[4pt]
\endhead
\endfoot
\endlastfoot
\begin{minipage}{\linewidth}
\centering\scriptsize
\textbf{(a) Proprietary Models}\par\vspace{3pt}
\renewcommand{\arraystretch}{1.18}
\setlength{\tabcolsep}{0.8pt}
\begin{tabular*}{\linewidth}{@{\extracolsep{\fill}}cl*{9}{r}@{}}
\toprule
\multirow{2}{*}{\textbf{Capability}} & \multicolumn{1}{c}{\multirow{2}{*}{\textbf{Task}}} & \multicolumn{3}{c}{\textbf{Opus 5}} & \multicolumn{3}{c}{\textbf{Sonnet 5}} & \multicolumn{3}{>{\columncolor{black!7}}c}{\textbf{\method}}\\
\cmidrule(lr){3-5}\cmidrule(lr){6-8}\cmidrule(lr){9-11}
 & & $F_j$ & $C_j$ & $S_j$ & $F_j$ & $C_j$ & $S_j$ & \multicolumn{3}{>{\columncolor{black!7}}c}{\detailours{$F_j$}{$C_j$}{$S_j$}}\\
\midrule
\multirow{2}{*}{\shortstack{C1: Instantaneous\\Conditions}} & T1 Single-Condition Triggering & 96.0 & 100.0 & 96.0 & 100.0 & 100.0 & 100.0 & \multicolumn{3}{>{\columncolor{black!7}}c}{\detailours{97.5}{99.0}{96.5}}\\
 & T2 Multi-Condition Triggering & 100.0 & 100.0 & 100.0 & 100.0 & 100.0 & 100.0 & \multicolumn{3}{>{\columncolor{black!7}}c}{\detailours{93.8}{76.0}{71.4}}\\
\midrule
\multirow{4}{*}{\shortstack{C2: Sequential\\Conditions}} & T3 Consecutive Exceedance & 100.0 & 100.0 & 100.0 & 85.1 & 100.0 & 85.1 & \multicolumn{3}{>{\columncolor{black!7}}c}{\detailours{96.4}{100.0}{96.4}}\\
 & T4 Trend Recognition & 100.0 & 100.0 & 100.0 & 100.0 & 100.0 & 100.0 & \multicolumn{3}{>{\columncolor{black!7}}c}{\detailours{91.6}{96.4}{88.3}}\\
 & T5 Mean Shift & 100.0 & 80.1 & 80.1 & 100.0 & 80.1 & 80.1 & \multicolumn{3}{>{\columncolor{black!7}}c}{\detailours{85.3}{82.7}{70.5}}\\
 & T6 Event Density & 100.0 & 100.0 & 100.0 & 97.3 & 100.0 & 97.3 & \multicolumn{3}{>{\columncolor{black!7}}c}{\detailours{68.3}{97.4}{66.5}}\\
\midrule
\multirow{3}{*}{\shortstack{C3: Event\\Tracking}} & T7 Single-Type Tracking & 96.5 & 100.0 & 96.5 & 100.0 & 100.0 & 100.0 & \multicolumn{3}{>{\columncolor{black!7}}c}{\detailours{83.4}{74.8}{62.4}}\\
 & T8 Multi-Type Tracking & 41.3 & 57.1 & 23.6 & 3.3 & 50.0 & 1.7 & \multicolumn{3}{>{\columncolor{black!7}}c}{\detailours{22.1}{40.3}{8.9}}\\
 & T9 Ordered Event Tracking & 94.1 & 100.0 & 94.1 & 0.0 & 0.0 & 0.0 & \multicolumn{3}{>{\columncolor{black!7}}c}{\detailours{63.8}{85.4}{54.4}}\\
\midrule
\multirow{2}{*}{\shortstack{C4: Event\\Reference}} & T10 Shape Reference & 47.3 & 100.0 & 47.3 & 4.5 & 100.0 & 4.5 & \multicolumn{3}{>{\columncolor{black!7}}c}{\detailours{38.5}{100.0}{38.5}}\\
 & T11 Numerical Reference & 92.5 & 83.1 & 76.9 & 93.5 & 83.8 & 78.4 & \multicolumn{3}{>{\columncolor{black!7}}c}{\detailours{85.5}{82.8}{70.8}}\\
\midrule
\multirow{3}{*}{\shortstack{C5: Online\\Instructions}} & T12 On-Demand Recall & 100.0 & 23.7 & 23.7 & 100.0 & 25.0 & 25.0 & \multicolumn{3}{>{\columncolor{black!7}}c}{\detailours{85.9}{39.3}{33.8}}\\
 & T13 Rule Replacement & 92.4 & 98.1 & 90.7 & 100.0 & 100.0 & 100.0 & \multicolumn{3}{>{\columncolor{black!7}}c}{\detailours{77.9}{76.7}{59.8}}\\
 & T14 Rule Addition & 70.2 & 79.4 & 55.8 & 40.0 & 65.6 & 26.2 & \multicolumn{3}{>{\columncolor{black!7}}c}{\detailours{66.1}{56.9}{37.6}}\\
\midrule
\multirow{4}{*}{\shortstack{C6: Ultra-Long\\Reliability}} & T15 Sustained Vigilance & 100.0 & 100.0 & 100.0 & 100.0 & 100.0 & 100.0 & \multicolumn{3}{>{\columncolor{black!7}}c}{\detailours{99.2}{100.0}{99.2}}\\
 & T16 Long-Gap Reference & 0.5 & 100.0 & 0.5 & 0.0 & 0.0 & 0.0 & \multicolumn{3}{>{\columncolor{black!7}}c}{\detailours{0.6}{100.0}{0.6}}\\
 & T17 Sustained Reporting & 99.4 & 6.0 & 6.0 & 99.4 & 41.3 & 41.1 & \multicolumn{3}{>{\columncolor{black!7}}c}{\detailours{99.4}{49.0}{48.6}}\\
 & T18 Cumulative Evidence & 15.4 & 43.8 & 6.7 & 0.0 & 0.0 & 0.0 & \multicolumn{3}{>{\columncolor{black!7}}c}{\detailours{11.5}{39.7}{4.6}}\\
\midrule
\multicolumn{2}{l}{\textbf{Overall}} & 80.3 & 81.7 & 66.5 & 68.0 & 69.2 & 57.7 & \multicolumn{3}{>{\columncolor{black!7}}c}{\detailours{70.4}{77.6}{56.1}}\\
\bottomrule
\end{tabular*}
\end{minipage}\\[12pt]
\begin{minipage}{\linewidth}
\centering\scriptsize
\vspace{8pt}
\textbf{(b) Open-Source Models (1/2)}\par\vspace{3pt}
\renewcommand{\arraystretch}{1.18}
\setlength{\tabcolsep}{0.8pt}
\begin{tabular*}{\linewidth}{@{\extracolsep{\fill}}cl*{12}{r}@{}}
\toprule
\multirow{2}{*}{\textbf{Capability}} & \multicolumn{1}{c}{\multirow{2}{*}{\textbf{Task}}} & \multicolumn{3}{c}{\textbf{\begin{tabular}[c]{@{}c@{}}DeepSeek-\\V4.1-Flash\end{tabular}}} & \multicolumn{3}{c}{\textbf{MiniMax-M3}} & \multicolumn{3}{c}{\textbf{Qwen3.8-27B}} & \multicolumn{3}{>{\columncolor{black!7}}c}{\textbf{\method}}\\
\cmidrule(lr){3-5}\cmidrule(lr){6-8}\cmidrule(lr){9-11}\cmidrule(lr){12-14}
 & & $F_j$ & $C_j$ & $S_j$ & $F_j$ & $C_j$ & $S_j$ & $F_j$ & $C_j$ & $S_j$ & \multicolumn{3}{>{\columncolor{black!7}}c}{\detailours{$F_j$}{$C_j$}{$S_j$}}\\
\midrule
\multirow{2}{*}{\shortstack{C1: Instantaneous\\Conditions}} & T1 Single-Condition Triggering & 100.0 & 100.0 & 100.0 & 100.0 & 100.0 & 100.0 & 86.0 & 100.0 & 86.0 & \multicolumn{3}{>{\columncolor{black!7}}c}{\detailours{97.5}{99.0}{96.5}}\\
 & T2 Multi-Condition Triggering & 100.0 & 100.0 & 100.0 & 100.0 & 100.0 & 100.0 & 6.0 & 98.9 & 5.9 & \multicolumn{3}{>{\columncolor{black!7}}c}{\detailours{93.8}{76.0}{71.4}}\\
\midrule
\multirow{4}{*}{\shortstack{C2: Sequential\\Conditions}} & T3 Consecutive Exceedance & 100.0 & 100.0 & 100.0 & 85.1 & 100.0 & 85.1 & 23.8 & 92.9 & 22.1 & \multicolumn{3}{>{\columncolor{black!7}}c}{\detailours{96.4}{100.0}{96.4}}\\
 & T4 Trend Recognition & 98.0 & 100.0 & 98.0 & 98.0 & 100.0 & 98.0 & 98.0 & 98.0 & 96.0 & \multicolumn{3}{>{\columncolor{black!7}}c}{\detailours{91.6}{96.4}{88.3}}\\
 & T5 Mean Shift & 91.0 & 79.0 & 71.9 & 100.0 & 80.1 & 80.1 & 24.7 & 81.1 & 20.0 & \multicolumn{3}{>{\columncolor{black!7}}c}{\detailours{85.3}{82.7}{70.5}}\\
 & T6 Event Density & 86.0 & 100.0 & 86.0 & 86.0 & 100.0 & 86.0 & 95.3 & 100.0 & 95.3 & \multicolumn{3}{>{\columncolor{black!7}}c}{\detailours{68.3}{97.4}{66.5}}\\
\midrule
\multirow{3}{*}{\shortstack{C3: Event\\Tracking}} & T7 Single-Type Tracking & 100.0 & 100.0 & 100.0 & 55.2 & 87.6 & 48.4 & 92.2 & 100.0 & 92.2 & \multicolumn{3}{>{\columncolor{black!7}}c}{\detailours{83.4}{74.8}{62.4}}\\
 & T8 Multi-Type Tracking & 2.8 & 25.9 & 0.7 & 10.6 & 20.4 & 2.2 & 0.0 & 0.0 & 0.0 & \multicolumn{3}{>{\columncolor{black!7}}c}{\detailours{22.1}{40.3}{8.9}}\\
 & T9 Ordered Event Tracking & 85.5 & 100.0 & 85.5 & 0.6 & 0.0 & 0.0 & 7.3 & 100.0 & 7.3 & \multicolumn{3}{>{\columncolor{black!7}}c}{\detailours{63.8}{85.4}{54.4}}\\
\midrule
\multirow{2}{*}{\shortstack{C4: Event\\Reference}} & T10 Shape Reference & 8.2 & 100.0 & 8.2 & 0.0 & 0.0 & 0.0 & 0.0 & 0.0 & 0.0 & \multicolumn{3}{>{\columncolor{black!7}}c}{\detailours{38.5}{100.0}{38.5}}\\
 & T11 Numerical Reference & 66.9 & 1.1 & 0.8 & 52.2 & 0.0 & 0.0 & 52.0 & 99.0 & 51.5 & \multicolumn{3}{>{\columncolor{black!7}}c}{\detailours{85.5}{82.8}{70.8}}\\
\midrule
\multirow{3}{*}{\shortstack{C5: Online\\Instructions}} & T12 On-Demand Recall & 0.0 & 0.0 & 0.0 & 91.9 & 23.4 & 21.5 & 99.3 & 17.7 & 17.5 & \multicolumn{3}{>{\columncolor{black!7}}c}{\detailours{85.9}{39.3}{33.8}}\\
 & T13 Rule Replacement & 0.0 & 0.0 & 0.0 & 100.0 & 100.0 & 100.0 & 85.1 & 100.0 & 85.1 & \multicolumn{3}{>{\columncolor{black!7}}c}{\detailours{77.9}{76.7}{59.8}}\\
 & T14 Rule Addition & 0.0 & 0.0 & 0.0 & 56.0 & 54.2 & 30.4 & 34.8 & 62.0 & 21.6 & \multicolumn{3}{>{\columncolor{black!7}}c}{\detailours{66.1}{56.9}{37.6}}\\
\midrule
\multirow{4}{*}{\shortstack{C6: Ultra-Long\\Reliability}} & T15 Sustained Vigilance & 100.0 & 100.0 & 100.0 & 100.0 & 100.0 & 100.0 & 78.8 & 100.0 & 78.8 & \multicolumn{3}{>{\columncolor{black!7}}c}{\detailours{99.2}{100.0}{99.2}}\\
 & T16 Long-Gap Reference & 1.5 & 100.0 & 1.5 & 1.0 & 100.0 & 1.0 & 1.3 & 0.0 & 0.0 & \multicolumn{3}{>{\columncolor{black!7}}c}{\detailours{0.6}{100.0}{0.6}}\\
 & T17 Sustained Reporting & 99.4 & 58.4 & 58.0 & 82.2 & 0.0 & 0.0 & 99.4 & 30.6 & 30.4 & \multicolumn{3}{>{\columncolor{black!7}}c}{\detailours{99.4}{49.0}{48.6}}\\
 & T18 Cumulative Evidence & 15.4 & 41.7 & 6.4 & 11.5 & 21.9 & 2.5 & 0.0 & 0.0 & 0.0 & \multicolumn{3}{>{\columncolor{black!7}}c}{\detailours{11.5}{39.7}{4.6}}\\
\midrule
\multicolumn{2}{l}{\textbf{Overall}} & 58.6 & 67.0 & 50.9 & 62.8 & 60.4 & 47.5 & 49.1 & 65.6 & 39.4 & \multicolumn{3}{>{\columncolor{black!7}}c}{\detailours{70.4}{77.6}{56.1}}\\
\bottomrule
\end{tabular*}
\end{minipage}\\[12pt]
\begin{minipage}{\linewidth}
\centering\scriptsize
\textbf{(c) Open-Source Models (2/2)}\par\vspace{3pt}
\renewcommand{\arraystretch}{1.18}
\setlength{\tabcolsep}{0.8pt}
\begin{tabular*}{\linewidth}{@{\extracolsep{\fill}}cl*{12}{r}@{}}
\toprule
\multirow{2}{*}{\textbf{Capability}} & \multicolumn{1}{c}{\multirow{2}{*}{\textbf{Task}}} & \multicolumn{3}{c}{\textbf{Qwen3.5-9B}} & \multicolumn{3}{c}{\textbf{Qwen3.5-4B}} & \multicolumn{3}{c}{\textbf{\begin{tabular}[c]{@{}c@{}}Qwen3.5-4B\\(streaming)\end{tabular}}} & \multicolumn{3}{>{\columncolor{black!7}}c}{\textbf{\method}}\\
\cmidrule(lr){3-5}\cmidrule(lr){6-8}\cmidrule(lr){9-11}\cmidrule(lr){12-14}
 & & $F_j$ & $C_j$ & $S_j$ & $F_j$ & $C_j$ & $S_j$ & $F_j$ & $C_j$ & $S_j$ & \multicolumn{3}{>{\columncolor{black!7}}c}{\detailours{$F_j$}{$C_j$}{$S_j$}}\\
\midrule
\multirow{2}{*}{\shortstack{C1: Instantaneous\\Conditions}} & T1 Single-Condition Triggering & 83.1 & 82.2 & 68.3 & 0.0 & 0.0 & 0.0 & 2.7 & 50.0 & 1.3 & \multicolumn{3}{>{\columncolor{black!7}}c}{\detailours{97.5}{99.0}{96.5}}\\
 & T2 Multi-Condition Triggering & 31.0 & 98.1 & 30.5 & 16.0 & 98.1 & 15.7 & 3.8 & 0.0 & 0.0 & \multicolumn{3}{>{\columncolor{black!7}}c}{\detailours{93.8}{76.0}{71.4}}\\
\midrule
\multirow{4}{*}{\shortstack{C2: Sequential\\Conditions}} & T3 Consecutive Exceedance & 0.0 & 0.0 & 0.0 & 0.0 & 0.0 & 0.0 & 0.0 & 0.0 & 0.0 & \multicolumn{3}{>{\columncolor{black!7}}c}{\detailours{96.4}{100.0}{96.4}}\\
 & T4 Trend Recognition & 24.0 & 100.0 & 24.0 & 29.9 & 93.0 & 27.8 & 6.7 & 50.0 & 3.3 & \multicolumn{3}{>{\columncolor{black!7}}c}{\detailours{91.6}{96.4}{88.3}}\\
 & T5 Mean Shift & 10.4 & 76.4 & 7.9 & 0.0 & 0.0 & 0.0 & 0.0 & 0.0 & 0.0 & \multicolumn{3}{>{\columncolor{black!7}}c}{\detailours{85.3}{82.7}{70.5}}\\
 & T6 Event Density & 10.7 & 16.7 & 1.8 & 4.4 & 11.0 & 0.5 & 0.0 & 0.0 & 0.0 & \multicolumn{3}{>{\columncolor{black!7}}c}{\detailours{68.3}{97.4}{66.5}}\\
\midrule
\multirow{3}{*}{\shortstack{C3: Event\\Tracking}} & T7 Single-Type Tracking & 0.0 & 0.0 & 0.0 & 0.0 & 0.0 & 0.0 & 8.1 & 0.0 & 0.0 & \multicolumn{3}{>{\columncolor{black!7}}c}{\detailours{83.4}{74.8}{62.4}}\\
 & T8 Multi-Type Tracking & 0.0 & 0.0 & 0.0 & 0.0 & 0.0 & 0.0 & 3.4 & 33.3 & 1.1 & \multicolumn{3}{>{\columncolor{black!7}}c}{\detailours{22.1}{40.3}{8.9}}\\
 & T9 Ordered Event Tracking & 0.0 & 0.0 & 0.0 & 0.0 & 0.0 & 0.0 & 4.0 & 0.0 & 0.0 & \multicolumn{3}{>{\columncolor{black!7}}c}{\detailours{63.8}{85.4}{54.4}}\\
\midrule
\multirow{2}{*}{\shortstack{C4: Event\\Reference}} & T10 Shape Reference & 0.0 & 0.0 & 0.0 & 3.7 & 76.8 & 2.8 & 4.3 & 0.0 & 0.0 & \multicolumn{3}{>{\columncolor{black!7}}c}{\detailours{38.5}{100.0}{38.5}}\\
 & T11 Numerical Reference & 59.6 & 0.0 & 0.0 & 0.0 & 0.0 & 0.0 & 0.0 & 0.0 & 0.0 & \multicolumn{3}{>{\columncolor{black!7}}c}{\detailours{85.5}{82.8}{70.8}}\\
\midrule
\multirow{3}{*}{\shortstack{C5: Online\\Instructions}} & T12 On-Demand Recall & 47.4 & 56.6 & 26.8 & 0.0 & 0.0 & 0.0 & 0.0 & 0.0 & 0.0 & \multicolumn{3}{>{\columncolor{black!7}}c}{\detailours{85.9}{39.3}{33.8}}\\
 & T13 Rule Replacement & 0.0 & 0.0 & 0.0 & 17.9 & 1.7 & 0.3 & 26.9 & 0.0 & 0.0 & \multicolumn{3}{>{\columncolor{black!7}}c}{\detailours{77.9}{76.7}{59.8}}\\
 & T14 Rule Addition & 38.5 & 69.1 & 26.6 & 0.0 & 0.0 & 0.0 & 0.9 & 100.0 & 0.9 & \multicolumn{3}{>{\columncolor{black!7}}c}{\detailours{66.1}{56.9}{37.6}}\\
\midrule
\multirow{4}{*}{\shortstack{C6: Ultra-Long\\Reliability}} & T15 Sustained Vigilance & 56.0 & 0.0 & 0.0 & 0.0 & 0.0 & 0.0 & 0.2 & 50.0 & 0.1 & \multicolumn{3}{>{\columncolor{black!7}}c}{\detailours{99.2}{100.0}{99.2}}\\
 & T16 Long-Gap Reference & 0.0 & 0.0 & 0.0 & 0.0 & 0.0 & 0.0 & 1.0 & 0.0 & 0.0 & \multicolumn{3}{>{\columncolor{black!7}}c}{\detailours{0.6}{100.0}{0.6}}\\
 & T17 Sustained Reporting & 14.6 & 70.0 & 10.2 & 0.0 & 0.0 & 0.0 & 7.6 & 0.0 & 0.0 & \multicolumn{3}{>{\columncolor{black!7}}c}{\detailours{99.4}{49.0}{48.6}}\\
 & T18 Cumulative Evidence & 0.2 & 2.0 & 0.0 & 0.1 & 9.0 & 0.0 & 0.1 & 0.0 & 0.0 & \multicolumn{3}{>{\columncolor{black!7}}c}{\detailours{11.5}{39.7}{4.6}}\\
\midrule
\multicolumn{2}{l}{\textbf{Overall}} & 20.9 & 31.7 & 10.9 & 4.0 & 16.1 & 2.6 & 3.9 & 15.7 & 0.4 & \multicolumn{3}{>{\columncolor{black!7}}c}{\detailours{70.4}{77.6}{56.1}}\\
\bottomrule
\end{tabular*}
\end{minipage}\\[12pt]
\begin{minipage}{\linewidth}
\centering\scriptsize
\vspace{8pt}
\textbf{(d) Time Series Language Models}\par\vspace{3pt}
\renewcommand{\arraystretch}{1.18}
\setlength{\tabcolsep}{0.8pt}
\begin{tabular*}{\linewidth}{@{\extracolsep{\fill}}cl*{9}{r}@{}}
\toprule
\multirow{2}{*}{\textbf{Capability}} & \multicolumn{1}{c}{\multirow{2}{*}{\textbf{Task}}} & \multicolumn{3}{c}{\textbf{TimeOmni-1-4B}} & \multicolumn{3}{c}{\textbf{\begin{tabular}[c]{@{}c@{}}TS-Reasoner-7B\\(streaming)\end{tabular}}} & \multicolumn{3}{>{\columncolor{black!7}}c}{\textbf{\method}}\\
\cmidrule(lr){3-5}\cmidrule(lr){6-8}\cmidrule(lr){9-11}
 & & $F_j$ & $C_j$ & $S_j$ & $F_j$ & $C_j$ & $S_j$ & \multicolumn{3}{>{\columncolor{black!7}}c}{\detailours{$F_j$}{$C_j$}{$S_j$}}\\
\midrule
\multirow{2}{*}{\shortstack{C1: Instantaneous\\Conditions}} & T1 Single-Condition Triggering & 0.0 & 0.0 & 0.0 & 8.4 & 5.2 & 0.4 & \multicolumn{3}{>{\columncolor{black!7}}c}{\detailours{97.5}{99.0}{96.5}}\\
 & T2 Multi-Condition Triggering & 0.0 & 0.0 & 0.0 & 7.8 & 15.0 & 1.2 & \multicolumn{3}{>{\columncolor{black!7}}c}{\detailours{93.8}{76.0}{71.4}}\\
\midrule
\multirow{4}{*}{\shortstack{C2: Sequential\\Conditions}} & T3 Consecutive Exceedance & 0.0 & 0.0 & 0.0 & 8.1 & 6.7 & 0.5 & \multicolumn{3}{>{\columncolor{black!7}}c}{\detailours{96.4}{100.0}{96.4}}\\
 & T4 Trend Recognition & 0.0 & 0.0 & 0.0 & 7.5 & 46.7 & 3.5 & \multicolumn{3}{>{\columncolor{black!7}}c}{\detailours{91.6}{96.4}{88.3}}\\
 & T5 Mean Shift & 0.0 & 0.0 & 0.0 & 4.8 & 40.0 & 1.9 & \multicolumn{3}{>{\columncolor{black!7}}c}{\detailours{85.3}{82.7}{70.5}}\\
 & T6 Event Density & 0.0 & 0.0 & 0.0 & 3.6 & 0.0 & 0.0 & \multicolumn{3}{>{\columncolor{black!7}}c}{\detailours{68.3}{97.4}{66.5}}\\
\midrule
\multirow{3}{*}{\shortstack{C3: Event\\Tracking}} & T7 Single-Type Tracking & 0.0 & 0.0 & 0.0 & 1.1 & 5.0 & 0.1 & \multicolumn{3}{>{\columncolor{black!7}}c}{\detailours{83.4}{74.8}{62.4}}\\
 & T8 Multi-Type Tracking & 0.0 & 0.0 & 0.0 & 3.9 & 10.0 & 0.4 & \multicolumn{3}{>{\columncolor{black!7}}c}{\detailours{22.1}{40.3}{8.9}}\\
 & T9 Ordered Event Tracking & 0.0 & 0.0 & 0.0 & 0.7 & 0.0 & 0.0 & \multicolumn{3}{>{\columncolor{black!7}}c}{\detailours{63.8}{85.4}{54.4}}\\
\midrule
\multirow{2}{*}{\shortstack{C4: Event\\Reference}} & T10 Shape Reference & 0.0 & 0.0 & 0.0 & 4.3 & 6.4 & 0.3 & \multicolumn{3}{>{\columncolor{black!7}}c}{\detailours{38.5}{100.0}{38.5}}\\
 & T11 Numerical Reference & 0.0 & 0.0 & 0.0 & 1.5 & 0.0 & 0.0 & \multicolumn{3}{>{\columncolor{black!7}}c}{\detailours{85.5}{82.8}{70.8}}\\
\midrule
\multirow{3}{*}{\shortstack{C5: Online\\Instructions}} & T12 On-Demand Recall & 3.5 & 0.0 & 0.0 & 1.6 & 0.0 & 0.0 & \multicolumn{3}{>{\columncolor{black!7}}c}{\detailours{85.9}{39.3}{33.8}}\\
 & T13 Rule Replacement & 0.0 & 0.0 & 0.0 & 1.6 & 0.0 & 0.0 & \multicolumn{3}{>{\columncolor{black!7}}c}{\detailours{77.9}{76.7}{59.8}}\\
 & T14 Rule Addition & 33.8 & 47.9 & 16.2 & 5.3 & 22.8 & 1.2 & \multicolumn{3}{>{\columncolor{black!7}}c}{\detailours{66.1}{56.9}{37.6}}\\
\midrule
\multirow{4}{*}{\shortstack{C6: Ultra-Long\\Reliability}} & T15 Sustained Vigilance & 0.0 & 0.0 & 0.0 & 0.2 & 7.3 & 0.0 & \multicolumn{3}{>{\columncolor{black!7}}c}{\detailours{99.2}{100.0}{99.2}}\\
 & T16 Long-Gap Reference & 1.3 & 0.0 & 0.0 & 1.2 & 0.0 & 0.0 & \multicolumn{3}{>{\columncolor{black!7}}c}{\detailours{0.6}{100.0}{0.6}}\\
 & T17 Sustained Reporting & 0.0 & 0.0 & 0.0 & 0.6 & 20.7 & 0.1 & \multicolumn{3}{>{\columncolor{black!7}}c}{\detailours{99.4}{49.0}{48.6}}\\
 & T18 Cumulative Evidence & 0.0 & 0.0 & 0.0 & 0.2 & 0.0 & 0.0 & \multicolumn{3}{>{\columncolor{black!7}}c}{\detailours{11.5}{39.7}{4.6}}\\
\midrule
\multicolumn{2}{l}{\textbf{Overall}} & 2.1 & 2.7 & 0.9 & 3.5 & 10.3 & 0.5 & \multicolumn{3}{>{\columncolor{black!7}}c}{\detailours{70.4}{77.6}{56.1}}\\
\bottomrule
\end{tabular*}
\end{minipage}\\[12pt]
\end{longtable}
\endgroup

\clearpage

\subsection{Ablation Studies}
\label{app:ablation-details}
\label{app:legacy-ablations}
\textbf{Setup.} All variants use the same 180 uniformly sampled ID/Real streams (11.2\% of full evaluation). Encoding variants train for 2,500 steps; memory variants reuse the main checkpoint.

\textbf{Encoding results.} Hybrid encoding scores 85.6/59.0 on ID/Real, compared with 66.0/18.7 for time series embeddings alone and 81.6/57.8 for statistical values alone (Table~\ref{tab:ablation-task-metrics-encoding}).
\begingroup
\newcommand{\detailours}[3]{\makebox[17.5pt][r]{#1}\hspace{4pt}\makebox[17.5pt][r]{#2}\hspace{4pt}\makebox[17.5pt][r]{#3}}
\setlength{\LTleft}{0pt}
\setlength{\LTright}{0pt}
\setlength{\LTcapwidth}{\linewidth}
\begin{longtable}{@{}p{\linewidth}@{}}
\caption{\textbf{Per-task encoding ablations} from Table~\ref{tab:ablation}(a). $F_j$ and $C_j$ are percentages; $S_j=100F_jC_j$. Panels report ID and Real Test; shading marks \method. Overall averages task scores. \textbf{Bold}: the best.}\label{tab:ablation-task-metrics-encoding}\\
\endfirsthead
\multicolumn{1}{c}{\small Table~\thetable: Per-task encoding ablations (continued).}\\[4pt]
\endhead
\endfoot
\endlastfoot
\begin{minipage}{\linewidth}
\centering
\scriptsize
\textbf{(a) ID Test}\par\vspace{3pt}
\renewcommand{\arraystretch}{1.10}
\setlength{\tabcolsep}{0.8pt}
\begin{tabular*}{\linewidth}{@{\extracolsep{\fill}}cl*{9}{r}@{}}
\toprule
\multirow{2}{*}{\textbf{Capability}} & \multicolumn{1}{c}{\multirow{2}{*}{\textbf{Task}}} & \multicolumn{3}{c}{\textbf{\begin{tabular}[c]{@{}c@{}}Time Series\\Embeddings Only\end{tabular}}} & \multicolumn{3}{c}{\textbf{\begin{tabular}[c]{@{}c@{}}Statistical\\Values Only\end{tabular}}} & \multicolumn{3}{>{\columncolor{black!7}}c}{\textbf{\begin{tabular}[c]{@{}c@{}}\method\\(trained for 2,500 steps)\end{tabular}}}\\
\cmidrule(lr){3-5}\cmidrule(lr){6-8}\cmidrule(lr){9-11}
 & & $F_j$ & $C_j$ & $S_j$ & $F_j$ & $C_j$ & $S_j$ & \multicolumn{3}{>{\columncolor{black!7}}c}{\detailours{$F_j$}{$C_j$}{$S_j$}}\\
\midrule
\multirow{2}{*}{\shortstack{C1: Instantaneous\\Conditions}} & T1 Single-Condition Triggering & 82.3 & 17.1 & 14.1 & 100.0 & 100.0 & 100.0 & \multicolumn{3}{>{\columncolor{black!7}}c}{\detailours{100.0}{100.0}{100.0}}\\
 & T2 Multi-Condition Triggering & 73.5 & 12.8 & 9.4 & 95.0 & 75.0 & 71.2 & \multicolumn{3}{>{\columncolor{black!7}}c}{\detailours{88.3}{75.0}{66.3}}\\
\midrule
\multirow{4}{*}{\shortstack{C2: Sequential\\Conditions}} & T3 Consecutive Exceedance & 97.1 & 72.3 & 70.2 & 100.0 & 100.0 & 100.0 & \multicolumn{3}{>{\columncolor{black!7}}c}{\detailours{100.0}{100.0}{100.0}}\\
 & T4 Trend Recognition & 83.3 & 98.9 & 82.4 & 93.3 & 100.0 & 93.3 & \multicolumn{3}{>{\columncolor{black!7}}c}{\detailours{80.0}{100.0}{80.0}}\\
 & T5 Mean Shift & 100.0 & 60.0 & 60.0 & 100.0 & 88.6 & 88.6 & \multicolumn{3}{>{\columncolor{black!7}}c}{\detailours{100.0}{77.2}{77.2}}\\
 & T6 Event Density & 93.3 & 100.0 & 93.3 & 100.0 & 100.0 & 100.0 & \multicolumn{3}{>{\columncolor{black!7}}c}{\detailours{100.0}{100.0}{100.0}}\\
\midrule
\multirow{3}{*}{\shortstack{C3: Event\\Tracking}} & T7 Single-Type Tracking & 100.0 & 100.0 & 100.0 & 100.0 & 100.0 & 100.0 & \multicolumn{3}{>{\columncolor{black!7}}c}{\detailours{100.0}{100.0}{100.0}}\\
 & T8 Multi-Type Tracking & 96.0 & 100.0 & 96.0 & 78.5 & 85.0 & 66.8 & \multicolumn{3}{>{\columncolor{black!7}}c}{\detailours{93.8}{100.0}{93.8}}\\
 & T9 Ordered Event Tracking & 100.0 & 100.0 & 100.0 & 86.7 & 100.0 & 86.7 & \multicolumn{3}{>{\columncolor{black!7}}c}{\detailours{100.0}{100.0}{100.0}}\\
\midrule
\multirow{2}{*}{\shortstack{C4: Event\\Reference}} & T10 Shape Reference & 100.0 & 100.0 & 100.0 & 55.1 & 100.0 & 55.1 & \multicolumn{3}{>{\columncolor{black!7}}c}{\detailours{100.0}{100.0}{100.0}}\\
 & T11 Numerical Reference & 93.1 & 47.8 & 44.5 & 100.0 & 96.8 & 96.8 & \multicolumn{3}{>{\columncolor{black!7}}c}{\detailours{100.0}{87.5}{87.5}}\\
\midrule
\multirow{3}{*}{\shortstack{C5: Online\\Instructions}} & T12 On-Demand Recall & 100.0 & 93.3 & 93.3 & 100.0 & 85.7 & 85.7 & \multicolumn{3}{>{\columncolor{black!7}}c}{\detailours{100.0}{84.0}{84.0}}\\
 & T13 Rule Replacement & 78.8 & 100.0 & 78.8 & 100.0 & 100.0 & 100.0 & \multicolumn{3}{>{\columncolor{black!7}}c}{\detailours{80.0}{100.0}{80.0}}\\
 & T14 Rule Addition & 89.7 & 56.0 & 50.2 & 98.7 & 65.2 & 64.3 & \multicolumn{3}{>{\columncolor{black!7}}c}{\detailours{96.0}{69.3}{66.5}}\\
\midrule
\multirow{4}{*}{\shortstack{C6: Ultra-Long\\Reliability}} & T15 Sustained Vigilance & 71.3 & 1.3 & 0.9 & 100.0 & 100.0 & 100.0 & \multicolumn{3}{>{\columncolor{black!7}}c}{\detailours{100.0}{100.0}{100.0}}\\
 & T16 Long-Gap Reference & 100.0 & 100.0 & 100.0 & 71.6 & 100.0 & 71.6 & \multicolumn{3}{>{\columncolor{black!7}}c}{\detailours{100.0}{100.0}{100.0}}\\
 & T17 Sustained Reporting & 100.0 & 94.7 & 94.7 & 100.0 & 80.5 & 80.5 & \multicolumn{3}{>{\columncolor{black!7}}c}{\detailours{100.0}{97.1}{97.1}}\\
 & T18 Cumulative Evidence & 0.0 & 0.0 & 0.0 & 20.0 & 40.8 & 8.2 & \multicolumn{3}{>{\columncolor{black!7}}c}{\detailours{20.0}{40.8}{8.2}}\\
\midrule
\multicolumn{2}{l}{\textbf{Overall}} & 86.6 & 69.7 & 66.0 & 88.8 & 89.9 & 81.6 & \multicolumn{3}{>{\columncolor{black!7}}c}{\detailours{\textbf{92.1}}{\textbf{90.6}}{\textbf{85.6}}}\\
\bottomrule
\end{tabular*}
\end{minipage}\\[4pt]
\begin{minipage}{\linewidth}
\centering
\scriptsize
\vspace{5pt}
\textbf{(b) Real Test}\par\vspace{3pt}
\renewcommand{\arraystretch}{1.10}
\setlength{\tabcolsep}{0.8pt}
\begin{tabular*}{\linewidth}{@{\extracolsep{\fill}}cl*{9}{r}@{}}
\toprule
\multirow{2}{*}{\textbf{Capability}} & \multicolumn{1}{c}{\multirow{2}{*}{\textbf{Task}}} & \multicolumn{3}{c}{\textbf{\begin{tabular}[c]{@{}c@{}}Time Series\\Embeddings Only\end{tabular}}} & \multicolumn{3}{c}{\textbf{\begin{tabular}[c]{@{}c@{}}Statistical\\Values Only\end{tabular}}} & \multicolumn{3}{>{\columncolor{black!7}}c}{\textbf{\begin{tabular}[c]{@{}c@{}}\method\\(trained for 2,500 steps)\end{tabular}}}\\
\cmidrule(lr){3-5}\cmidrule(lr){6-8}\cmidrule(lr){9-11}
 & & $F_j$ & $C_j$ & $S_j$ & $F_j$ & $C_j$ & $S_j$ & \multicolumn{3}{>{\columncolor{black!7}}c}{\detailours{$F_j$}{$C_j$}{$S_j$}}\\
\midrule
\multirow{2}{*}{\shortstack{C1: Instantaneous\\Conditions}} & T1 Single-Condition Triggering & 51.0 & 32.7 & 16.7 & 100.0 & 100.0 & 100.0 & \multicolumn{3}{>{\columncolor{black!7}}c}{\detailours{100.0}{100.0}{100.0}}\\
 & T2 Multi-Condition Triggering & 93.8 & 14.1 & 13.2 & 67.0 & 74.3 & 49.8 & \multicolumn{3}{>{\columncolor{black!7}}c}{\detailours{91.0}{76.8}{69.9}}\\
\midrule
\multirow{4}{*}{\shortstack{C2: Sequential\\Conditions}} & T3 Consecutive Exceedance & 61.0 & 10.0 & 6.1 & 100.0 & 100.0 & 100.0 & \multicolumn{3}{>{\columncolor{black!7}}c}{\detailours{93.3}{100.0}{93.3}}\\
 & T4 Trend Recognition & 51.2 & 86.2 & 44.1 & 89.3 & 100.0 & 89.3 & \multicolumn{3}{>{\columncolor{black!7}}c}{\detailours{82.7}{100.0}{82.7}}\\
 & T5 Mean Shift & 33.3 & 50.0 & 16.7 & 80.0 & 80.1 & 64.1 & \multicolumn{3}{>{\columncolor{black!7}}c}{\detailours{73.3}{80.9}{59.3}}\\
 & T6 Event Density & 60.0 & 25.0 & 15.0 & 60.0 & 66.7 & 40.0 & \multicolumn{3}{>{\columncolor{black!7}}c}{\detailours{73.3}{100.0}{73.3}}\\
\midrule
\multirow{3}{*}{\shortstack{C3: Event\\Tracking}} & T7 Single-Type Tracking & 8.3 & 6.7 & 0.6 & 53.1 & 50.0 & 26.6 & \multicolumn{3}{>{\columncolor{black!7}}c}{\detailours{73.7}{70.0}{51.6}}\\
 & T8 Multi-Type Tracking & 31.3 & 50.0 & 15.7 & 22.5 & 41.7 & 9.4 & \multicolumn{3}{>{\columncolor{black!7}}c}{\detailours{34.2}{47.9}{16.4}}\\
 & T9 Ordered Event Tracking & 47.1 & 87.5 & 41.2 & 80.5 & 100.0 & 80.5 & \multicolumn{3}{>{\columncolor{black!7}}c}{\detailours{71.3}{95.3}{68.0}}\\
\midrule
\multirow{2}{*}{\shortstack{C4: Event\\Reference}} & T10 Shape Reference & 45.6 & 100.0 & 45.6 & 17.1 & 100.0 & 17.1 & \multicolumn{3}{>{\columncolor{black!7}}c}{\detailours{28.8}{100.0}{28.8}}\\
 & T11 Numerical Reference & 56.7 & 7.7 & 4.4 & 100.0 & 96.6 & 96.6 & \multicolumn{3}{>{\columncolor{black!7}}c}{\detailours{80.2}{98.5}{79.0}}\\
\midrule
\multirow{3}{*}{\shortstack{C5: Online\\Instructions}} & T12 On-Demand Recall & 80.0 & 50.0 & 40.0 & 100.0 & 87.2 & 87.2 & \multicolumn{3}{>{\columncolor{black!7}}c}{\detailours{80.2}{66.7}{53.5}}\\
 & T13 Rule Replacement & 35.6 & 42.6 & 15.2 & 84.2 & 57.3 & 48.3 & \multicolumn{3}{>{\columncolor{black!7}}c}{\detailours{97.1}{86.7}{84.2}}\\
 & T14 Rule Addition & 26.7 & 36.0 & 9.6 & 91.0 & 46.9 & 42.6 & \multicolumn{3}{>{\columncolor{black!7}}c}{\detailours{60.4}{44.3}{26.8}}\\
\midrule
\multirow{4}{*}{\shortstack{C6: Ultra-Long\\Reliability}} & T15 Sustained Vigilance & 23.3 & 10.0 & 2.3 & 100.0 & 100.0 & 100.0 & \multicolumn{3}{>{\columncolor{black!7}}c}{\detailours{100.0}{100.0}{100.0}}\\
 & T16 Long-Gap Reference & 0.0 & 0.0 & 0.0 & 4.7 & 100.0 & 4.7 & \multicolumn{3}{>{\columncolor{black!7}}c}{\detailours{5.1}{100.0}{5.1}}\\
 & T17 Sustained Reporting & 98.2 & 51.6 & 50.7 & 98.2 & 60.4 & 59.3 & \multicolumn{3}{>{\columncolor{black!7}}c}{\detailours{98.2}{61.6}{60.4}}\\
 & T18 Cumulative Evidence & 0.0 & 0.0 & 0.0 & 60.0 & 40.6 & 24.3 & \multicolumn{3}{>{\columncolor{black!7}}c}{\detailours{20.0}{50.0}{10.0}}\\
\midrule
\multicolumn{2}{l}{\textbf{Overall}} & 44.6 & 36.7 & 18.7 & \textbf{72.6} & 77.9 & 57.8 & \multicolumn{3}{>{\columncolor{black!7}}c}{\detailours{70.2}{\textbf{82.1}}{\textbf{59.0}}}\\
\bottomrule
\end{tabular*}
\end{minipage}\\[4pt]
\end{longtable}
\endgroup

\clearpage
\textbf{Memory results.} Removing text memory lowers the ID score from 89.8 to 59.1; removing historical waveforms lowers C4 from 93.8 to 65.7. Removing uncommitted context or delaying memory reduces the ID score to 71.9 or 64.8 (Table~\ref{tab:ablation-task-metrics-memory}).
\begingroup
\newcommand{\detailours}[3]{\makebox[16.5pt][r]{#1}\hspace{2.5pt}\makebox[16.5pt][r]{#2}\hspace{2.5pt}\makebox[16.5pt][r]{#3}}
\setlength{\LTleft}{0pt}
\setlength{\LTright}{0pt}
\setlength{\LTcapwidth}{\linewidth}
\begin{longtable}{@{}p{\linewidth}@{}}
\caption{\textbf{Per-task memory ablations} from Table~\ref{tab:ablation}(b). $F_j$ and $C_j$ are percentages; $S_j=100F_jC_j$. Panels report ID and Real Test; shading marks \method. Overall averages task scores. \textbf{Bold}: the best.}\label{tab:ablation-task-metrics-memory}\\
\endfirsthead
\multicolumn{1}{c}{\small Table~\thetable: Per-task memory ablations (continued).}\\[4pt]
\endhead
\endfoot
\endlastfoot
\begin{minipage}{\linewidth}
\centering
\fontsize{6.5}{7.8}\selectfont
\textbf{(a) ID Test}\par\vspace{3pt}
\renewcommand{\arraystretch}{1.10}
\setlength{\tabcolsep}{0.35pt}
\begin{tabular*}{\linewidth}{@{\extracolsep{\fill}}cl*{15}{r}@{}}
\toprule
\multirow{2}{*}{\textbf{Capability}} & \multicolumn{1}{c}{\multirow{2}{*}{\textbf{Task}}} & \multicolumn{3}{c}{\textbf{\begin{tabular}[c]{@{}c@{}}w/o Text\\Memory\end{tabular}}} & \multicolumn{3}{c}{\textbf{\begin{tabular}[c]{@{}c@{}}w/o Waveform\\Memory\end{tabular}}} & \multicolumn{3}{c}{\textbf{\begin{tabular}[c]{@{}c@{}}w/o Uncommitted\\Context\end{tabular}}} & \multicolumn{3}{c}{\textbf{\begin{tabular}[c]{@{}c@{}}Memory Delayed\\by One Window\end{tabular}}} & \multicolumn{3}{>{\columncolor{black!7}}c}{\textbf{\method}}\\
\cmidrule(lr){3-5}\cmidrule(lr){6-8}\cmidrule(lr){9-11}\cmidrule(lr){12-14}\cmidrule(lr){15-17}
 & & $F_j$ & $C_j$ & $S_j$ & $F_j$ & $C_j$ & $S_j$ & $F_j$ & $C_j$ & $S_j$ & $F_j$ & $C_j$ & $S_j$ & \multicolumn{3}{>{\columncolor{black!7}}c}{\detailours{$F_j$}{$C_j$}{$S_j$}}\\
\midrule
\multirow{2}{*}{\shortstack{C1: Instantaneous\\Conditions}} & T1 Single-Condition Triggering & 100.0 & 100.0 & 100.0 & 100.0 & 100.0 & 100.0 & 100.0 & 100.0 & 100.0 & 100.0 & 100.0 & 100.0 & \multicolumn{3}{>{\columncolor{black!7}}c}{\detailours{100.0}{100.0}{100.0}}\\
 & T2 Multi-Condition Triggering & 96.0 & 79.4 & 76.2 & 96.0 & 79.4 & 76.2 & 96.0 & 79.4 & 76.2 & 96.0 & 79.4 & 76.2 & \multicolumn{3}{>{\columncolor{black!7}}c}{\detailours{96.0}{79.4}{76.2}}\\
\midrule
\multirow{4}{*}{\shortstack{C2: Sequential\\Conditions}} & T3 Consecutive Exceedance & 100.0 & 100.0 & 100.0 & 100.0 & 100.0 & 100.0 & 100.0 & 100.0 & 100.0 & 100.0 & 100.0 & 100.0 & \multicolumn{3}{>{\columncolor{black!7}}c}{\detailours{100.0}{100.0}{100.0}}\\
 & T4 Trend Recognition & 93.3 & 100.0 & 93.3 & 93.3 & 100.0 & 93.3 & 93.3 & 100.0 & 93.3 & 93.3 & 100.0 & 93.3 & \multicolumn{3}{>{\columncolor{black!7}}c}{\detailours{93.3}{100.0}{93.3}}\\
 & T5 Mean Shift & 100.0 & 79.5 & 79.5 & 100.0 & 80.2 & 80.2 & 100.0 & 79.5 & 79.5 & 100.0 & 79.5 & 79.5 & \multicolumn{3}{>{\columncolor{black!7}}c}{\detailours{100.0}{79.5}{79.5}}\\
 & T6 Event Density & 100.0 & 100.0 & 100.0 & 100.0 & 100.0 & 100.0 & 100.0 & 100.0 & 100.0 & 100.0 & 100.0 & 100.0 & \multicolumn{3}{>{\columncolor{black!7}}c}{\detailours{100.0}{100.0}{100.0}}\\
\midrule
\multirow{3}{*}{\shortstack{C3: Event\\Tracking}} & T7 Single-Type Tracking & 28.0 & 75.0 & 21.0 & 100.0 & 100.0 & 100.0 & 80.5 & 70.0 & 56.3 & 48.0 & 70.8 & 34.0 & \multicolumn{3}{>{\columncolor{black!7}}c}{\detailours{100.0}{100.0}{100.0}}\\
 & T8 Multi-Type Tracking & 0.0 & 0.0 & 0.0 & 100.0 & 100.0 & 100.0 & 83.8 & 57.5 & 48.2 & 63.8 & 43.8 & 27.9 & \multicolumn{3}{>{\columncolor{black!7}}c}{\detailours{100.0}{100.0}{100.0}}\\
 & T9 Ordered Event Tracking & 70.5 & 100.0 & 70.5 & 86.7 & 100.0 & 86.7 & 59.3 & 100.0 & 59.3 & 66.0 & 100.0 & 66.0 & \multicolumn{3}{>{\columncolor{black!7}}c}{\detailours{86.7}{100.0}{86.7}}\\
\midrule
\multirow{2}{*}{\shortstack{C4: Event\\Reference}} & T10 Shape Reference & 100.0 & 100.0 & 100.0 & 45.7 & 100.0 & 45.7 & 100.0 & 100.0 & 100.0 & 100.0 & 100.0 & 100.0 & \multicolumn{3}{>{\columncolor{black!7}}c}{\detailours{100.0}{100.0}{100.0}}\\
 & T11 Numerical Reference & 16.0 & 75.0 & 12.0 & 100.0 & 85.6 & 85.6 & 90.0 & 54.5 & 49.1 & 85.3 & 43.7 & 37.3 & \multicolumn{3}{>{\columncolor{black!7}}c}{\detailours{100.0}{87.7}{87.7}}\\
\midrule
\multirow{3}{*}{\shortstack{C5: Online\\Instructions}} & T12 On-Demand Recall & 100.0 & 33.3 & 33.3 & 100.0 & 90.7 & 90.7 & 100.0 & 50.0 & 50.0 & 100.0 & 40.0 & 40.0 & \multicolumn{3}{>{\columncolor{black!7}}c}{\detailours{100.0}{90.7}{90.7}}\\
 & T13 Rule Replacement & 25.7 & 66.7 & 17.1 & 80.0 & 100.0 & 80.0 & 74.7 & 51.8 & 38.7 & 48.8 & 34.4 & 16.8 & \multicolumn{3}{>{\columncolor{black!7}}c}{\detailours{80.0}{100.0}{80.0}}\\
 & T14 Rule Addition & 88.2 & 54.5 & 48.1 & 98.8 & 68.3 & 67.5 & 96.7 & 58.9 & 57.0 & 96.2 & 64.9 & 62.4 & \multicolumn{3}{>{\columncolor{black!7}}c}{\detailours{98.8}{68.3}{67.5}}\\
\midrule
\multirow{4}{*}{\shortstack{C6: Ultra-Long\\Reliability}} & T15 Sustained Vigilance & 100.0 & 100.0 & 100.0 & 100.0 & 100.0 & 100.0 & 100.0 & 100.0 & 100.0 & 100.0 & 100.0 & 100.0 & \multicolumn{3}{>{\columncolor{black!7}}c}{\detailours{100.0}{100.0}{100.0}}\\
 & T16 Long-Gap Reference & 100.0 & 100.0 & 100.0 & 21.0 & 100.0 & 21.0 & 100.0 & 100.0 & 100.0 & 100.0 & 100.0 & 100.0 & \multicolumn{3}{>{\columncolor{black!7}}c}{\detailours{100.0}{100.0}{100.0}}\\
 & T17 Sustained Reporting & 98.5 & 12.5 & 12.3 & 100.0 & 94.4 & 94.4 & 100.0 & 40.6 & 40.6 & 100.0 & 29.2 & 29.2 & \multicolumn{3}{>{\columncolor{black!7}}c}{\detailours{100.0}{94.4}{94.4}}\\
 & T18 Cumulative Evidence & 0.0 & 0.0 & 0.0 & 100.0 & 60.3 & 60.3 & 73.3 & 61.9 & 45.4 & 13.3 & 24.6 & 3.3 & \multicolumn{3}{>{\columncolor{black!7}}c}{\detailours{100.0}{60.3}{60.3}}\\
\midrule
\multicolumn{2}{l}{\textbf{Overall}} & 73.1 & 70.9 & 59.1 & 90.1 & 92.2 & 82.3 & 91.5 & 78.0 & 71.9 & 83.9 & 72.8 & 64.8 & \multicolumn{3}{>{\columncolor{black!7}}c}{\detailours{\textbf{97.5}}{\textbf{92.2}}{\textbf{89.8}}}\\
\bottomrule
\end{tabular*}
\end{minipage}\\[4pt]
\begin{minipage}{\linewidth}
\centering
\fontsize{6.5}{7.8}\selectfont
\vspace{5pt}
\textbf{(b) Real Test}\par\vspace{3pt}
\renewcommand{\arraystretch}{1.10}
\setlength{\tabcolsep}{0.35pt}
\begin{tabular*}{\linewidth}{@{\extracolsep{\fill}}cl*{15}{r}@{}}
\toprule
\multirow{2}{*}{\textbf{Capability}} & \multicolumn{1}{c}{\multirow{2}{*}{\textbf{Task}}} & \multicolumn{3}{c}{\textbf{\begin{tabular}[c]{@{}c@{}}w/o Text\\Memory\end{tabular}}} & \multicolumn{3}{c}{\textbf{\begin{tabular}[c]{@{}c@{}}w/o Waveform\\Memory\end{tabular}}} & \multicolumn{3}{c}{\textbf{\begin{tabular}[c]{@{}c@{}}w/o Uncommitted\\Context\end{tabular}}} & \multicolumn{3}{c}{\textbf{\begin{tabular}[c]{@{}c@{}}Memory Delayed\\by One Window\end{tabular}}} & \multicolumn{3}{>{\columncolor{black!7}}c}{\textbf{\method}}\\
\cmidrule(lr){3-5}\cmidrule(lr){6-8}\cmidrule(lr){9-11}\cmidrule(lr){12-14}\cmidrule(lr){15-17}
 & & $F_j$ & $C_j$ & $S_j$ & $F_j$ & $C_j$ & $S_j$ & $F_j$ & $C_j$ & $S_j$ & $F_j$ & $C_j$ & $S_j$ & \multicolumn{3}{>{\columncolor{black!7}}c}{\detailours{$F_j$}{$C_j$}{$S_j$}}\\
\midrule
\multirow{2}{*}{\shortstack{C1: Instantaneous\\Conditions}} & T1 Single-Condition Triggering & 100.0 & 100.0 & 100.0 & 100.0 & 100.0 & 100.0 & 100.0 & 100.0 & 100.0 & 100.0 & 100.0 & 100.0 & \multicolumn{3}{>{\columncolor{black!7}}c}{\detailours{100.0}{100.0}{100.0}}\\
 & T2 Multi-Condition Triggering & 97.1 & 71.9 & 69.9 & 97.1 & 71.9 & 69.9 & 97.1 & 71.9 & 69.9 & 97.1 & 71.9 & 69.9 & \multicolumn{3}{>{\columncolor{black!7}}c}{\detailours{97.1}{71.9}{69.9}}\\
\midrule
\multirow{4}{*}{\shortstack{C2: Sequential\\Conditions}} & T3 Consecutive Exceedance & 100.0 & 100.0 & 100.0 & 100.0 & 100.0 & 100.0 & 100.0 & 100.0 & 100.0 & 100.0 & 100.0 & 100.0 & \multicolumn{3}{>{\columncolor{black!7}}c}{\detailours{100.0}{100.0}{100.0}}\\
 & T4 Trend Recognition & 76.0 & 100.0 & 76.0 & 76.0 & 100.0 & 76.0 & 76.0 & 100.0 & 76.0 & 76.0 & 100.0 & 76.0 & \multicolumn{3}{>{\columncolor{black!7}}c}{\detailours{76.0}{100.0}{76.0}}\\
 & T5 Mean Shift & 80.0 & 81.9 & 65.6 & 80.0 & 81.9 & 65.6 & 80.0 & 81.9 & 65.6 & 80.0 & 81.9 & 65.6 & \multicolumn{3}{>{\columncolor{black!7}}c}{\detailours{80.0}{81.9}{65.6}}\\
 & T6 Event Density & 66.7 & 100.0 & 66.7 & 66.7 & 100.0 & 66.7 & 66.7 & 100.0 & 66.7 & 66.7 & 100.0 & 66.7 & \multicolumn{3}{>{\columncolor{black!7}}c}{\detailours{66.7}{100.0}{66.7}}\\
\midrule
\multirow{3}{*}{\shortstack{C3: Event\\Tracking}} & T7 Single-Type Tracking & 10.3 & 25.0 & 2.6 & 67.0 & 73.3 & 49.1 & 36.4 & 54.2 & 19.7 & 26.4 & 50.0 & 13.2 & \multicolumn{3}{>{\columncolor{black!7}}c}{\detailours{67.0}{73.3}{49.1}}\\
 & T8 Multi-Type Tracking & 35.8 & 48.6 & 17.4 & 40.9 & 39.6 & 16.2 & 29.9 & 44.4 & 13.3 & 38.0 & 61.1 & 23.2 & \multicolumn{3}{>{\columncolor{black!7}}c}{\detailours{40.9}{39.6}{16.2}}\\
 & T9 Ordered Event Tracking & 63.3 & 100.0 & 63.3 & 69.8 & 100.0 & 69.8 & 47.2 & 91.7 & 43.3 & 71.3 & 100.0 & 71.3 & \multicolumn{3}{>{\columncolor{black!7}}c}{\detailours{69.8}{100.0}{69.8}}\\
\midrule
\multirow{2}{*}{\shortstack{C4: Event\\Reference}} & T10 Shape Reference & 23.6 & 100.0 & 23.6 & 28.0 & 100.0 & 28.0 & 36.0 & 100.0 & 36.0 & 28.1 & 100.0 & 28.1 & \multicolumn{3}{>{\columncolor{black!7}}c}{\detailours{36.1}{100.0}{36.1}}\\
 & T11 Numerical Reference & 58.3 & 57.8 & 33.7 & 80.2 & 85.1 & 68.3 & 69.6 & 86.3 & 60.0 & 68.3 & 62.1 & 42.4 & \multicolumn{3}{>{\columncolor{black!7}}c}{\detailours{80.2}{85.1}{68.3}}\\
\midrule
\multirow{3}{*}{\shortstack{C5: Online\\Instructions}} & T12 On-Demand Recall & 80.4 & 31.7 & 25.5 & 80.4 & 67.2 & 54.0 & 80.4 & 26.7 & 21.4 & 80.4 & 46.7 & 37.5 & \multicolumn{3}{>{\columncolor{black!7}}c}{\detailours{80.4}{67.2}{54.0}}\\
 & T13 Rule Replacement & 35.0 & 75.5 & 26.4 & 79.0 & 100.0 & 79.0 & 65.3 & 52.5 & 34.3 & 48.9 & 63.0 & 30.8 & \multicolumn{3}{>{\columncolor{black!7}}c}{\detailours{79.0}{100.0}{79.0}}\\
 & T14 Rule Addition & 73.8 & 30.5 & 22.5 & 55.4 & 45.6 & 25.3 & 63.0 & 35.8 & 22.6 & 64.7 & 45.6 & 29.6 & \multicolumn{3}{>{\columncolor{black!7}}c}{\detailours{55.4}{45.6}{25.3}}\\
\midrule
\multirow{4}{*}{\shortstack{C6: Ultra-Long\\Reliability}} & T15 Sustained Vigilance & 100.0 & 100.0 & 100.0 & 100.0 & 100.0 & 100.0 & 100.0 & 100.0 & 100.0 & 100.0 & 100.0 & 100.0 & \multicolumn{3}{>{\columncolor{black!7}}c}{\detailours{100.0}{100.0}{100.0}}\\
 & T16 Long-Gap Reference & 2.6 & 100.0 & 2.6 & 4.0 & 100.0 & 4.0 & 1.0 & 100.0 & 1.0 & 1.9 & 100.0 & 1.9 & \multicolumn{3}{>{\columncolor{black!7}}c}{\detailours{1.3}{100.0}{1.3}}\\
 & T17 Sustained Reporting & 92.1 & 52.0 & 47.9 & 98.2 & 47.6 & 46.7 & 98.2 & 48.4 & 47.6 & 98.2 & 48.2 & 47.3 & \multicolumn{3}{>{\columncolor{black!7}}c}{\detailours{98.2}{47.6}{46.7}}\\
 & T18 Cumulative Evidence & 0.0 & 0.0 & 0.0 & 20.0 & 50.0 & 10.0 & 33.3 & 42.5 & 14.2 & 13.3 & 34.2 & 4.6 & \multicolumn{3}{>{\columncolor{black!7}}c}{\detailours{20.0}{50.0}{10.0}}\\
\midrule
\multicolumn{2}{l}{\textbf{Overall}} & 60.8 & 70.8 & 46.9 & 69.0 & \textbf{81.2} & 57.1 & 65.6 & 74.2 & 49.5 & 64.4 & 75.8 & 50.4 & \multicolumn{3}{>{\columncolor{black!7}}c}{\detailours{\textbf{69.3}}{\textbf{81.2}}{\textbf{57.4}}}\\
\bottomrule
\end{tabular*}
\end{minipage}\\[4pt]
\end{longtable}
\endgroup

\clearpage

\subsection{Backbone Analysis}
\label{app:backbone}
\textbf{Setup.} Backbones share the training data, encoder, memory settings, and optimization recipe, and use the same 180 ID/Real test streams. Qwen3.5 uses its native tokenizer and 100 warm-up steps; other backbones use 200. Figure~\ref{fig:backbone-analysis} shows 95\% bootstrap intervals from 2,000 within-task resamples.

\begin{figure*}[!htbp]
\centering
\includegraphics[width=\textwidth]{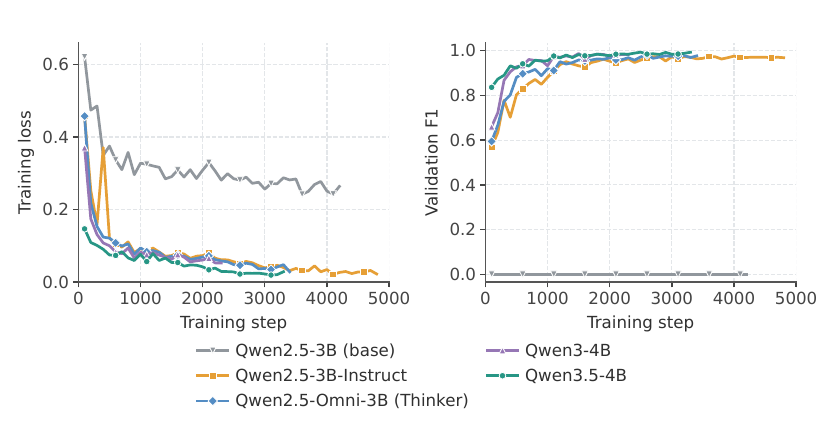}
\caption{\textbf{Backbone convergence.} Training loss (left) and validation response-decision F1 (right) during initial training. Validation uses teacher forcing and is distinct from streaming Timing F1.}
\label{fig:backbone-convergence}
\end{figure*}

\textbf{Results.} Training loss and validation F1 stabilize by the end of training (Figure~\ref{fig:backbone-convergence}). Qwen3.5-4B achieves the highest overall ID and Real scores (Table~\ref{tab:backbone}). The base Qwen2.5-3B model's zero scores reflect tied response-control-token logits in its frozen output head.

\begin{table*}[!htbp]
\centering
\caption{Backbone comparison by capability on 180 uniformly sampled ID/Real streams (11.2\% of full evaluation; Joint Score $S$). \textbf{Bold}: best.}
\label{tab:backbone}
\begingroup\fontsize{7.5}{9}\selectfont
\renewcommand{\arraystretch}{1.15}
\setlength{\tabcolsep}{2pt}
\begin{tabular}{@{}>{\raggedright\arraybackslash}p{72pt}*{14}{c}@{}}
\toprule
\multirow{2}{*}[-0.5ex]{\textbf{Backbone}} & \multicolumn{2}{c}{\textbf{\begin{tabular}[c]{@{}c@{}}C1: Instantaneous\\Conditions\end{tabular}}} & \multicolumn{2}{c}{\textbf{\begin{tabular}[c]{@{}c@{}}C2: Sequential\\Conditions\end{tabular}}} & \multicolumn{2}{c}{\textbf{\begin{tabular}[c]{@{}c@{}}C3: Event\\Tracking\end{tabular}}} & \multicolumn{2}{c}{\textbf{\begin{tabular}[c]{@{}c@{}}C4: Event\\Reference\end{tabular}}} & \multicolumn{2}{c}{\textbf{\begin{tabular}[c]{@{}c@{}}C5: Online\\Instructions\end{tabular}}} & \multicolumn{2}{c}{\textbf{\begin{tabular}[c]{@{}c@{}}C6: Ultra-Long\\Reliability\end{tabular}}} & \multicolumn{2}{c}{\textbf{Overall}}\\
\cmidrule(lr){2-3}\cmidrule(lr){4-5}\cmidrule(lr){6-7}\cmidrule(lr){8-9}\cmidrule(lr){10-11}\cmidrule(lr){12-13}\cmidrule(lr){14-15}
 & ID & Real & ID & Real & ID & Real & ID & Real & ID & Real & ID & Real & ID & Real\\
\midrule
Qwen2.5-3B (base) & 0.0 & 0.0 & 0.0 & 0.0 & 0.0 & 0.0 & 0.0 & 0.0 & 0.0 & 0.0 & 0.0 & 0.0 & 0.0 & 0.0\\
Qwen2.5-3B-Instruct & 83.2 & 84.8 & 86.3 & 61.4 & 74.3 & 33.1 & 86.8 & 43.9 & 65.5 & 36.1 & 72.8 & 37.3 & 77.5 & 47.8\\
Qwen2.5-Omni-3B (Thinker) & 85.3 & 81.8 & 92.3 & 69.8 & 82.6 & 25.8 & \textbf{96.2} & 49.0 & 58.3 & 31.1 & 82.1 & \textbf{39.9} & 82.4 & 48.4\\
Qwen3-4B & 86.0 & 75.4 & 90.0 & 71.9 & 91.6 & 33.1 & 91.1 & \textbf{58.4} & \textbf{84.3} & \textbf{55.1} & 79.1 & 38.5 & 86.6 & 54.1\\
\rowcolor{blue!5}
Qwen3.5-4B & \textbf{88.1} & \textbf{84.9} & \textbf{93.2} & \textbf{77.1} & \textbf{95.6} & \textbf{45.0} & 93.8 & 52.2 & 79.4 & 52.8 & \textbf{88.7} & 39.5 & \textbf{89.8} & \textbf{57.4}\\
\bottomrule
\end{tabular}
\endgroup
\end{table*}

\subsection{Generalization}
\label{app:generalization-details}
\subsubsection{Instruction Paraphrasing}
\label{app:paraphrase}
\textbf{Setup.} Table~\ref{tab:paraphrase} evaluates the same 180 ID streams before and after instruction paraphrasing, keeping signals, reference answers, arrival times, programs, and checkpoints fixed. DeepSeek-V4.1-Flash rewrites instructions while preserving numerical values and task requirements, with a separate model call checking equivalence. It rewrites 242 of 243 instructions, updates, and queries; opening-instruction overlap is low (\textbf{0.26} BLEU-4 and \textbf{0.36} token Jaccard), indicating substantial rewording.

\textbf{Results.} \method changes from 89.5 to 88.5, while the five code baselines with original scores above 40 lose 17.6--32.1 points. The 95\% bootstrap interval for our paired difference is $[-3.1,+1.1]$ (2,000 within-task resamples). Together with the low lexical overlap, this supports stable generalization to differently worded user instructions.

\begin{table*}[!htbp]
\caption{Instruction paraphrasing (IP; Joint Score $S$) by capability on paired ID streams before (ID) and after instruction paraphrasing (IP). Overall averages all 18 tasks. \textbf{Bold}: best.}
\label{tab:paraphrase}\label{tab:paraphrase-baselines}
\centering\begingroup\fontsize{7.5}{9}\selectfont
\renewcommand{\arraystretch}{1.15}
\setlength{\tabcolsep}{2pt}
\begin{tabular}{@{}>{\raggedright\arraybackslash}p{72pt}*{14}{c}@{}}\toprule
\multirow{2}{*}[-0.5ex]{\textbf{Method}} & \multicolumn{2}{c}{\textbf{\begin{tabular}[c]{@{}c@{}}C1: Instantaneous\\Conditions\end{tabular}}} & \multicolumn{2}{c}{\textbf{\begin{tabular}[c]{@{}c@{}}C2: Sequential\\Conditions\end{tabular}}} & \multicolumn{2}{c}{\textbf{\begin{tabular}[c]{@{}c@{}}C3: Event\\Tracking\end{tabular}}} & \multicolumn{2}{c}{\textbf{\begin{tabular}[c]{@{}c@{}}C4: Event\\Reference\end{tabular}}} & \multicolumn{2}{c}{\textbf{\begin{tabular}[c]{@{}c@{}}C5: Online\\Instructions\end{tabular}}} & \multicolumn{2}{c}{\textbf{\begin{tabular}[c]{@{}c@{}}C6: Ultra-Long\\Reliability\end{tabular}}} & \multicolumn{2}{c}{\textbf{Overall}}\\
\cmidrule(lr){2-3}\cmidrule(lr){4-5}\cmidrule(lr){6-7}\cmidrule(lr){8-9}\cmidrule(lr){10-11}\cmidrule(lr){12-13}\cmidrule(lr){14-15}
 & ID & IP & ID & IP & ID & IP & ID & IP & ID & IP & ID & IP & ID & IP\\\midrule
\multicolumn{15}{@{}l@{}}{\textbf{Proprietary Models}}\\
Opus 5 & 95.0 & \textbf{83.7} & \textbf{95.9} & 89.6 & 84.2 & 44.8 & 69.2 & 31.3 & 50.2 & 55.0 & 72.5 & 49.9 & 78.1 & 60.4\\
Sonnet 5 & \textbf{100.0} & 77.4 & 88.1 & 62.3 & 41.8 & 1.0 & 72.2 & 43.6 & 54.7 & 23.8 & 74.5 & 35.0 & 71.3 & 39.2\\
\multicolumn{15}{@{}l@{}}{\textbf{Open-Source Models}}\\
DeepSeek-V4.1-Flash & \textbf{100.0} & 53.7 & 92.1 & 59.3 & 65.3 & 20.8 & 0.0 & 0.5 & 0.0 & 0.0 & 58.4 & 18.2 & 55.4 & 26.7\\
MiniMax-M3 & \textbf{100.0} & 50.0 & 85.6 & 58.1 & 17.2 & 2.9 & 25.3 & 5.0 & 45.6 & 20.9 & 68.5 & 17.7 & 58.6 & 26.9\\
Qwen3.8-27B & 58.8 & 46.1 & 53.5 & 26.8 & 41.6 & 0.6 & 8.7 & 8.7 & 37.3 & 0.6 & 54.2 & 23.4 & 44.6 & 17.5\\
Qwen3.5-9B & 60.6 & 36.9 & 10.8 & 5.0 & 0.0 & 0.0 & 0.0 & 12.1 & 14.0 & 0.0 & 0.0 & 0.0 & 11.5 & 6.6\\
Qwen3.5-4B & 7.8 & 0.2 & 5.6 & 6.5 & 0.0 & 0.0 & 0.0 & 0.0 & 0.1 & 0.1 & 0.0 & 0.0 & 2.1 & 1.5\\
\midrule
\multicolumn{15}{@{}l@{}}{\textbf{Time Series Language Models}}\\
TimeOmni-1-4B & 0.0 & 0.0 & 0.0 & 0.0 & 0.0 & 0.0 & 0.0 & 0.0 & 3.1 & 1.4 & 0.0 & 0.0 & 0.5 & 0.2\\
\midrule
\multicolumn{15}{@{}l@{}}{\textbf{Ours}}\\
\rowcolor{blue!5}
\method & 86.1 & 80.6 & 94.8 & \textbf{93.4} & \textbf{93.8} & \textbf{94.9} & \textbf{93.3} & \textbf{88.7} & \textbf{81.8} & \textbf{88.5} & \textbf{86.6} & \textbf{82.7} & \textbf{89.5} & \textbf{88.5}\\
\bottomrule\end{tabular}\endgroup\end{table*}

\subsubsection{Numerical Generalization}
\label{app:numerical-generalization}
\textbf{Setup.} Table~\ref{tab:numerical-shifts} evaluates numerical generalization on 850 held-out streams across all 18 tasks, with shifted numerical ranges and generation settings. Programs and checkpoints are frozen.

\textbf{Results.} \method scores 61.5 under numerical shift, compared with 90.0 on ID Test. Performance varies across capabilities, consistent with their differing demands on learned numerical reasoning: C2 Sequential Conditions declines from 92.6 to 20.4, while C4 Event Reference retains 84.0. Despite the substantial drop, our overall score remains above DeepSeek-V4.1-Flash (54.3).

\begin{table*}[!htbp]
\caption{Numerical shifts (NS; Joint Score $S$) by capability on ID and numerical-shift (NS) testbeds. Code-executing models and \method use all 889 ID and 850 NS streams; the two direct streaming baselines use 90 streams per split due to evaluation cost. Overall averages all 18 tasks. \textbf{Bold}: best.}
\label{tab:numerical-shifts}
\centering\begingroup\fontsize{7.5}{9}\selectfont
\renewcommand{\arraystretch}{1.15}
\setlength{\tabcolsep}{1.25pt}
\begin{tabular*}{\linewidth}{@{\extracolsep{\fill}}>{\raggedright\arraybackslash}p{89pt}*{14}{c}@{}}\toprule
\multirow{2}{*}[-0.5ex]{\textbf{Method}} & \multicolumn{2}{c}{\textbf{\begin{tabular}[c]{@{}c@{}}C1: Instantaneous\\Conditions\end{tabular}}} & \multicolumn{2}{c}{\textbf{\begin{tabular}[c]{@{}c@{}}C2: Sequential\\Conditions\end{tabular}}} & \multicolumn{2}{c}{\textbf{\begin{tabular}[c]{@{}c@{}}C3: Event\\Tracking\end{tabular}}} & \multicolumn{2}{c}{\textbf{\begin{tabular}[c]{@{}c@{}}C4: Event\\Reference\end{tabular}}} & \multicolumn{2}{c}{\textbf{\begin{tabular}[c]{@{}c@{}}C5: Online\\Instructions\end{tabular}}} & \multicolumn{2}{c}{\textbf{\begin{tabular}[c]{@{}c@{}}C6: Ultra-Long\\Reliability\end{tabular}}} & \multicolumn{2}{c}{\textbf{Overall}}\\
\cmidrule(lr){2-3}\cmidrule(lr){4-5}\cmidrule(lr){6-7}\cmidrule(lr){8-9}\cmidrule(lr){10-11}\cmidrule(lr){12-13}\cmidrule(lr){14-15}
 & ID & NS & ID & NS & ID & NS & ID & NS & ID & NS & ID & NS & ID & NS\\\midrule
\multicolumn{15}{@{}l@{}}{\textbf{Proprietary Models}}\\
Opus 5 & 98.0 & 97.0 & \textbf{96.1} & \textbf{98.0} & 86.6 & \textbf{89.2} & 74.4 & 77.1 & 49.0 & 44.2 & 72.8 & \textbf{67.0} & 79.3 & \textbf{78.2}\\
Sonnet 5 & \textbf{100.0} & \textbf{100.0} & 88.3 & 78.5 & 39.4 & 39.6 & 75.6 & 78.6 & 50.7 & 51.3 & 71.5 & 63.7 & 70.0 & 66.6\\
\multicolumn{15}{@{}l@{}}{\textbf{Open-Source Models}}\\
DeepSeek-V4.1-Flash & \textbf{100.0} & \textbf{100.0} & 87.8 & 84.4 & 64.9 & 65.5 & 0.0 & 0.6 & 0.0 & 0.0 & 62.7 & 60.5 & 55.4 & 54.3\\
MiniMax-M3 & 99.9 & \textbf{100.0} & 82.7 & 71.9 & 13.6 & 1.5 & 22.0 & 22.1 & 47.3 & 42.9 & 66.6 & 64.1 & 56.9 & 51.2\\
Qwen3.8-27B & 54.2 & 51.9 & 56.4 & 44.5 & 34.9 & 31.8 & 17.8 & 20.6 & 38.1 & 38.9 & 50.3 & 46.5 & 43.9 & 40.1\\
Qwen3.5-9B & 53.4 & 52.5 & 8.2 & 9.4 & 0.0 & 0.0 & 0.0 & 0.0 & 11.6 & 11.8 & 0.0 & 0.0 & 9.7 & 9.9\\
Qwen3.5-4B & 9.3 & 11.6 & 6.7 & 5.1 & 0.0 & 0.0 & 0.0 & 0.0 & 0.1 & 0.0 & 0.0 & 0.0 & 2.5 & 2.4\\
\mbox{Qwen3.5-4B (streaming)} & 2.1 & 0.9 & 0.0 & 0.4 & 0.0 & 0.0 & 0.0 & 0.0 & 0.0 & 0.0 & 0.0 & 0.0 & 0.2 & 0.2\\
\midrule
\multicolumn{15}{@{}l@{}}{\textbf{Time Series Language Models}}\\
TimeOmni-1-4B & 0.0 & 0.0 & 0.0 & 0.0 & 0.0 & 0.0 & 0.0 & 0.0 & 4.3 & 2.2 & 0.0 & 0.0 & 0.7 & 0.4\\
\mbox{TS-Reasoner-7B (streaming)} & 1.0 & 0.7 & 1.1 & 0.7 & 0.1 & 0.1 & 0.0 & 0.1 & 0.5 & 0.1 & 0.0 & 0.0 & 0.5 & 0.3\\
\midrule
\multicolumn{15}{@{}l@{}}{\textbf{Ours}}\\
\rowcolor{blue!5}
\method & 87.6 & 86.2 & 92.6 & 20.4 & \textbf{95.2} & 75.1 & \textbf{93.8} & \textbf{84.0} & \textbf{84.2} & \textbf{70.9} & \textbf{86.9} & 61.6 & \textbf{90.0} & 61.5\\
\bottomrule\end{tabular*}\endgroup\end{table*}

\subsubsection{Third-Party Dataset Evaluation}
\label{app:external-details}
\textbf{Setup.} We evaluate domain transfer on KPI (AIOps Challenge 2018), which is absent from our original training data. Each series is split chronologically into 70\% training and 30\% test data; 18 series with at least three test anomaly segments yield 11,322 chunks and 638 reference segments. One-hour chunks are interpolated to 512 samples. Reports are mapped to original sample positions by their reported values, falling back to the largest deviation from the chunk median. Event-F1 PA counts a segment as detected if it contains a predicted point; predictions outside reference segments are false positives. Precision, Recall, and Event-F1 PA are computed per series and averaged equally. Baseline values come from Table~4 of \citet{agentic_coding_monitoring-1}, which reports the best of ten trials for LLM methods.

\textbf{Results.} In Table~\ref{tab:external}, \method achieves Event-F1 PA of 0.200 without adaptation and 0.539 after training on the KPI training partition. Adaptation raises recall from 0.197 to 0.568. The zero-shot result exceeds the two published prompting baselines; the adapted result exceeds AnomalyTransformer but is slightly below TFAD. The adapted checkpoint is used only for this experiment.

\begin{table}[!htbp]
\centering
\caption{KPI evaluation ($\uparrow$; per-series means). Published baselines from \citet{agentic_coding_monitoring-1}.}
\label{tab:external}
\begingroup\fontsize{7.5}{9}\selectfont
\setlength{\tabcolsep}{4pt}
\renewcommand{\arraystretch}{1.12}
\begin{tabular*}{\linewidth}{@{\extracolsep{\fill}}clccc@{}}
\toprule
\textbf{Setting} & \multicolumn{1}{c}{\textbf{Method}} & \textbf{Precision \boldmath$\uparrow$} & \textbf{Recall \boldmath$\uparrow$} & \textbf{Event-F1 PA \boldmath$\uparrow$}\\
\midrule
\multirow{3}{*}{Zero-shot} & LLMAD~\citep{liu2025large} & 0.033 & 0.940 & 0.053\\
 & SigLLM-Prompter~\citep{alnegheimish2024large} & 0.007 & 0.498 & 0.014\\
 & \method & 0.340 & 0.197 & 0.200\\
\midrule
\multirow{3}{*}{KPI-trained} & TFAD~\citep{zhang2022tfad} & 0.690 & 0.670 & 0.564\\
 & AnomalyTransformer~\citep{xu2022anomaly} & 0.482 & 0.395 & 0.282\\
 & \method & 0.602 & 0.568 & 0.539\\
\bottomrule
\end{tabular*}
\endgroup
\end{table}

\subsection{Efficiency and Edge Deployment}
\label{app:efficiency-details}\label{app:deployment-details}

\begin{table}[H]
\centering
\caption{Efficiency by group and pooled over all 54 streams. Each group contains 18 streams replayed at 1 chunk/s. Pooled values match Table~\ref{tab:runtime}. Parentheses give ratios to H100 \method ($c=15$, $1\times$), computed before rounding. Tokens are input tokens per 1k chunks.}
\label{tab:efficiency-groups}
\begingroup\fontsize{7.5}{9}\selectfont
\setlength{\tabcolsep}{2pt}
\renewcommand{\arraystretch}{1.12}
\textbf{(a) Quality and input tokens}\par
\begin{tabular*}{\linewidth}{@{\extracolsep{\fill}}l*{8}{c}@{}}
\toprule
\multirow{2}{*}[-2pt]{\textbf{Model}} & \multicolumn{2}{c}{\textbf{Group 1}} & \multicolumn{2}{c}{\textbf{Group 2}} & \multicolumn{2}{c}{\textbf{Group 3}} & \multicolumn{2}{c}{\textbf{Pooled (54 streams)}}\\
\cmidrule(lr){2-3}\cmidrule(lr){4-5}\cmidrule(lr){6-7}\cmidrule(lr){8-9}
 & \boldmath$S\uparrow$ & \textbf{Tokens}\boldmath$\downarrow$ & \boldmath$S\uparrow$ & \textbf{Tokens}\boldmath$\downarrow$ & \boldmath$S\uparrow$ & \textbf{Tokens}\boldmath$\downarrow$ & \boldmath$S\uparrow$ & \textbf{Tokens}\boldmath$\downarrow$\\
\midrule
\multicolumn{9}{@{}l}{\textbf{H100}}\\
Qwen3.5-4B (streaming) & 0.14 & 8,691,413 & 0.05 & 8,640,881 & 5.75 & 8,353,534 & 2.0 & 8,558,122\,\textcolor{red}{(151.4\texttimes)}\\
TS-Reasoner-7B (streaming) & 0.37 & 592,390 & 0.16 & 588,440 & 0.46 & 602,393 & 0.3 & 594,623\,\textcolor{red}{(10.5\texttimes)}\\
\method ($c=15$) & 60.32 & 56,450 & 61.85 & 56,312 & 44.24 & 56,837 & 53.8 & 56,541\,\textcolor{black!55}{(1\texttimes)}\\
\midrule
\multicolumn{9}{@{}l}{\textbf{Jetson AGX Orin}}\\
\method ($c=10$) & 59.13 & 53,185 & 61.86 & 52,987 & 44.74 & 53,305 & 53.7 & 53,164\,\textcolor{black!55}{(0.94\texttimes)}\\
\method ($c=15$) & 60.12 & 56,454 & 63.16 & 56,314 & 44.24 & 56,744 & 54.3 & 56,511\,\textcolor{black!55}{(1.00\texttimes)}\\
\method ($c=20$) & 58.80 & 59,888 & 63.12 & 59,655 & 45.76 & 60,104 & 55.3 & 59,890\,\textcolor{red}{(1.06\texttimes)}\\
\bottomrule
\end{tabular*}
\par\medskip
\textbf{(b) Processing speed and latency}\par
\begin{tabular*}{\linewidth}{@{\extracolsep{\fill}}l*{8}{c}@{}}
\toprule
\multirow{2}{*}[-2pt]{\textbf{Model}} & \multicolumn{2}{c}{\textbf{Group 1}} & \multicolumn{2}{c}{\textbf{Group 2}} & \multicolumn{2}{c}{\textbf{Group 3}} & \multicolumn{2}{c}{\textbf{Pooled (54 streams)}}\\
\cmidrule(lr){2-3}\cmidrule(lr){4-5}\cmidrule(lr){6-7}\cmidrule(lr){8-9}
 & \textbf{RTF}\boldmath$\downarrow$ & \textbf{p95 (s)}\boldmath$\downarrow$ & \textbf{RTF}\boldmath$\downarrow$ & \textbf{p95 (s)}\boldmath$\downarrow$ & \textbf{RTF}\boldmath$\downarrow$ & \textbf{p95 (s)}\boldmath$\downarrow$ & \textbf{RTF}\boldmath$\downarrow$ & \textbf{p95 (s)}\boldmath$\downarrow$\\
\midrule
\multicolumn{9}{@{}l}{\textbf{H100}}\\
Qwen3.5-4B (streaming) & 0.716 & 2.165 & 0.760 & 13.115 & 0.772 & 25.787 & 0.749\,\textcolor{red}{(6.4\texttimes)} & 16.615\,\textcolor{red}{(206\texttimes)}\\
TS-Reasoner-7B (streaming) & 0.275 & 0.475 & 0.277 & 0.527 & 0.274 & 0.510 & 0.276\,\textcolor{red}{(2.4\texttimes)} & 0.507\,\textcolor{red}{(6.3\texttimes)}\\
\method ($c=15$) & 0.104 & 0.078 & 0.103 & 0.078 & 0.143 & 0.439 & 0.117\,\textcolor{black!55}{(1\texttimes)} & 0.081\,\textcolor{black!55}{(1\texttimes)}\\
\midrule
\multicolumn{9}{@{}l}{\textbf{Jetson AGX Orin}}\\
\method ($c=10$) & 0.394 & 3.105 & 0.323 & 0.276 & 0.404 & 4.929 & 0.375\,\textcolor{red}{(3.2\texttimes)} & 2.246\,\textcolor{red}{(27.8\texttimes)}\\
\method ($c=15$) & 0.336 & 0.288 & 0.323 & 0.262 & 0.412 & 2.629 & 0.358\,\textcolor{red}{(3.1\texttimes)} & 0.649\,\textcolor{red}{(8.0\texttimes)}\\
\method ($c=20$) & 0.357 & 0.308 & 0.320 & 0.252 & 0.399 & 1.994 & 0.360\,\textcolor{red}{(3.1\texttimes)} & 0.592\,\textcolor{red}{(7.3\texttimes)}\\
\bottomrule
\end{tabular*}
\endgroup
\end{table}

\textbf{Efficiency.} \\\textbf{Setup.} Table~\ref{tab:efficiency} evaluates 54 complete Real Test streams (three per task, 31,033 chunks) at one 512-sample chunk per second. Each model runs on a dedicated H100 with batch size one, a complete first-stream warm-up per session, and one measured repetition per stream. The cohort consists of three disjoint 18-stream groups. Joint Score follows the task-level formula in Appendix~\ref{app:streaming-evaluation}, as in Table~\ref{tab:main}, recomputed across all 54 streams. Input tokens and active processing time are pooled before normalization, and p95 is computed over all chunk latencies rather than averaged across groups.
\textbf{Results.} Table~\ref{tab:efficiency-groups} reports each group and the pooled results from Table~\ref{tab:runtime}. Relative to H100 \method ($c=15$, $1\times$), Qwen3.5-4B streaming uses $151.4\times$ the input tokens and has $6.4\times$ the RTF and $206\times$ the p95 latency. The corresponding ratios for TS-Reasoner-7B are $10.5\times$, $2.4\times$, and $6.3\times$.

\textbf{Scheduling ablations.}\\
\textbf{Setup.} Figure~\ref{fig:runtime-scheduling} uses a separate complete T13 stream (557 chunks), with $W=60$, $c=15$, and a 160-chunk warm-up. Each configuration is measured once; the main figure shows seconds 30--200. \textbf{Results.} Without elastic synchronization, consecutive chunk starts are spaced by at least one second. Without early memory updates, memory decoding waits until the window end, using the same snapshot and prompt from $W-c$. Removing both combines these restrictions. Responses and memory contents remain identical across configurations. Full-stream p95 latencies are 0.093\,s (\method), 0.644\,s (without elastic synchronization), 0.531\,s (without early updates), and 15.380\,s (without both). Figure~\ref{fig:runtime-backlog} shows the corresponding backlog accumulation and recovery.

\begingroup
\setlength{\intextsep}{6pt}
\begin{figure}[H]
\centering
\includegraphics[width=\linewidth,trim=0 7bp 0 2bp,clip]{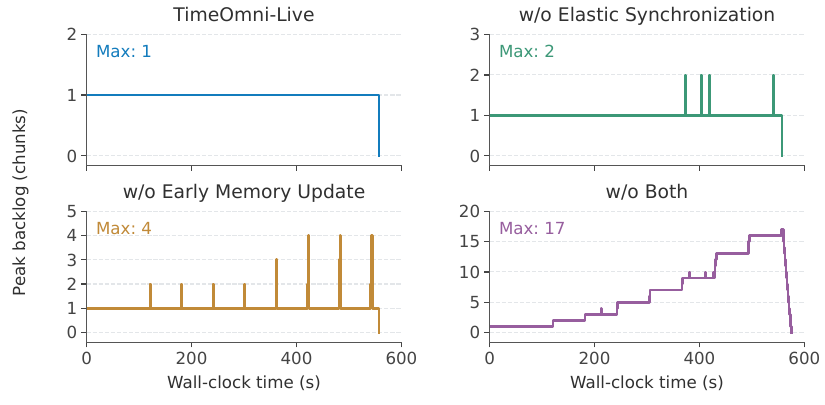}

\vspace{-4pt}\caption{\textbf{Streaming backlog at 1 chunk/s.} Each panel shows the maximum number of arrived but unfinished chunks within each one-second interval, including the chunk being processed. Configurations are shown separately to avoid overlapping traces; note the different vertical scales.}
\label{fig:runtime-backlog}

\end{figure}
\endgroup

\textbf{Edge deployment.} \\\textbf{Setup.} Jetson AGX Orin uses the same checkpoint and 54-stream replay as H100, with BF16 backbone weights, FP32 encoder/adapter weights, and no quantization. It runs in MAXN mode with approximately 61\,GiB shared memory. \textbf{Results.} Across 480 memory updates, deadline misses fall from 117 at $c=10$ to 59 at $c=15$ and 14 at $c=20$ (Table~\ref{tab:jetson-overlap}). Pooled p95 latency falls from 2.246\,s to 0.592\,s, a $3.8\times$ reduction, consistent with fewer memory-update stalls. All settings achieve RTF below one. Overlap changes the retained history, so task quality is evaluated separately for each setting.
\begingroup
\setlength{\intextsep}{3pt}
\begin{table}[H]
\centering
\begin{minipage}{0.70\linewidth}
\caption{Overlap sensitivity on Jetson.}
\label{tab:jetson-overlap}
\centering\footnotesize
\setlength{\tabcolsep}{6pt}
\renewcommand{\arraystretch}{1.15}
\begin{tabular}{@{}ccc@{}}
\toprule
\boldmath$c$ & \textbf{Missed updates}\boldmath$\downarrow$ & \textbf{p95 latency (s)}\boldmath$\downarrow$\\
\midrule
10 & 117/480 & 2.246\,\textcolor{red}{(3.8\texttimes)}\\
15 & 59/480 & 0.649\,\textcolor{red}{(1.1\texttimes)}\\
20 & 14/480 & 0.592\,\textcolor{black!55}{(1\texttimes)}\\
\bottomrule
\end{tabular}

\end{minipage}
\end{table}
\endgroup

\clearpage

\section{Task Examples and Scoring Criteria}
\label{app:case-studies}
The following examples are organized by the six capability groups. Each provides a task instruction, signal evidence, reference responses, and the quantities used to score their content. These are task examples from \dataset with reference answers, not model prediction traces. The shared timing and content scoring rules are defined in Appendix~\ref{app:streaming-evaluation}.
\begingroup
\setlength{\parindent}{0pt}
\setlength{\tabcolsep}{5pt}
\renewcommand{\arraystretch}{1.06}
\subsection{C1: Instantaneous Conditions}
\begin{table}[!ht]
\centering
\small
\setlength{\abovecaptionskip}{0pt}
\setlength{\belowcaptionskip}{7pt}
\caption{Task 1 (Single-Condition Triggering): Example with ground-truth response and scoring criteria.}
\label{tab:case-task1}
\begin{tabularx}{\textwidth}{@{}>{\raggedright\arraybackslash}p{2.05cm}Y@{}}
\toprule
\multicolumn{2}{@{}l@{}}{\textbf{Capability C1: Instantaneous Conditions}}\\
\midrule
\textbf{Task} & \textbf{\textit{T1: Single-Condition Triggering}}\\[4pt]
\textbf{User} & The signal arrives one chunk at a time (512 samples per chunk; chunks are numbered from 0). Alert me whenever the reading rises above 3.3. One alarm per crossing: report the chunk where the reading first goes over the line, not the chunks that stay over. Each time, report exactly: "peaked at \textless{}value\textgreater{}", where \textless{}value\textgreater{} is the highest reading in that chunk.\\[6pt]
 & Input time series stream (shown in full; supplied sequentially in chunks of 512 samples):\\[2pt]
 & \begin{minipage}[t]{\linewidth}\vspace{0pt}\centering
\includegraphics[width=\linewidth]{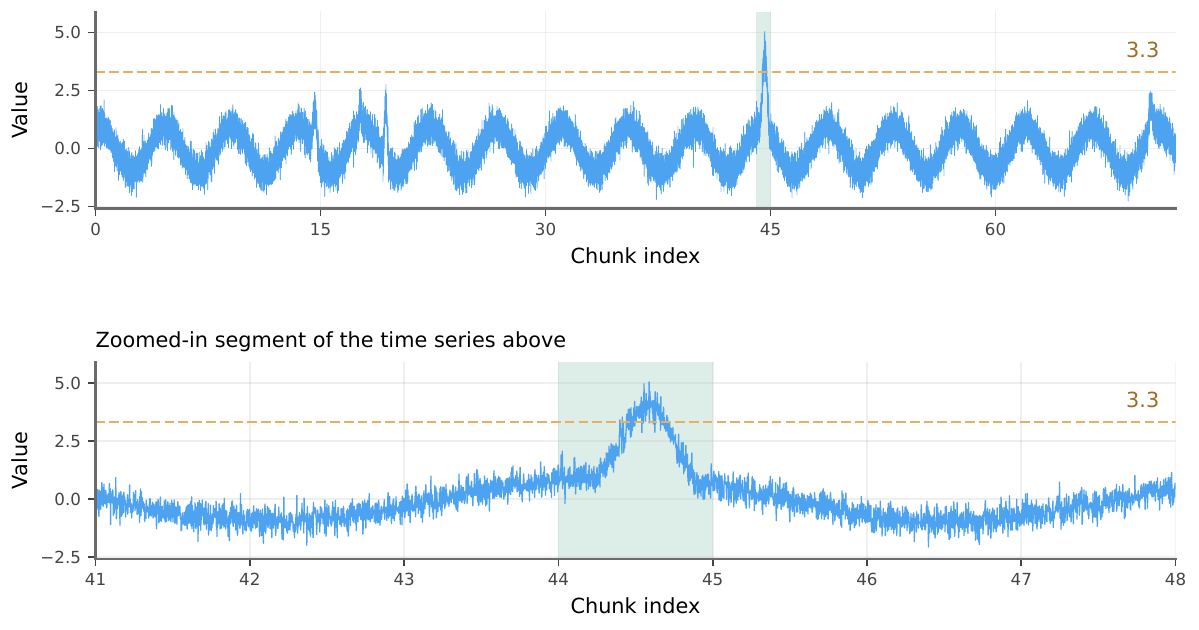}
\end{minipage}\\[4pt]
 & \textit{Note:} Only raw samples are provided as signal input. Shading, labels and guide lines are for illustration only and are not model inputs.\\[6pt]
\midrule
\textbf{Ground Truth} & \textbf{Response time:} chunk 44. \quad \textbf{Reported content:} peaked at 5.0391.\\[6pt]
\midrule
\textbf{Scoring criteria} & \textbf{Timing:} detect the threshold-crossing event. \textbf{Content:} score the reported excess above 3.3 against the ground-truth excess, 1.7391.\\[4pt]
\bottomrule
\end{tabularx}
\end{table}
\clearpage

\begin{table}[!ht]
\centering
\small
\setlength{\abovecaptionskip}{0pt}
\setlength{\belowcaptionskip}{7pt}
\caption{Task 2 (Multi-Condition Triggering): Example with ground-truth response and scoring criteria.}
\label{tab:case-task2}
\begin{tabularx}{\textwidth}{@{}>{\raggedright\arraybackslash}p{2.05cm}Y@{}}
\toprule
\multicolumn{2}{@{}l@{}}{\textbf{Capability C1: Instantaneous Conditions}}\\
\midrule
\textbf{Task} & \textbf{\textit{T2: Multi-Condition Triggering}}\\[4pt]
\textbf{User} & The signal arrives one chunk at a time (512 samples per chunk; chunks are numbered from 0). Notify me exactly when the two conditions coincide: the average level is above 1.37, and the fluctuation is above 1.57. Report every chunk in which both conditions hold. Report both readings in one message, exactly: "average at \textless{}reading\textgreater{}, fluctuation at \textless{}reading\textgreater{}", where the first \textless{}reading\textgreater{} is that chunk's mean and the second is that chunk's standard deviation.\\[6pt]
 & Input time series stream (shown in full; supplied sequentially in chunks of 512 samples):\\[2pt]
 & \begin{minipage}[t]{\linewidth}\vspace{0pt}\centering
\includegraphics[width=\linewidth]{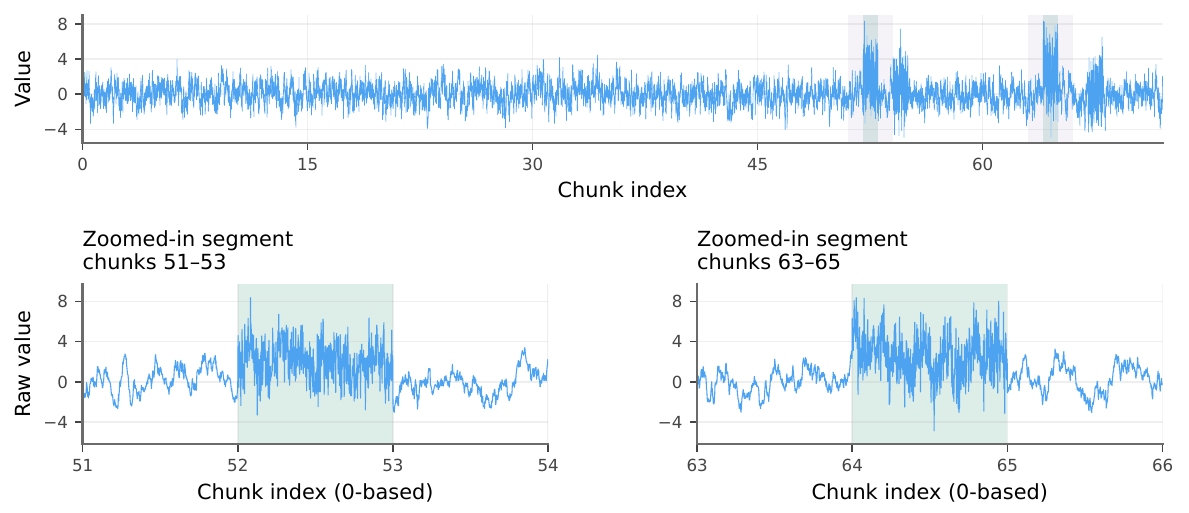}
\end{minipage}\\[4pt]
 & \textit{Note:} Only raw samples are provided as signal input. Shading, labels and guide lines are for illustration only and are not model inputs.\\[6pt]
\midrule
\textbf{Ground Truth} & \textbf{Response time:} chunk 52. \quad \textbf{Reported content:} average at 2.02, fluctuation at 1.88.\\[4pt]
 & \textbf{Response time:} chunk 64. \quad \textbf{Reported content:} average at 2.54, fluctuation at 2.15.\\[4pt]
\midrule
\textbf{Evidence} & Average is the mean and fluctuation is the standard deviation of the 512 samples within the current chunk (divisor 512). Each curve subtracts its threshold from that statistic; both must be positive. \textit{For explanation only; not model input.}\\[3pt]
 & \begin{minipage}[t]{\linewidth}\vspace{0pt}\centering
\includegraphics[width=\linewidth]{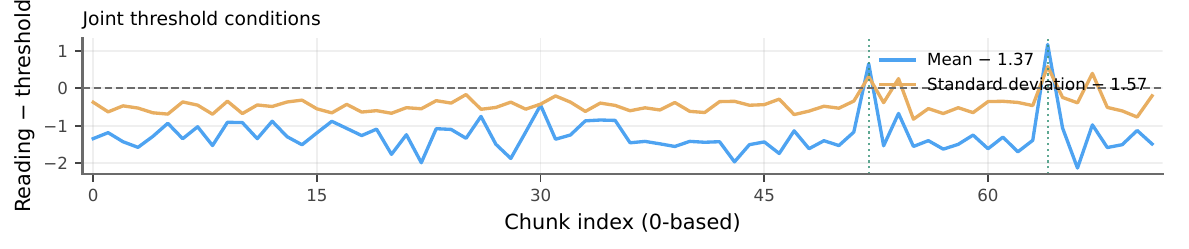}
\end{minipage}\\[4pt]
\midrule
\textbf{Scoring criteria} & \textbf{Timing:} respond when both conditions hold simultaneously. \textbf{Content:} compare both readings as excesses relative to their respective thresholds.\\[4pt]
\bottomrule
\end{tabularx}
\end{table}
\clearpage

\subsection{C2: Sequential Conditions}
\begin{table}[!ht]
\centering
\small
\setlength{\abovecaptionskip}{0pt}
\setlength{\belowcaptionskip}{7pt}
\caption{Task 3 (Consecutive Exceedance): Example with ground-truth response and scoring criteria.}
\label{tab:case-task3}
\begin{tabularx}{\textwidth}{@{}>{\raggedright\arraybackslash}p{2.05cm}Y@{}}
\toprule
\multicolumn{2}{@{}l@{}}{\textbf{Capability C2: Sequential Conditions}}\\
\midrule
\textbf{Task} & \textbf{\textit{T3: Consecutive Exceedance}}\\[4pt]
\textbf{User} & The signal arrives one chunk at a time (512 samples per chunk; chunks are numbered from 0). I only care if the level remains above 5.1 for 4 chunks straight -- then raise an alarm. Speak at the chunk that completes the run, once per run; the run has to break and build up again before the next alarm. When you call it, report exactly: "peaked at \textless{}value\textgreater{}", where \textless{}value\textgreater{} is the highest reading across those consecutive chunks.\\[6pt]
 & Input time series stream (shown in full; supplied sequentially in chunks of 512 samples):\\[2pt]
 & \begin{minipage}[t]{\linewidth}\vspace{0pt}\centering
\includegraphics[width=\linewidth]{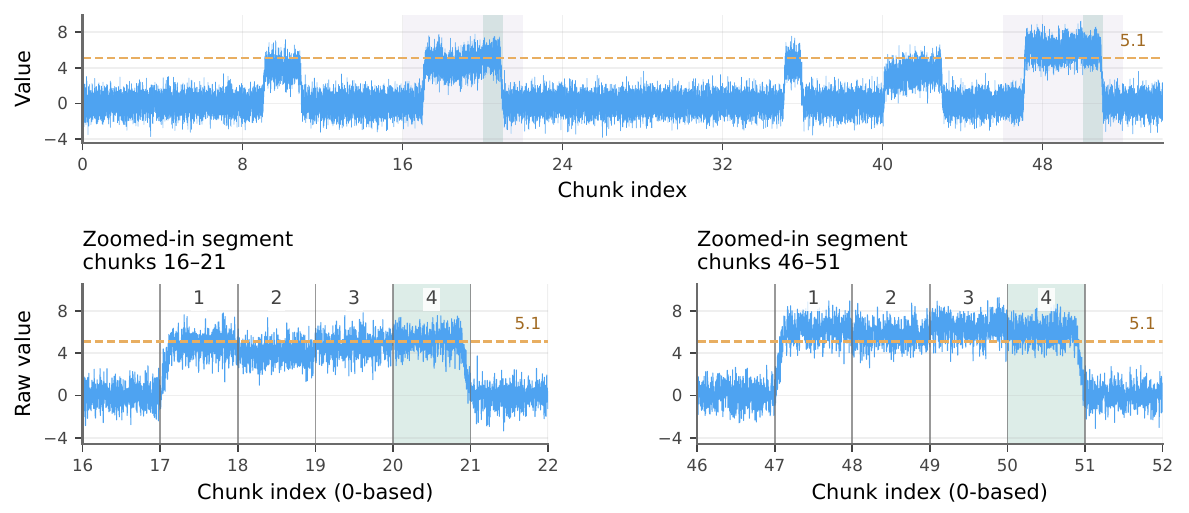}
\end{minipage}\\[4pt]
 & \textit{Note:} Only raw samples are provided as signal input. Shading, labels and guide lines are for illustration only and are not model inputs.\\[6pt]
\midrule
\textbf{Ground Truth} & \textbf{Response time:} chunk 20. \quad \textbf{Reported content:} peaked at 7.8192.\\[4pt]
 & \textbf{Response time:} chunk 50. \quad \textbf{Reported content:} peaked at 9.2415.\\[4pt]
\midrule
\textbf{Evidence} & Each close-up includes the preceding and following chunks for context. The four consecutive chunks are separated by vertical guides and numbered 1--4. Their chunk maxima exceed 5.1; the response is due at the fourth chunk. \textit{For explanation only; not model input.}\\[3pt]
 & \begin{minipage}[t]{\linewidth}\vspace{0pt}\centering
\includegraphics[width=\linewidth]{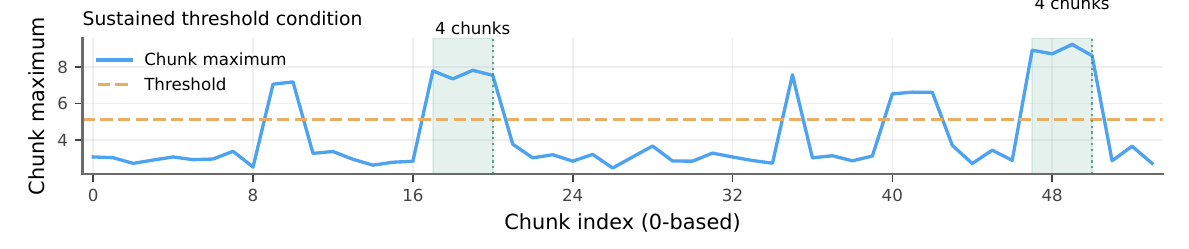}
\end{minipage}\\[4pt]
\midrule
\textbf{Scoring criteria} & \textbf{Timing:} respond when the required consecutive run is complete. \textbf{Content:} compare the reported run-wide peak as an excess relative to the threshold.\\[4pt]
\bottomrule
\end{tabularx}
\end{table}
\clearpage

\begin{table}[!ht]
\centering
\small
\setlength{\abovecaptionskip}{0pt}
\setlength{\belowcaptionskip}{7pt}
\caption{Task 4 (Trend Recognition): Example with ground-truth response and scoring criteria.}
\label{tab:case-task4}
\begin{tabularx}{\textwidth}{@{}>{\raggedright\arraybackslash}p{2.05cm}Y@{}}
\toprule
\multicolumn{2}{@{}l@{}}{\textbf{Capability C2: Sequential Conditions}}\\
\midrule
\textbf{Task} & \textbf{\textit{T4: Trend Recognition}}\\[4pt]
\textbf{User} & The signal arrives one chunk at a time (512 samples per chunk; chunks are numbered from 0). Speak up when the value has clearly increased for 4 straight chunks. "Clearly" means each chunk mean moving more than 2.257 from the previous chunk. Speak at the chunk that completes the run, once per run; a chunk that fails to move clearly resets the count. Report exactly: "moving \textless{}up|down\textgreater{}, covered \textless{}amount\textgreater{}", where \textless{}amount\textgreater{} is the total change of the chunk mean across those consecutive chunks.\\[6pt]
 & Input time series stream (shown in full; supplied sequentially in chunks of 512 samples):\\[2pt]
 & \begin{minipage}[t]{\linewidth}\vspace{0pt}\centering
\includegraphics[width=\linewidth]{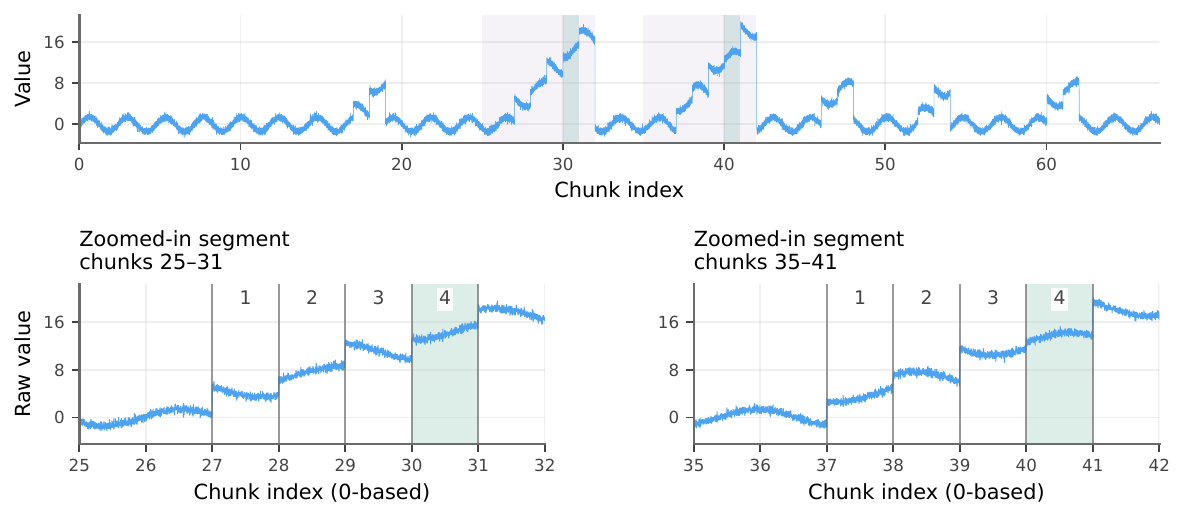}
\end{minipage}\\[4pt]
 & \textit{Note:} Only raw samples are provided as signal input. Shading, labels and guide lines are for illustration only and are not model inputs.\\[6pt]
\midrule
\textbf{Ground Truth} & \textbf{Response time:} chunk 30. \quad \textbf{Reported content:} moving up, covered 13.0855.\\[4pt]
 & \textbf{Response time:} chunk 40. \quad \textbf{Reported content:} moving up, covered 13.656.\\[4pt]
\midrule
\textbf{Evidence} & Chunks numbered 1--4 each have a higher mean than the preceding chunk, with every increase exceeding the stated margin. The first increase is measured from the preceding, unnumbered chunk; the fourth completes the run and triggers the response. \textit{For explanation only; not model input.}\\[3pt]
 & \begin{minipage}[t]{\linewidth}\vspace{0pt}\centering
\includegraphics[width=\linewidth]{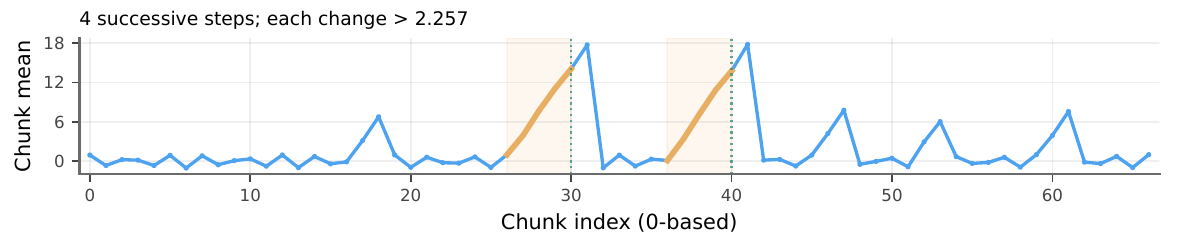}
\end{minipage}\\[4pt]
\midrule
\textbf{Scoring criteria} & \textbf{Timing:} respond at completion of the required consecutive steps. \textbf{Content:} check the direction and total movement.\\[4pt]
\bottomrule
\end{tabularx}
\end{table}
\clearpage

\begin{table}[!ht]
\centering
\small
\setlength{\abovecaptionskip}{0pt}
\setlength{\belowcaptionskip}{7pt}
\caption{Task 5 (Mean Shift): Example with ground-truth response and scoring criteria.}
\label{tab:case-task5}
\begin{tabularx}{\textwidth}{@{}>{\raggedright\arraybackslash}p{2.05cm}Y@{}}
\toprule
\multicolumn{2}{@{}l@{}}{\textbf{Capability C2: Sequential Conditions}}\\
\midrule
\textbf{Task} & \textbf{\textit{T5: Mean Shift}}\\[4pt]
\textbf{User} & The signal arrives one chunk at a time (512 samples per chunk; chunks are numbered from 0). Tell me the moment the signal moves clearly off its recent norm and keeps that for 4 chunks. "Clearly" means the chunk mean moving more than 0.319 away from the median of the 8 chunks just before the shift began, and staying that far away for 4 chunks in a row. Speak at the last of those 4 chunks, once; after an alarm, stay silent until a chunk comes along that no longer counts as clearly shifted from the median of the 8 chunks before it. Report exactly: "settled \textless{}higher|lower\textgreater{} by \textless{}amount\textgreater{}", where \textless{}amount\textgreater{} is the average chunk mean after the change minus the average chunk mean before it.\\[6pt]
 & Input time series stream (shown in full; supplied sequentially in chunks of 512 samples):\\[2pt]
 & \begin{minipage}[t]{\linewidth}\vspace{0pt}\centering
\includegraphics[width=\linewidth]{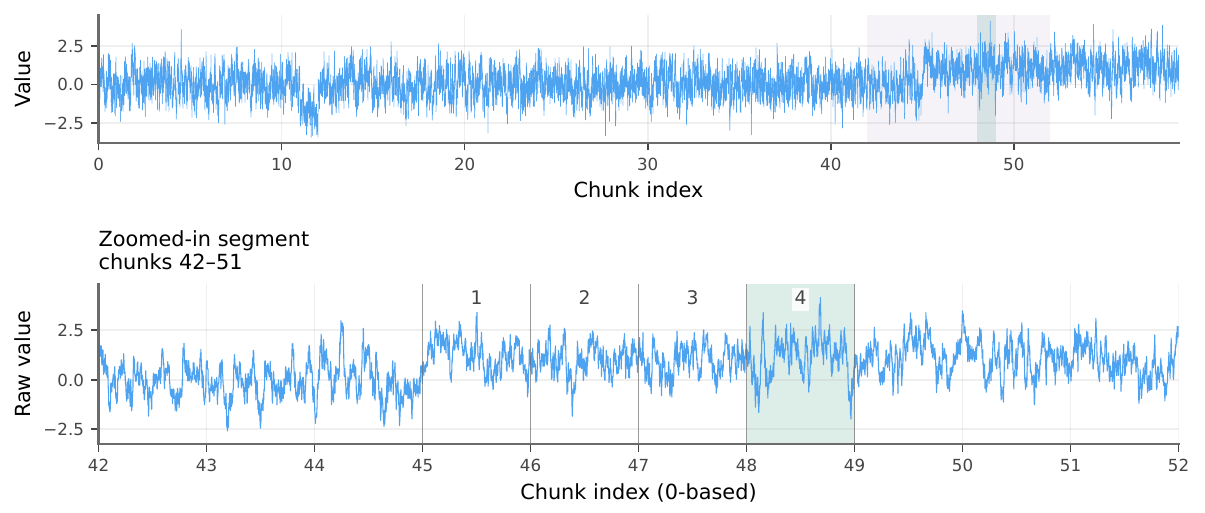}
\end{minipage}\\[4pt]
 & \textit{Note:} Only raw samples are provided as signal input. Shading, labels and guide lines are for illustration only and are not model inputs.\\[6pt]
\midrule
\textbf{Ground Truth} & \textbf{Response time:} chunk 48. \quad \textbf{Reported content:} settled higher by 1.1149.\\[4pt]
\midrule
\textbf{Evidence} & Chunks numbered 1--4 show the shifted level persisting for the required four chunks. The response is due at the fourth chunk, rather than at the initial shift. The plot below shows the chunk means and shift onset. \textit{For explanation only; not model input.}\\[3pt]
 & \begin{minipage}[t]{\linewidth}\vspace{0pt}\centering
\includegraphics[width=\linewidth]{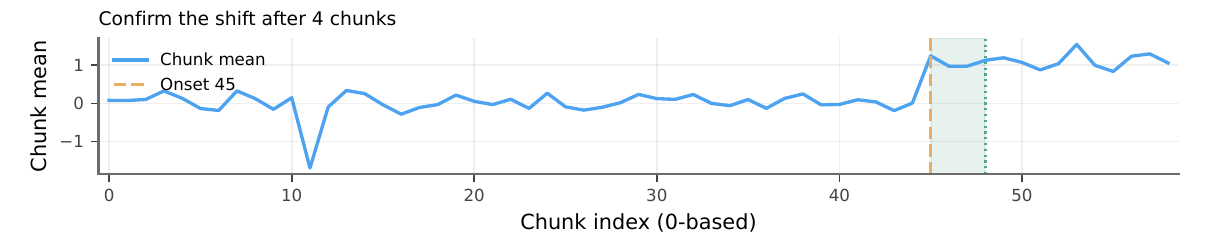}
\end{minipage}\\[4pt]
\midrule
\textbf{Scoring criteria} & \textbf{Timing:} respond once the level shift has persisted for the required duration. \textbf{Content:} check the shift direction and magnitude.\\[4pt]
\bottomrule
\end{tabularx}
\end{table}
\clearpage

\begin{table}[!ht]
\centering
\small
\setlength{\abovecaptionskip}{0pt}
\setlength{\belowcaptionskip}{7pt}
\caption{Task 6 (Event Density): Example with ground-truth response and scoring criteria.}
\label{tab:case-task6}
\begin{tabularx}{\textwidth}{@{}>{\raggedright\arraybackslash}p{2.05cm}Y@{}}
\toprule
\multicolumn{2}{@{}l@{}}{\textbf{Capability C2: Sequential Conditions}}\\
\midrule
\textbf{Task} & \textbf{\textit{T6: Event Density}}\\[4pt]
\textbf{User} & The signal arrives one chunk at a time (512 samples per chunk; chunks are numbered from 0). Raise an alarm once 3 spikes have occurred within the last 6 chunks. A spike means the signal shooting past 5.8; smaller wiggles are not spikes. Speak at the chunk where the count inside the window first reaches 3; after that, stay silent until the count inside the window drops below 3 again. Report exactly: "spanned \textless{}n\textgreater{} chunks", where \textless{}n\textgreater{} is the chunk index of the last spike minus that of the first, among the 3 spikes that set off the alarm.\\[6pt]
 & Input time series stream (shown in full; supplied sequentially in chunks of 512 samples):\\[2pt]
 & \begin{minipage}[t]{\linewidth}\vspace{0pt}\centering
\includegraphics[width=\linewidth]{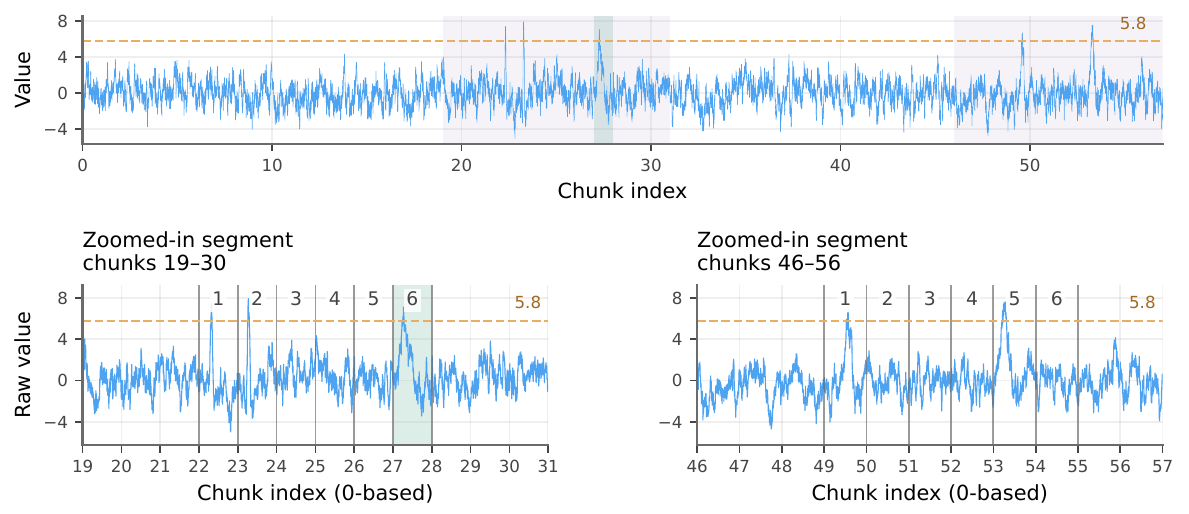}
\end{minipage}\\[4pt]
 & \textit{Note:} Only raw samples are provided as signal input. Shading, labels and guide lines are for illustration only and are not model inputs.\\[6pt]
\midrule
\textbf{Ground Truth} & \textbf{Response time:} chunk 27. \quad \textbf{Reported content:} spanned 5 chunks.\\[4pt]
\midrule
\textbf{Evidence} & The six chunks numbered 1--6 contain three spikes, triggering the response in the final chunk. The reported span is the gap between the first and last spike (5 chunks), not the window length (6 chunks). In the other close-up, chunks 49--54 are numbered 1--6, starting at the first spike; only 2 spikes occur in this window (\textless{} 3), so no response is required. \textit{For explanation only; not model input.}\\[3pt]
 & \begin{minipage}[t]{\linewidth}\vspace{0pt}\centering
\includegraphics[width=\linewidth]{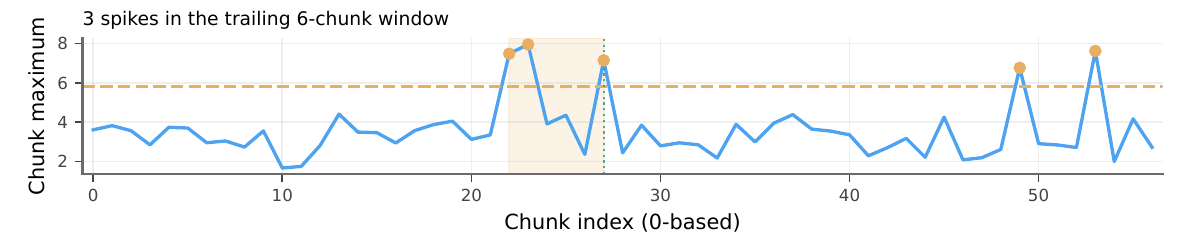}
\end{minipage}\\[4pt]
\midrule
\textbf{Scoring criteria} & \textbf{Timing:} respond when the trailing window contains the required number of spikes. \textbf{Content:} exactly match the span from the first spike to the last.\\[4pt]
\bottomrule
\end{tabularx}
\end{table}
\clearpage

\subsection{C3: Event Tracking}
\begin{table}[!ht]
\centering
\small
\setlength{\abovecaptionskip}{0pt}
\setlength{\belowcaptionskip}{7pt}
\caption{Task 7 (Single-Type Tracking): Example with ground-truth response and scoring criteria.}
\label{tab:case-task7}
\begin{tabularx}{\textwidth}{@{}>{\raggedright\arraybackslash}p{2.05cm}Y@{}}
\toprule
\multicolumn{2}{@{}l@{}}{\textbf{Capability C3: Event Tracking}}\\
\midrule
\textbf{Task} & \textbf{\textit{T7: Single-Type Tracking}}\\[4pt]
\textbf{User} & The signal arrives one chunk at a time (512 samples per chunk; chunks are numbered from 0). Ignore the first 3 upward spikes. Starting with the 4th, report every one. A fault means the signal shooting past 3.1; smaller wiggles are not faults. Report each as: "fault \#\textless{}n\textgreater{} --- \textless{}g\textgreater{} chunks since the last", where \textless{}n\textgreater{} counts every fault since the start (silent ones included) and \textless{}g\textgreater{} is this fault's chunk index minus the previous fault's.\\[6pt]
 & Input time series stream (shown in full; supplied sequentially in chunks of 512 samples):\\[2pt]
 & \begin{minipage}[t]{\linewidth}\vspace{0pt}\centering
\includegraphics[width=\linewidth]{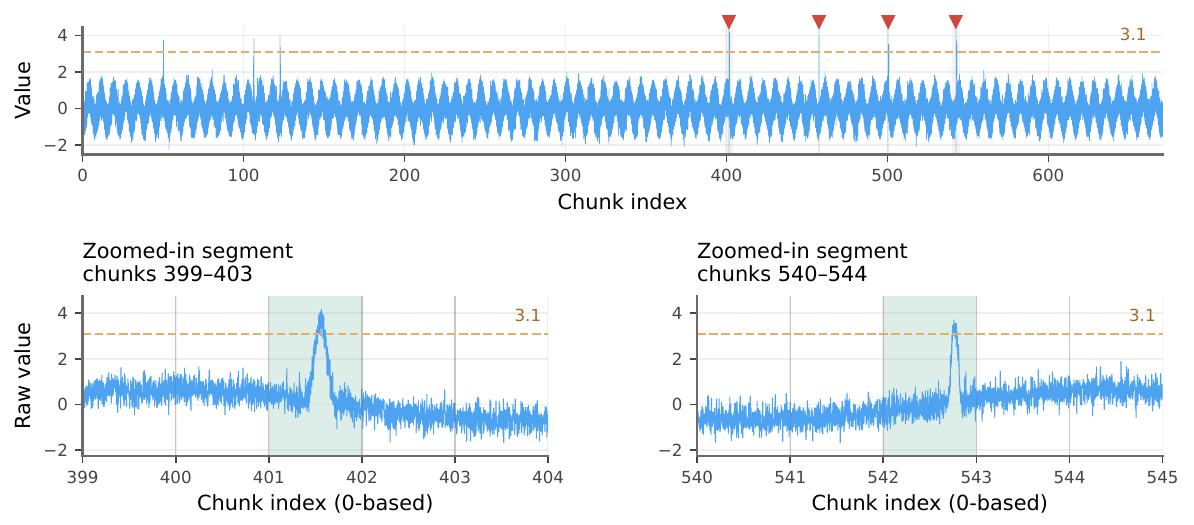}
\end{minipage}\\[4pt]
 & \textit{Note:} Only raw samples are provided as signal input. Shading, labels and guide lines are for illustration only and are not model inputs.\\[6pt]
\midrule
\textbf{Ground Truth} & \textbf{Response time:} chunk 401. \quad \textbf{Reported content:} fault \#4 --- 279 chunks since the last.\\[4pt]
 & \textbf{Response time:} chunk 457. \quad \textbf{Reported content:} fault \#5 --- 56 chunks since the last.\\[4pt]
 & \textbf{Response time:} chunk 500. \quad \textbf{Reported content:} fault \#6 --- 43 chunks since the last.\\[4pt]
 & \textbf{Response time:} chunk 542. \quad \textbf{Reported content:} fault \#7 --- 42 chunks since the last.\\[4pt]
\midrule
\textbf{Evidence} & The count includes all faults, including the first three silent ones. The close-ups show fault \#4 and fault \#7; their response gaps are measured from the immediately preceding fault. \textit{For explanation only; not model input.}\\[3pt]
 & \begin{minipage}[t]{\linewidth}\vspace{0pt}\centering
\includegraphics[width=\linewidth]{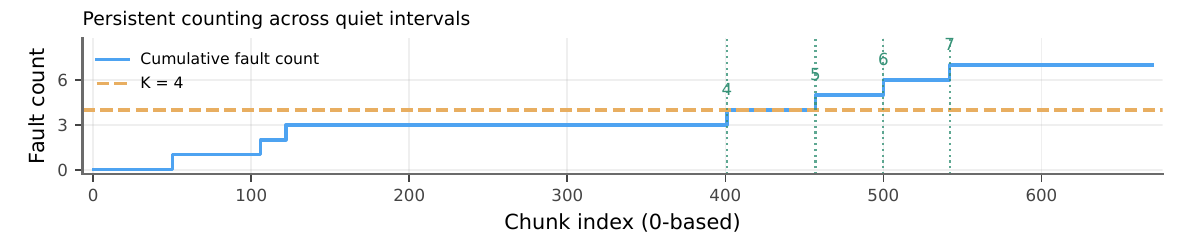}
\end{minipage}\\[4pt]
\midrule
\textbf{Scoring criteria} & \textbf{Timing:} report each qualifying fault starting at the specified count. \textbf{Content:} exactly match the fault index and score the gap since the previous fault.\\[4pt]
\bottomrule
\end{tabularx}
\end{table}
\clearpage

\begin{table}[!ht]
\centering
\small
\setlength{\abovecaptionskip}{0pt}
\setlength{\belowcaptionskip}{7pt}
\caption{Task 8 (Multi-Type Tracking): Example with ground-truth response and scoring criteria.}
\label{tab:case-task8}
\begin{tabularx}{\textwidth}{@{}>{\raggedright\arraybackslash}p{2.05cm}Y@{}}
\toprule
\multicolumn{2}{@{}l@{}}{\textbf{Capability C3: Event Tracking}}\\
\midrule
\textbf{Task} & \textbf{\textit{T8: Multi-Type Tracking}}\\[4pt]
\textbf{User} & The signal arrives one chunk at a time (512 samples per chunk; chunks are numbered from 0). Keep an eye on the sine-shaped bursts for me. 2 of them? Fine. The 3rd and beyond -- speak up every time. A burst means a shaped pulse dropping below -6.2; the background's own wiggles never count. Report each as: "sine \#\textless{}n\textgreater{} --- square at \textless{}count\textgreater{}, triangle at \textless{}count\textgreater{}", where \textless{}n\textgreater{} is how many sine bursts have occurred so far including this one (so the 3rd sine is "sine \#3"), and each \textless{}count\textgreater{} is how many bursts of that other shape have occurred so far, up to and including this chunk.\\[6pt]
 & Input time series stream (shown in full; supplied sequentially in chunks of 512 samples):\\[2pt]
 & \begin{minipage}[t]{\linewidth}\vspace{0pt}\centering
\includegraphics[width=\linewidth]{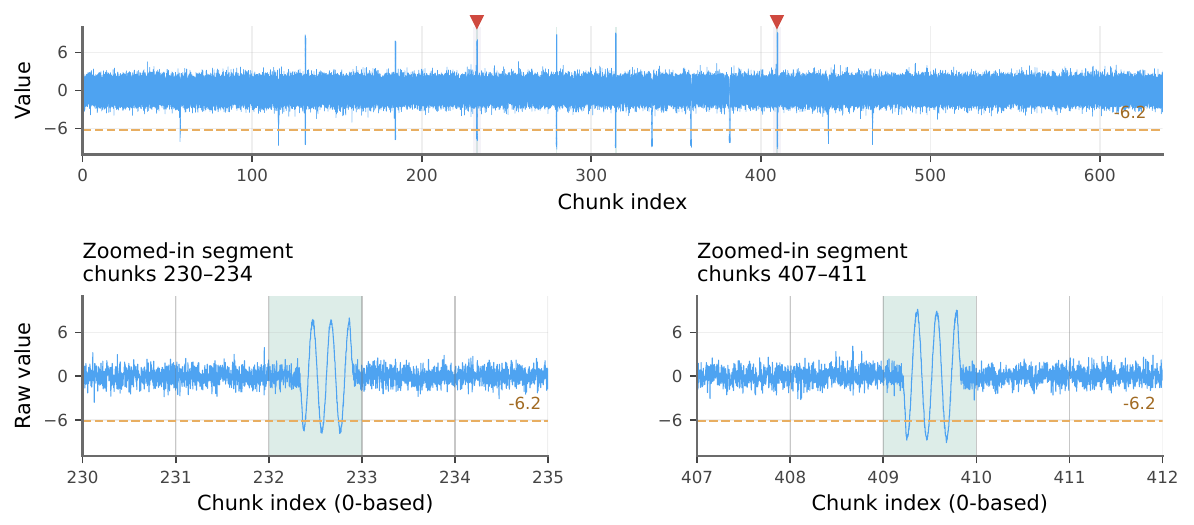}
\end{minipage}\\[4pt]
 & \textit{Note:} Only raw samples are provided as signal input. Shading, labels and guide lines are for illustration only and are not model inputs.\\[6pt]
\midrule
\textbf{Ground Truth} & \textbf{Response time:} chunk 232. \quad \textbf{Reported content:} sine \#3 --- square at 0, triangle at 2.\\[4pt]
 & \textbf{Response time:} chunk 279. \quad \textbf{Reported content:} sine \#4 --- square at 0, triangle at 2.\\[4pt]
 & \textbf{Response time:} chunk 314. \quad \textbf{Reported content:} sine \#5 --- square at 0, triangle at 2.\\[4pt]
 & \textbf{Response time:} chunk 409. \quad \textbf{Reported content:} sine \#6 --- square at 3, triangle at 2.\\[4pt]
\midrule
\textbf{Evidence} & The dashed horizontal line marks sine count = 3. Green dotted lines mark responses at chunks 232, 279, 314 and 409; the labels 3, 4, 5 and 6 give the sine count at each response: each new sine burst from the third onward triggers a report. Quiet chunks and other shapes do not trigger a report even while the sine count remains at or above 3. Counts include the current burst. \textit{For explanation only; not model input.}\\[3pt]
 & \begin{minipage}[t]{\linewidth}\vspace{0pt}\centering
\includegraphics[width=\linewidth]{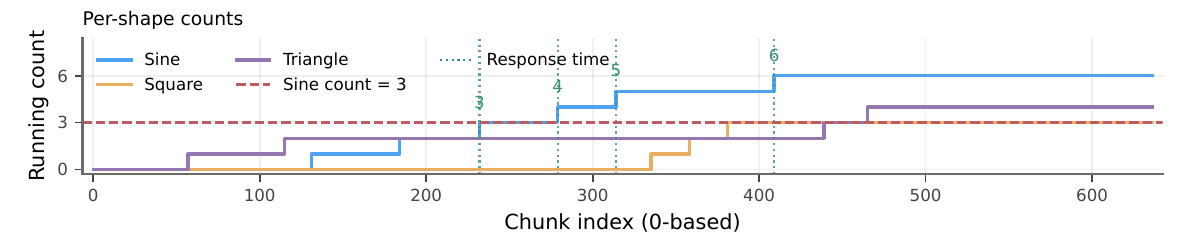}
\end{minipage}\\[4pt]
\midrule
\textbf{Scoring criteria} & \textbf{Timing:} report target-shape bursts starting at the specified count. \textbf{Content:} exactly match the target shape, its index, and the counts of the other shapes.\\[4pt]
\bottomrule
\end{tabularx}
\end{table}
\clearpage

\begin{table}[!ht]
\centering
\small
\setlength{\abovecaptionskip}{0pt}
\setlength{\belowcaptionskip}{7pt}
\caption{Task 9 (Ordered Event Tracking): Example with ground-truth response and scoring criteria.}
\label{tab:case-task9}
\begin{tabularx}{\textwidth}{@{}>{\raggedright\arraybackslash}p{2.05cm}Y@{}}
\toprule
\multicolumn{2}{@{}l@{}}{\textbf{Capability C3: Event Tracking}}\\
\midrule
\textbf{Task} & \textbf{\textit{T9: Ordered Event Tracking}}\\[4pt]
\textbf{User} & The signal arrives one chunk at a time (512 samples per chunk; chunks are numbered from 0). Only this sequence matters: first a sine, then a triangle within 24 chunks. Speak at the triangle. Lone bursts and late follow-ups are fine. A burst means a shaped pulse reaching past 5.9; the background's own wiggles never count. Call it exactly as: "\textless{}g\textgreater{} chunks apart", where \textless{}g\textgreater{} is the chunk index of the triangle minus the chunk index of the sine that came before it.\\[6pt]
 & Input time series stream (shown in full; supplied sequentially in chunks of 512 samples):\\[2pt]
 & \begin{minipage}[t]{\linewidth}\vspace{0pt}\centering
\includegraphics[width=\linewidth]{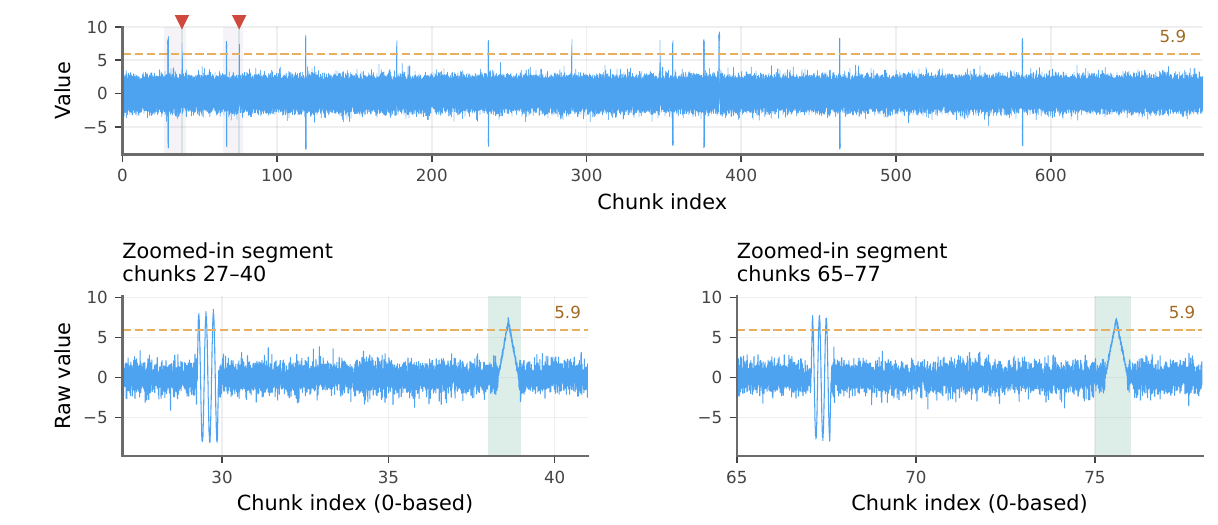}
\end{minipage}\\[4pt]
 & \textit{Note:} Only raw samples are provided as signal input. Shading, labels and guide lines are for illustration only and are not model inputs.\\[6pt]
\midrule
\textbf{Ground Truth} & \textbf{Response time:} chunk 38. \quad \textbf{Reported content:} 9 chunks apart.\\[4pt]
 & \textbf{Response time:} chunk 75. \quad \textbf{Reported content:} 8 chunks apart.\\[4pt]
\midrule
\textbf{Evidence} & Each close-up contains a qualifying sine-triangle pair with surrounding samples. The evidence shows chunks 0--200: green arrows point from sine to triangle, with gaps of 9 and 8 chunks. Green dotted lines mark the responses at the triangle chunks (38 and 75). \textit{For explanation only; not model input.}\\[3pt]
 & \begin{minipage}[t]{\linewidth}\vspace{0pt}\centering
\includegraphics[width=\linewidth]{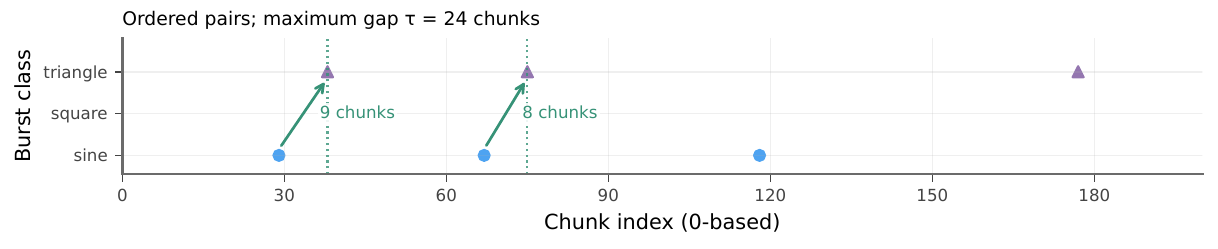}
\end{minipage}\\[4pt]
\midrule
\textbf{Scoring criteria} & \textbf{Timing:} respond at the second burst of a qualifying ordered pair. \textbf{Content:} score the gap between the two bursts.\\[4pt]
\bottomrule
\end{tabularx}
\end{table}
\clearpage

\subsection{C4: Event Reference}
\begin{table}[!ht]
\centering
\small
\setlength{\abovecaptionskip}{0pt}
\setlength{\belowcaptionskip}{7pt}
\caption{Task 10 (Shape Reference): Example with ground-truth response and scoring criteria.}
\label{tab:case-task10}
\begin{tabularx}{\textwidth}{@{}>{\raggedright\arraybackslash}p{2.05cm}Y@{}}
\toprule
\multicolumn{2}{@{}l@{}}{\textbf{Capability C4: Event Reference}}\\
\midrule
\textbf{Task} & \textbf{\textit{T10: Shape Reference}}\\[4pt]
\textbf{User} & The signal arrives one chunk at a time (512 samples per chunk; chunks are numbered from 0). Right at the start you'll see one burst -- keep its picture in mind. I want to know every time that exact look returns. The rest, stay quiet. A burst means a shaped pulse dropping below -6.0; the background's own wiggles never count. Report each return exactly as: "\textless{}g\textgreater{} chunks after the opener", where \textless{}g\textgreater{} is the chunk index of this burst minus the chunk index of the opening burst.\\[6pt]
 & Input time series stream (shown in full; supplied sequentially in chunks of 512 samples):\\[2pt]
 & \begin{minipage}[t]{\linewidth}\vspace{0pt}\centering
\includegraphics[width=\linewidth]{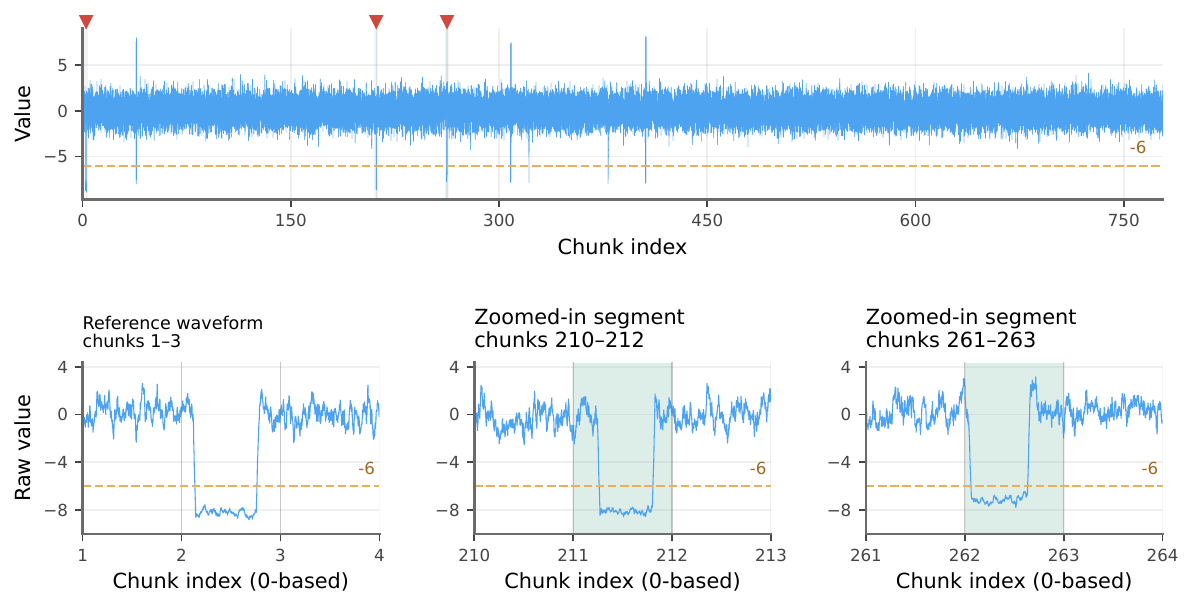}
\end{minipage}\\[4pt]
 & \textit{Note:} Only raw samples are provided as signal input. Shading, labels and guide lines are for illustration only and are not model inputs.\\[6pt]
\midrule
\textbf{Ground Truth} & \textbf{Response time:} chunk 211. \quad \textbf{Reported content:} 209 chunks after the opener.\\[4pt]
 & \textbf{Response time:} chunk 262. \quad \textbf{Reported content:} 260 chunks after the opener.\\[4pt]
\midrule
\textbf{Evidence} & The opening burst is square-shaped. Other bursts are sine- or triangle-shaped, so crossing the threshold alone does not trigger a response. Examples of both nonmatching shapes are shown below; only recurrences of the opening waveform are reported. \textit{For explanation only; not model input.}\\[3pt]
 & \begin{minipage}[t]{\linewidth}\vspace{0pt}\centering
\includegraphics[width=\linewidth]{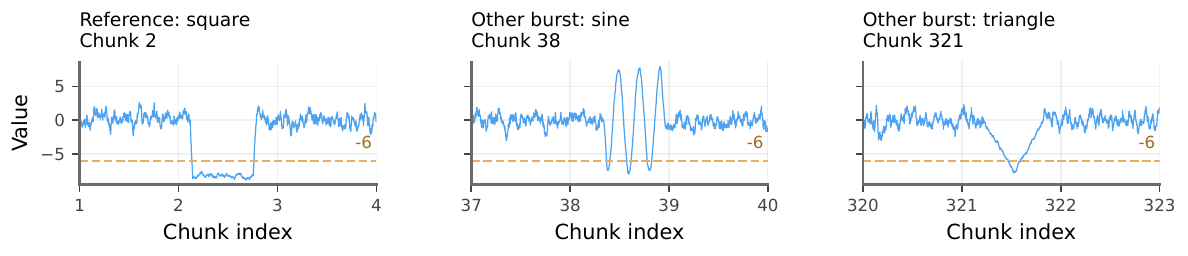}
\end{minipage}\\[4pt]
\midrule
\textbf{Scoring criteria} & \textbf{Timing:} respond when the opening waveform recurs. \textbf{Content:} score the elapsed chunks since the reference burst.\\[4pt]
\bottomrule
\end{tabularx}
\end{table}
\clearpage

\begin{table}[!ht]
\centering
\small
\setlength{\abovecaptionskip}{0pt}
\setlength{\belowcaptionskip}{7pt}
\caption{Task 11 (Numerical Reference): Example with ground-truth response and scoring criteria.}
\label{tab:case-task11}
\begin{tabularx}{\textwidth}{@{}>{\raggedright\arraybackslash}p{2.05cm}Y@{}}
\toprule
\multicolumn{2}{@{}l@{}}{\textbf{Capability C4: Event Reference}}\\
\midrule
\textbf{Task} & \textbf{\textit{T11: Numerical Reference}}\\[4pt]
\textbf{User} & The signal arrives one chunk at a time (512 samples per chunk; chunks are numbered from 0). Track the highest peak so far. A newcomer that tops it gets reported with both values, new and old. The rest, silence. A burst only counts if it reaches past 5.3; anything tamer never enters the record book. The first burst that reaches past 5.3 sets the initial record and is not reported; from then on, report only bursts that beat the current record. Use exactly: "new record \textless{}value\textgreater{}, previous \textless{}value\textgreater{}", where the first is the peak (for a below-the-line watch: the lowest point) of this burst and the second is that of the previous record.\\[6pt]
 & Input time series stream (shown in full; supplied sequentially in chunks of 512 samples):\\[2pt]
 & \begin{minipage}[t]{\linewidth}\vspace{0pt}\centering
\includegraphics[width=\linewidth]{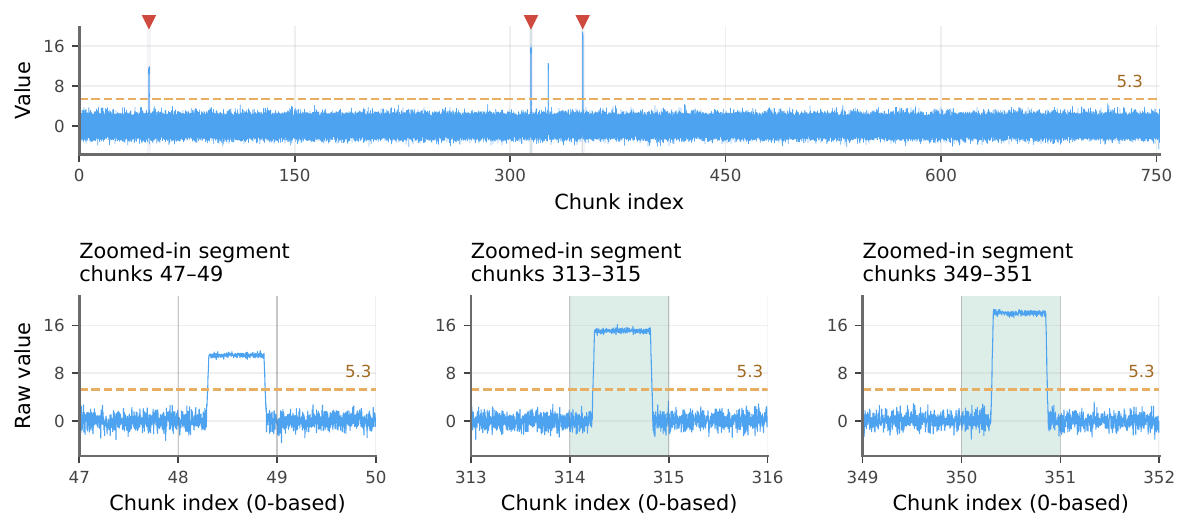}
\end{minipage}\\[4pt]
 & \textit{Note:} Only raw samples are provided as signal input. Shading, labels and guide lines are for illustration only and are not model inputs.\\[6pt]
\midrule
\textbf{Ground Truth} & \textbf{Response time:} chunk 314. \quad \textbf{Reported content:} new record 15.1, previous 11.\\[4pt]
 & \textbf{Response time:} chunk 350. \quad \textbf{Reported content:} new record 18.1, previous 15.1.\\[4pt]
\midrule
\textbf{Evidence} & The initial peak at chunk 48 sets the record to 11 without a response. The peaks at chunks 314 and 350 raise it to 15.1 and 18.1, respectively. Their increases over the previous record are 4.1 and 3.0. \textit{For explanation only; not model input.}\\[3pt]
 & \begin{minipage}[t]{\linewidth}\vspace{0pt}\centering
\includegraphics[width=\linewidth]{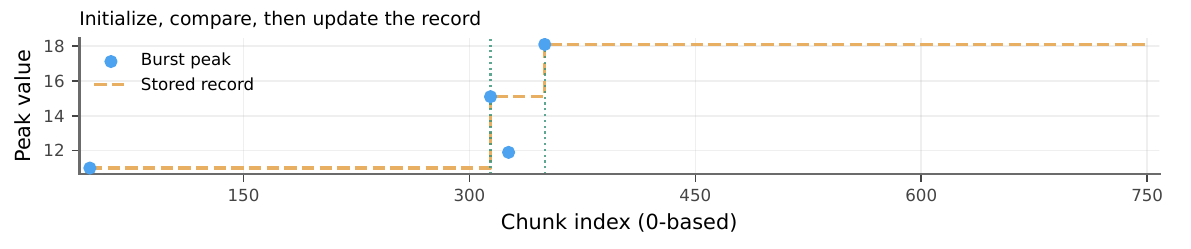}
\end{minipage}\\[4pt]
\midrule
\textbf{Scoring criteria} & \textbf{Timing:} the first qualifying burst only initializes the record; respond when a later burst exceeds it. \textbf{Content:} score the previous peak and the absolute difference between the reported new and previous peaks, then average the two field scores.\\[4pt]
\bottomrule
\end{tabularx}
\end{table}
\clearpage

\subsection{C5: Online Instructions}
\begin{table}[!ht]
\centering
\small
\setlength{\abovecaptionskip}{0pt}
\setlength{\belowcaptionskip}{7pt}
\caption{Task 12 (On-Demand Recall): Example with ground-truth response and scoring criteria.}
\label{tab:case-task12}
\begin{tabularx}{\textwidth}{@{}>{\raggedright\arraybackslash}p{2.05cm}Y@{}}
\toprule
\multicolumn{2}{@{}l@{}}{\textbf{Capability C5: Online Instructions}}\\
\midrule
\textbf{Task} & \textbf{\textit{T12: On-Demand Recall}}\\[4pt]
\textbf{User} & The signal arrives one chunk at a time (512 samples per chunk; chunks are numbered from 0). Just keep watch and keep count of the bursts -- I'll ask you along the way. Answer my questions with the number; otherwise stay quiet. A burst means a shaped pulse dropping below -3.7; the background's own wiggles never count. Answer with just the number, or just the shape name, and nothing else.\\[6pt]
 & \textbf{Question at chunk 268:} What's the count on triangle-shaped bursts up to now? Answer with just the number.\\[4pt]
 & \textbf{Question at chunk 520:} How many square-shaped bursts so far? Answer with just the number.\\[4pt]
 & Input time series stream (shown in full; supplied sequentially in chunks of 512 samples):\\[2pt]
 & \begin{minipage}[t]{\linewidth}\vspace{0pt}\centering
\includegraphics[width=\linewidth]{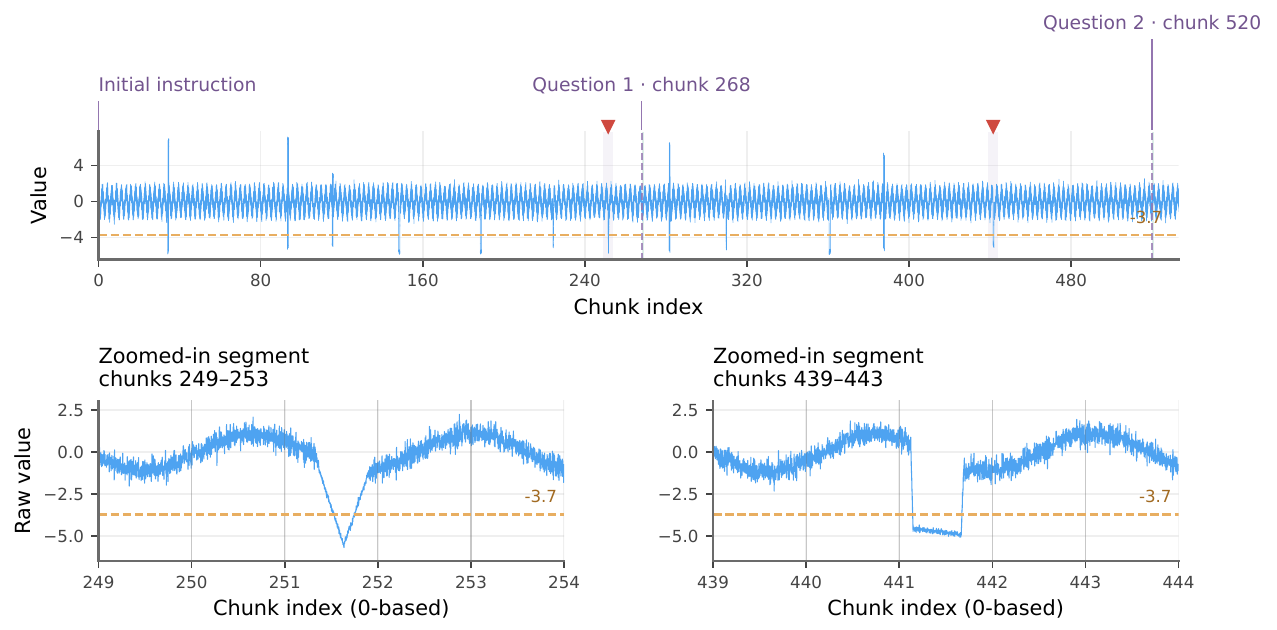}
\end{minipage}\\[4pt]
 & \textit{Note:} Only raw samples are provided as signal input. Shading, labels and guide lines are for illustration only and are not model inputs.\\[6pt]
\midrule
\textbf{Ground Truth} & \textbf{Response time:} chunk 268. \quad \textbf{Reported content:} 2.\\[4pt]
 & \textbf{Response time:} chunk 520. \quad \textbf{Reported content:} 4.\\[4pt]
\midrule
\textbf{Evidence} & The initial instruction requests counting but does not specify which shape will be queried. Separate shape counts must therefore be retained. Question 1 arrives at chunk 268, after the triangle burst at 251 shown in the first close-up; Question 2 arrives at 520, after the square burst at 441 in the second. Green dotted lines mark the question/response times; labels 2 and 4 give the answers. Bursts themselves require silence. \textit{For explanation only; not model input.}\\[3pt]
 & \begin{minipage}[t]{\linewidth}\vspace{0pt}\centering
\includegraphics[width=\linewidth]{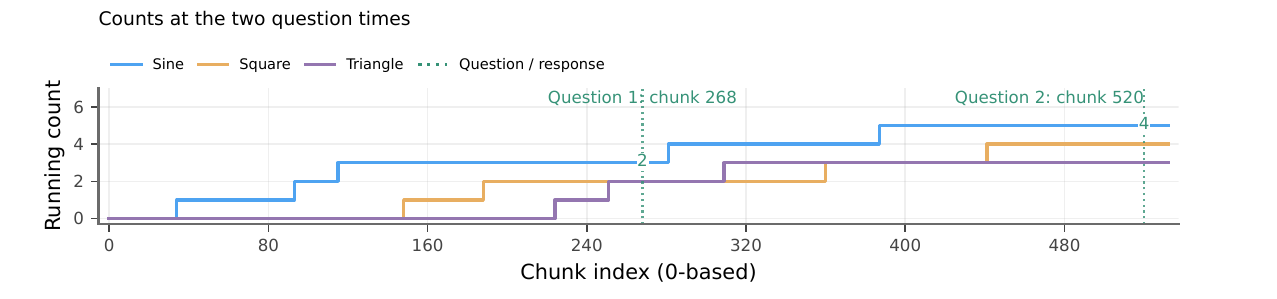}
\end{minipage}\\[4pt]
\midrule
\textbf{Scoring criteria} & \textbf{Timing:} respond only to the in-stream questions. \textbf{Content:} score the requested count from a number-only reply.\\[4pt]
\bottomrule
\end{tabularx}
\end{table}
\clearpage

\begin{table}[!ht]
\centering
\small
\setlength{\abovecaptionskip}{0pt}
\setlength{\belowcaptionskip}{7pt}
\caption{Task 13 (Rule Replacement): Example with ground-truth response and scoring criteria.}
\label{tab:case-task13}
\begin{tabularx}{\textwidth}{@{}>{\raggedright\arraybackslash}p{2.05cm}Y@{}}
\toprule
\multicolumn{2}{@{}l@{}}{\textbf{Capability C5: Online Instructions}}\\
\midrule
\textbf{Task} & \textbf{\textit{T13: Rule Replacement}}\\[4pt]
\textbf{User} & The signal arrives one chunk at a time (512 samples per chunk; chunks are numbered from 0). Count the faults quietly. The 5th and every one after it gets reported. A fault means the signal shooting past 6.8; smaller wiggles are not faults. Report each as: "fault \#\textless{}n\textgreater{} --- \textless{}g\textgreater{} chunks since the last", where \textless{}n\textgreater{} counts every fault since the start (silent ones included) and \textless{}g\textgreater{} is this fault's chunk index minus the previous fault's.\\[6pt]
 & \textbf{Instruction update at chunk 108:} New rule: report from the 7th one on. Forget the 5th.\\[4pt]
 & Input time series stream (shown in full; supplied sequentially in chunks of 512 samples):\\[2pt]
 & \begin{minipage}[t]{\linewidth}\vspace{0pt}\centering
\includegraphics[width=\linewidth]{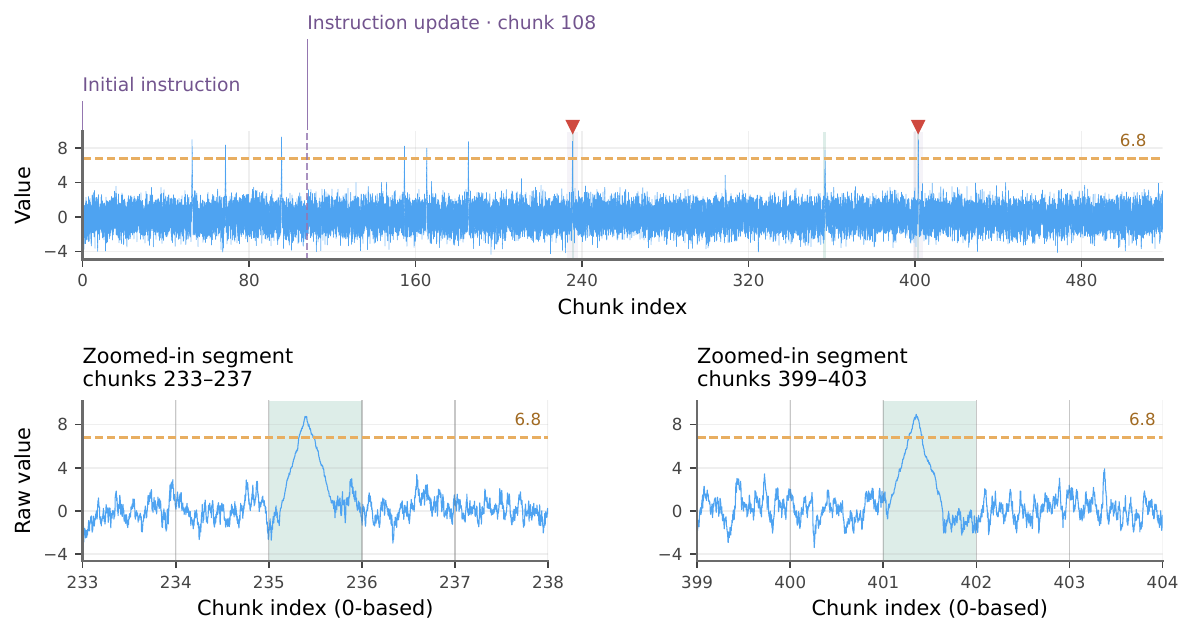}
\end{minipage}\\[4pt]
 & \textit{Note:} Only raw samples are provided as signal input. Shading, labels and guide lines are for illustration only and are not model inputs.\\[6pt]
\midrule
\textbf{Ground Truth} & \textbf{Response time:} chunk 235. \quad \textbf{Reported content:} fault \#7 --- 50 chunks since the last.\\[4pt]
 & \textbf{Response time:} chunk 356. \quad \textbf{Reported content:} fault \#8 --- 121 chunks since the last.\\[4pt]
 & \textbf{Response time:} chunk 401. \quad \textbf{Reported content:} fault \#9 --- 45 chunks since the last.\\[4pt]
\midrule
\textbf{Evidence} & At chunk 108, the reporting threshold changes from 5 to 7 without resetting the count. Orange circles mark faults 5 and 6: the original rule would require reports, but the updated rule requires silence. Green lines mark the actual required reports for faults 7, 8 and 9. This sample has no required report before the update. \textit{For explanation only; not model input.}\\[3pt]
 & \begin{minipage}[t]{\linewidth}\vspace{0pt}\centering
\includegraphics[width=\linewidth]{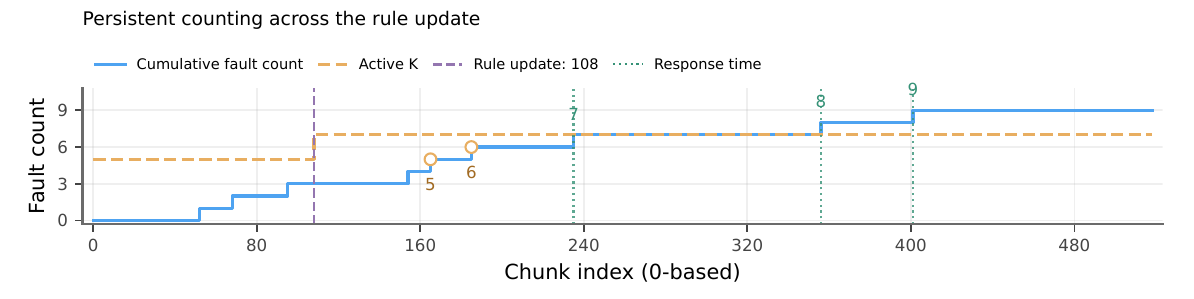}
\end{minipage}\\[4pt]
\midrule
\textbf{Scoring criteria} & \textbf{Timing:} apply the new count threshold immediately without resetting the count. \textbf{Content:} score the fault index and the gap from the preceding fault, with equal field weights.\\[4pt]
\bottomrule
\end{tabularx}
\end{table}
\clearpage

\begin{table}[!ht]
\centering
\small
\setlength{\abovecaptionskip}{0pt}
\setlength{\belowcaptionskip}{7pt}
\caption{Task 14 (Rule Addition): Example with ground-truth response and scoring criteria.}
\label{tab:case-task14}
\begin{tabularx}{\textwidth}{@{}>{\raggedright\arraybackslash}p{2.05cm}Y@{}}
\toprule
\multicolumn{2}{@{}l@{}}{\textbf{Capability C5: Online Instructions}}\\
\midrule
\textbf{Task} & \textbf{\textit{T14: Rule Addition}}\\[4pt]
\textbf{User} & The signal arrives one chunk at a time (512 samples per chunk; chunks are numbered from 0). The line to watch is 10.3. Any burst going past it gets reported with its peak. Rule: whenever a burst's highest reading rises above 10.3, say "peaked at \textless{}reading\textgreater{}", where \textless{}reading\textgreater{} is the highest reading in that chunk. A second rule will be added part-way through the stream; when both rules apply to the same burst, put both in one message, in either order.\\[6pt]
 & \textbf{Instruction update at chunk 76:} One more rule -- sine-shaped bursts, the 2nd and beyond (count from the beginning), report them as well. Rule: whenever a sine burst is the 2nd or later, counting from the very start, say "sine \#\textless{}n\textgreater{}", where \textless{}n\textgreater{} is how many sine bursts have occurred so far including this one. When both rules apply to the same burst, put both in one message, in either order.\\[4pt]
 & Input time series stream (shown in full; supplied sequentially in chunks of 512 samples):\\[2pt]
 & \begin{minipage}[t]{\linewidth}\vspace{0pt}\centering
\includegraphics[width=\linewidth]{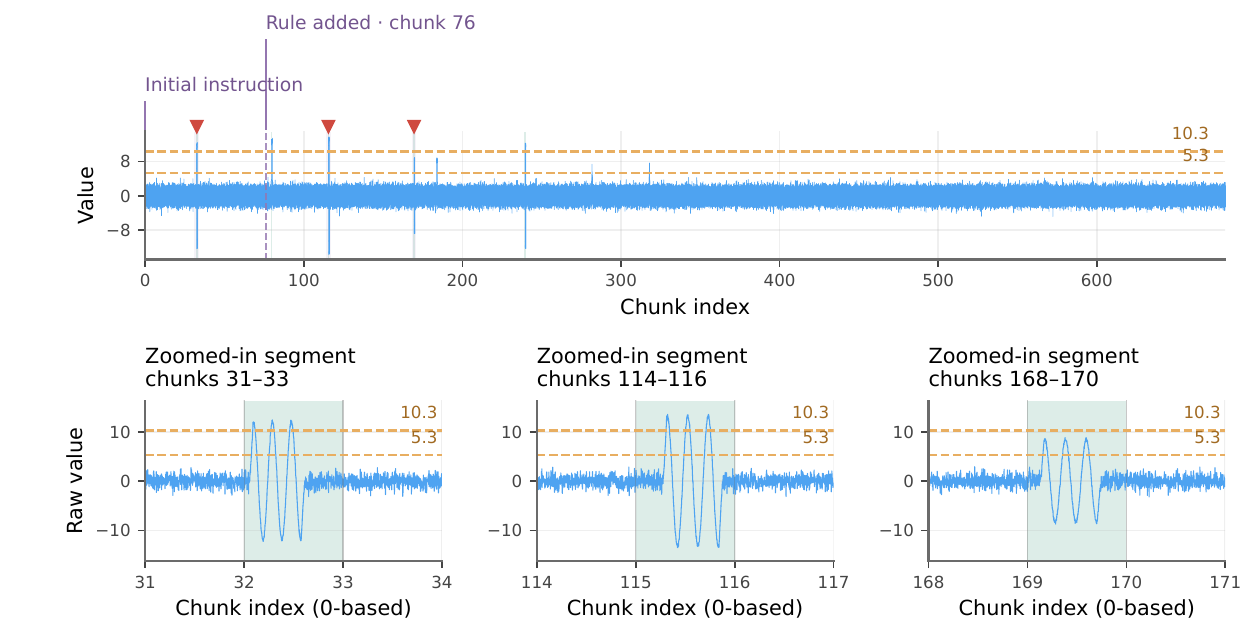}
\end{minipage}\\[4pt]
 & \textit{Note:} Only raw samples are provided as signal input. Shading, labels and guide lines are for illustration only and are not model inputs.\\[6pt]
\midrule
\textbf{Ground Truth} & \textbf{Response time:} chunk 32. \quad \textbf{Reported content:} peaked at 12.\\[4pt]
 & \textbf{Response time:} chunk 79. \quad \textbf{Reported content:} peaked at 12.6.\\[4pt]
 & \textbf{Response time:} chunk 115. \quad \textbf{Reported content:} sine \#2, peaked at 13.1.\\[4pt]
 & \textbf{Response time:} chunk 169. \quad \textbf{Reported content:} sine \#3.\\[4pt]
 & \textbf{Response time:} chunk 239. \quad \textbf{Reported content:} sine \#4, peaked at 11.8.\\[4pt]
\midrule
\textbf{Evidence} & At chunk 76, a shape-count rule is added while the original peak rule remains active. The new count includes bursts from before the update. The purple line marks this insertion. Blue labels give peak readings; orange labels give sine-burst counts. Green dotted lines mark responses. At chunks 115 and 239 both rules fire, so both values belong in one reply. \textit{For explanation only; not model input.}\\[3pt]
 & \begin{minipage}[t]{\linewidth}\vspace{0pt}\centering
\includegraphics[width=\linewidth]{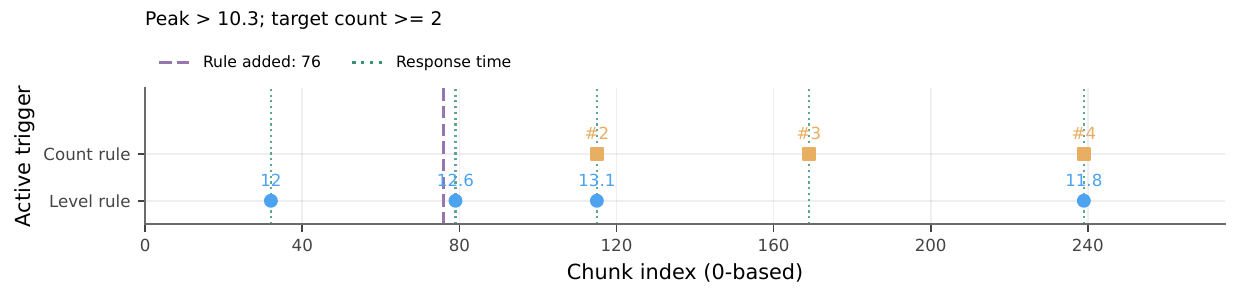}
\end{minipage}\\[4pt]
\midrule
\textbf{Scoring criteria} & \textbf{Timing:} combine both facts in one reply when the active rules fire together. \textbf{Content:} score the rule set, target count and/or excess above the peak threshold, with equal weights over the applicable fields.\\[4pt]
\bottomrule
\end{tabularx}
\end{table}
\clearpage

\subsection{C6: Ultra-Long Reliability}
\begin{table}[!ht]
\centering
\small
\setlength{\abovecaptionskip}{0pt}
\setlength{\belowcaptionskip}{7pt}
\caption{Task 15 (Sustained Vigilance): Example with ground-truth response and scoring criteria.}
\label{tab:case-task15}
\begin{tabularx}{\textwidth}{@{}>{\raggedright\arraybackslash}p{2.05cm}Y@{}}
\toprule
\multicolumn{2}{@{}l@{}}{\textbf{Capability C6: Ultra-Long Reliability}}\\
\midrule
\textbf{Task} & \textbf{\textit{T15: Sustained Vigilance}}\\[4pt]
\textbf{User} & The signal arrives one chunk at a time (512 samples per chunk; chunks are numbered from 0). The line is 3.4. Report every crossing, however long the quiet stretches get. One alarm per crossing: report the chunk where the reading first goes over the line, not the chunks that stay over. Each time, report exactly: "peaked at \textless{}value\textgreater{}", where \textless{}value\textgreater{} is the highest reading in that chunk.\\[6pt]
 & Input time series stream (shown in full; supplied sequentially in chunks of 512 samples):\\[2pt]
 & \begin{minipage}[t]{\linewidth}\vspace{0pt}\centering
\includegraphics[width=\linewidth]{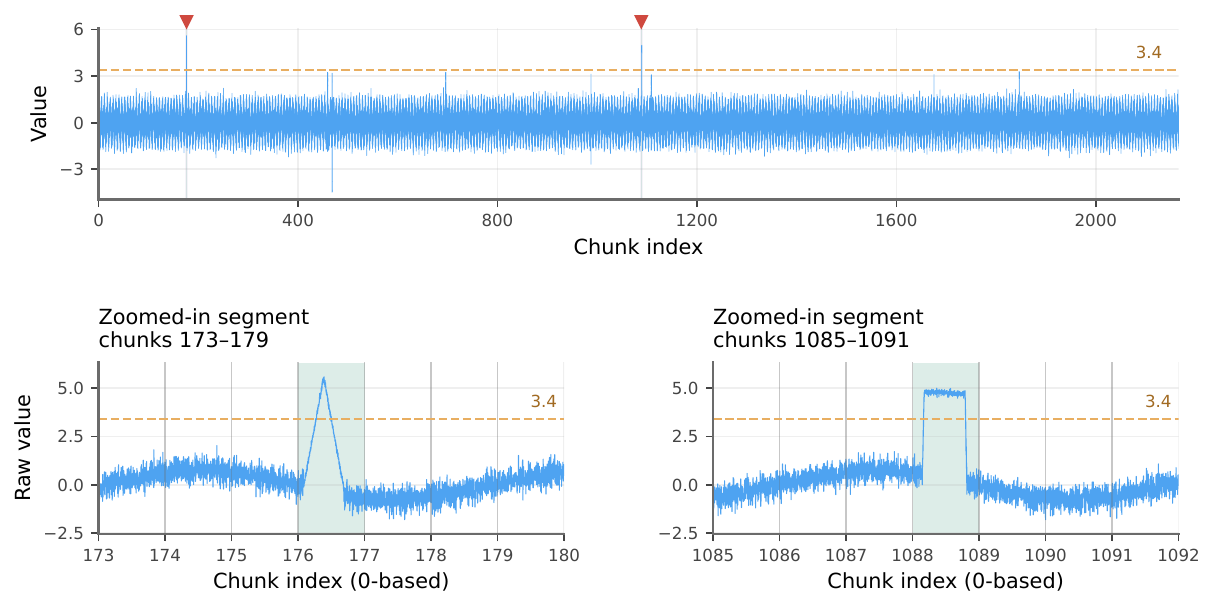}
\end{minipage}\\[4pt]
 & \textit{Note:} Only raw samples are provided as signal input. Shading, labels and guide lines are for illustration only and are not model inputs.\\[6pt]
\midrule
\textbf{Ground Truth} & \textbf{Response time:} chunk 176. \quad \textbf{Reported content:} peaked at 5.5813.\\[4pt]
 & \textbf{Response time:} chunk 1088. \quad \textbf{Reported content:} peaked at 4.993.\\[4pt]
\midrule
\textbf{Scoring criteria} & \textbf{Timing:} report only the first chunk of each crossing. \textbf{Content:} score the reported peak as an excess above the threshold.\\[4pt]
\bottomrule
\end{tabularx}
\end{table}
\clearpage

\begin{table}[!ht]
\centering
\small
\setlength{\abovecaptionskip}{0pt}
\setlength{\belowcaptionskip}{7pt}
\caption{Task 16 (Long-Gap Reference): Example with ground-truth response and scoring criteria.}
\label{tab:case-task16}
\begin{tabularx}{\textwidth}{@{}>{\raggedright\arraybackslash}p{2.05cm}Y@{}}
\toprule
\multicolumn{2}{@{}l@{}}{\textbf{Capability C6: Ultra-Long Reliability}}\\
\midrule
\textbf{Task} & \textbf{\textit{T16: Long-Gap Reference}}\\[4pt]
\textbf{User} & The signal arrives one chunk at a time (512 samples per chunk; chunks are numbered from 0). The very first thing on this line is a sample burst. Matching bursts after it: report each one. Non-matching: not a word. A burst means a shaped pulse dropping below -10.8; the background's own wiggles never count. Report each return exactly as: "\textless{}g\textgreater{} chunks after the opener", where \textless{}g\textgreater{} is the chunk index of this burst minus the chunk index of the opening burst.\\[6pt]
 & Input time series stream (shown in full; supplied sequentially in chunks of 512 samples):\\[2pt]
 & \begin{minipage}[t]{\linewidth}\vspace{0pt}\centering
\includegraphics[width=\linewidth]{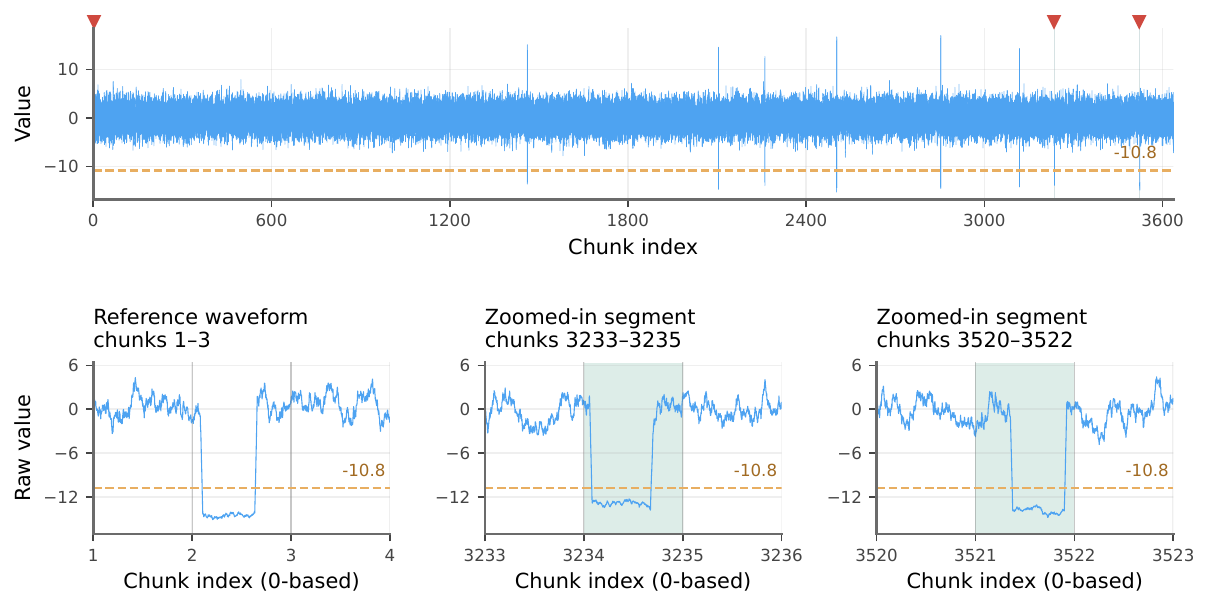}
\end{minipage}\\[4pt]
 & \textit{Note:} Only raw samples are provided as signal input. Shading, labels and guide lines are for illustration only and are not model inputs.\\[6pt]
\midrule
\textbf{Ground Truth} & \textbf{Response time:} chunk 3234. \quad \textbf{Reported content:} 3232 chunks after the opener.\\[4pt]
 & \textbf{Response time:} chunk 3521. \quad \textbf{Reported content:} 3519 chunks after the opener.\\[4pt]
\midrule
\textbf{Evidence} & The opening waveform is square-shaped. All six intervening bursts are sine-shaped and do not match it; one is shown below. The square waveform returns at chunks 3234 and 3521, giving gaps of 3232 and 3519 from the reference at chunk 2. \textit{For explanation only; not model input.}\\[3pt]
 & \begin{minipage}[t]{\linewidth}\vspace{0pt}\centering
\includegraphics[width=\linewidth]{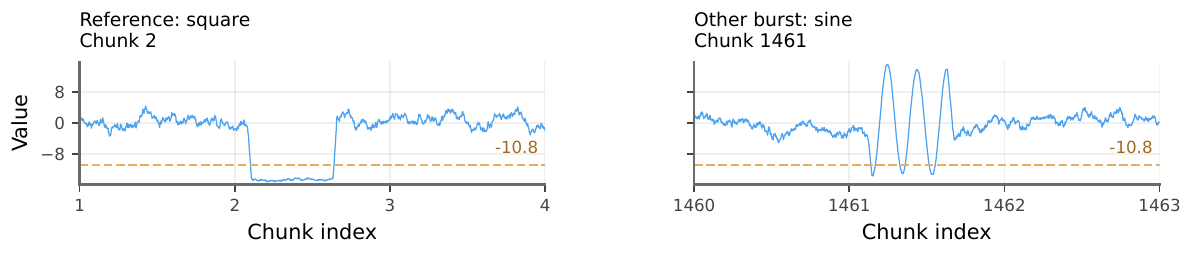}
\end{minipage}\\[4pt]
\midrule
\textbf{Scoring criteria} & \textbf{Timing:} keep the opening burst silent and report each later matching burst. \textbf{Content:} score the return chunk index minus the reference chunk index.\\[4pt]
\bottomrule
\end{tabularx}
\end{table}
\clearpage

\begin{table}[!ht]
\centering
\small
\setlength{\abovecaptionskip}{0pt}
\setlength{\belowcaptionskip}{7pt}
\caption{Task 17 (Sustained Reporting): Example with ground-truth response and scoring criteria.}
\label{tab:case-task17}
\begin{tabularx}{\textwidth}{@{}>{\raggedright\arraybackslash}p{2.05cm}Y@{}}
\toprule
\multicolumn{2}{@{}l@{}}{\textbf{Capability C6: Ultra-Long Reliability}}\\
\midrule
\textbf{Task} & \textbf{\textit{T17: Sustained Reporting}}\\[4pt]
\textbf{User} & The signal arrives one chunk at a time (512 samples per chunk; chunks are numbered from 0). Long shift: keep count of the bursts by shape. Every 500 chunks -- at chunk 500, 1000, and so on -- give me the running totals unprompted. I may also ask in between. Bursts themselves: no comment. Two kinds of replies. Scheduled report: when chunk 500, 1000, 1500, ... arrives, say exactly "sine \textless{}n\textgreater{}, square \textless{}n\textgreater{}, triangle \textless{}n\textgreater{}", where each \textless{}n\textgreater{} is how many bursts of that shape arrived before this chunk (bursts in the current chunk are not counted yet). Answer to my question: reply with just the number; the number is the count of that shape that arrived before the chunk the question comes in.\\[6pt]
 & \textbf{Question at chunk 1916:} How many sine-shaped bursts so far? Answer with just the number.\\[4pt]
 & \textbf{Question at chunk 2098:} How many triangle-shaped bursts so far? Answer with just the number.\\[4pt]
 & Input time series stream (shown in full; supplied sequentially in chunks of 512 samples):\\[2pt]
 & \begin{minipage}[t]{\linewidth}\vspace{0pt}\centering
\includegraphics[width=.90\linewidth]{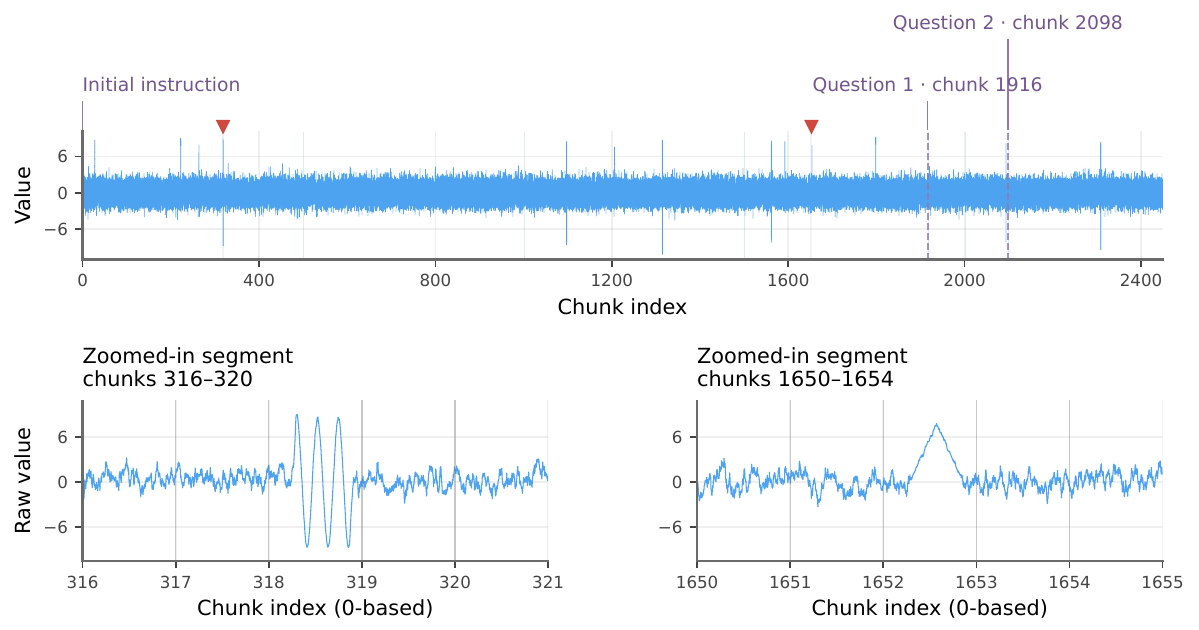}
\end{minipage}\\[4pt]
 & \textit{Note:} Only raw samples are provided as signal input. Shading, labels and guide lines are for illustration only and are not model inputs.\\[6pt]
\midrule
\textbf{Ground Truth} & \textbf{Response time:} chunk 500. \quad \textbf{Reported content:} sine 1, square 1, triangle 2.\\[4pt]
 & \textbf{Response time:} chunk 1000. \quad \textbf{Reported content:} sine 1, square 1, triangle 2.\\[4pt]
 & \textbf{Response time:} chunk 1500. \quad \textbf{Reported content:} sine 3, square 1, triangle 3.\\[4pt]
 & \textbf{Response time:} chunk 1916. \quad \textbf{Reported content:} 4.\\[4pt]
 & \textbf{Response time:} chunk 2000. \quad \textbf{Reported content:} sine 4, square 3, triangle 4.\\[4pt]
 & \textbf{Response time:} chunk 2098. \quad \textbf{Reported content:} 4.\\[4pt]
\midrule
\textbf{Evidence} & Green dotted lines mark scheduled reports; their labels list sine / square / triangle counts. Purple lines mark the two questions, and the labels 4 give their answers. Counts exclude the current chunk. The raw close-ups show a sine burst and a triangle burst contributing to later totals. \textit{For explanation only; not model input.}\\[3pt]
 & \begin{minipage}[t]{\linewidth}\vspace{0pt}\centering
\includegraphics[width=.90\linewidth]{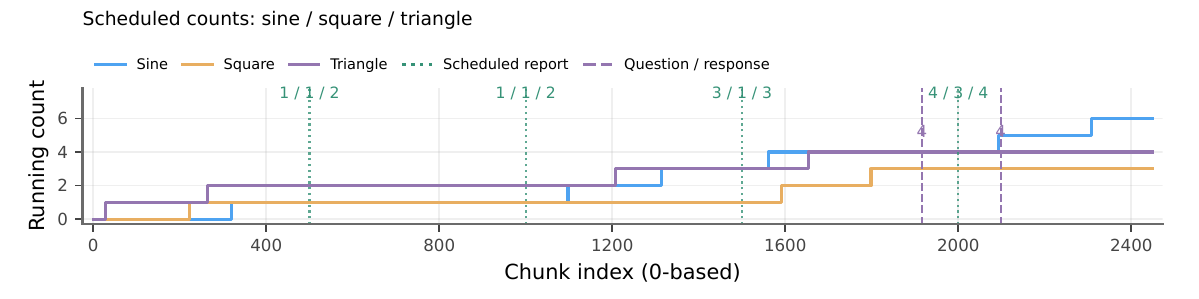}
\end{minipage}\\[4pt]
\midrule
\textbf{Scoring criteria} & \textbf{Timing:} report at scheduled chunks and answer each intervening question. \textbf{Content:} scheduled reports require exact per-shape counts; question replies are scored as numbers, always excluding current-chunk bursts.\\[4pt]
\bottomrule
\end{tabularx}
\end{table}
\clearpage

\begin{table}[!ht]
\centering
\small
\setlength{\abovecaptionskip}{0pt}
\setlength{\belowcaptionskip}{7pt}
\caption{Task 18 (Cumulative Evidence): Example with ground-truth response and scoring criteria.}
\label{tab:case-task18}
\begin{tabularx}{\textwidth}{@{}>{\raggedright\arraybackslash}p{2.05cm}Y@{}}
\toprule
\multicolumn{2}{@{}l@{}}{\textbf{Capability C6: Ultra-Long Reliability}}\\
\midrule
\textbf{Task} & \textbf{\textit{T18: Cumulative Evidence}}\\[4pt]
\textbf{User} & The signal arrives one chunk at a time (512 samples per chunk; chunks are numbered from 0). This machine ages slowly. If the signal's level walks away from where it began and the cumulative shift (skip anything under a quarter) tops 60, report it. A single report will do. Treat the first 100 chunks as normal: their average level is the starting level, and the spread (standard deviation) of their samples is one unit. From chunk 100 on, keep a running score: add this chunk's mean minus the starting level, measured in those units, then subtract 0.25; if the score falls below zero, reset it to zero. Keep a second score the same way for downward moves, using the starting level minus this chunk's mean. The moment either score exceeds 60, say so once and then stay silent. Report exactly: "started near chunk \textless{}n\textgreater{}, moved \textless{}amount\textgreater{}", where \textless{}n\textgreater{} is the chunk at which the winning score last rose from zero, and \textless{}amount\textgreater{} is how far this chunk's mean is from the starting level, in the original reading units, always given as a positive number.\\[6pt]
 & Input time series stream (shown in full; supplied sequentially in chunks of 512 samples):\\[2pt]
 & \begin{minipage}[t]{\linewidth}\vspace{0pt}\centering
\includegraphics[width=\linewidth]{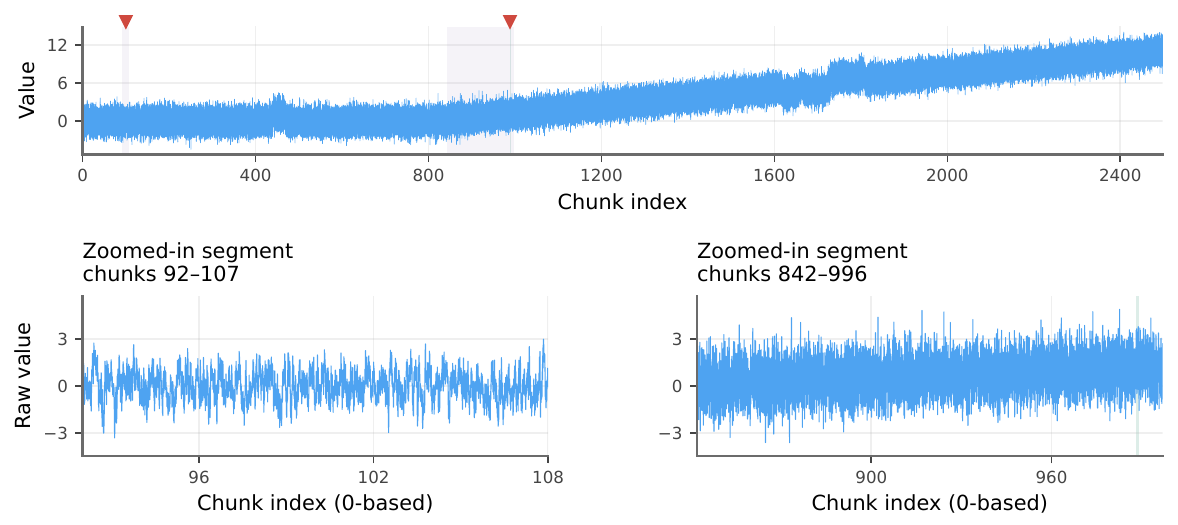}
\end{minipage}\\[4pt]
 & \textit{Note:} Only raw samples are provided as signal input. Shading, labels and guide lines are for illustration only and are not model inputs.\\[6pt]
\midrule
\textbf{Ground Truth} & \textbf{Response time:} chunk 988. \quad \textbf{Reported content:} started near chunk 850, moved 1.4909.\\[4pt]
\midrule
\textbf{Evidence} & The first 100 chunks fix the baseline mean and sample standard deviation. Left (first field): normalized mean shifts, minus 0.25 per chunk, accumulate with resets at zero. The winning score first exceeds 60 at chunk 988, after 139 updates since its last departure from zero at chunk 850. The reported 850 is this start time, not the score itself. Right (second field): at chunk 988, the absolute difference between the current chunk mean and the baseline mean is 1.4909 in raw signal units. This is a current-chunk magnitude, not an accumulated sum. \textit{For explanation only; not model input.}\\[3pt]
 & \begin{minipage}[t]{\linewidth}\vspace{0pt}\centering
\includegraphics[width=\linewidth]{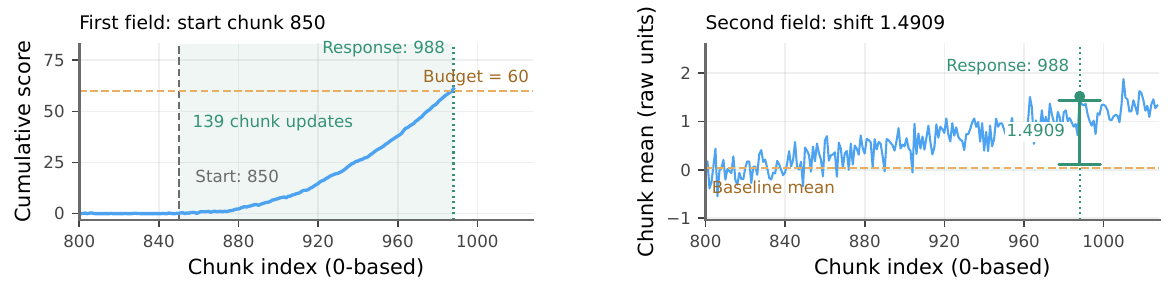}
\end{minipage}\\[4pt]
\midrule
\textbf{Scoring criteria} & \textbf{Timing:} respond once at the first budget crossing. \textbf{Content:} score the start chunk (full credit within 30 chunks; zero at 90 or more) and the current mean's shift from baseline, then average both field scores.\\[4pt]
\bottomrule
\end{tabularx}
\end{table}
\clearpage

\endgroup

\end{document}